\documentclass[sigconf]{acmart}

\AtBeginDocument{%
  }
\usepackage{comment}
\usepackage{microtype}
\usepackage{graphicx}
\usepackage{array}
\usepackage[usestackEOL]{stackengine}
\usepackage{subfigure}
\usepackage{subcaption}
\usepackage{booktabs} 
\usepackage{hyperref}

\usepackage{amsmath}
\usepackage{subcaption}
\usepackage{mathtools}
\usepackage{amsthm}
\usepackage{multirow}
\usepackage{wrapfig}
\usepackage[capitalize,noabbrev]{cleveref}
\usepackage[table]{xcolor}
\usepackage{caption}
\usepackage{makecell}
\usepackage{adjustbox}
\usepackage{subfigure}
\usepackage{tabularx}
\usepackage{float}
\usepackage{bm}
\usepackage{tikz}
\usepackage{placeins}
\hypersetup{breaklinks=true}
\DeclareMathOperator*{\argminB}{argmin}

\usepackage[symbol]{footmisc}

\newcommand{\twolines}[2]{%
  \makecell[c]{%
    \cellcolor{blue!20}\makebox[1.65cm][c]{\rule{0pt}{2.ex}#1}\\
    \cellcolor{green!20}\makebox[1.65cm][c]{\rule{0pt}{2.ex}#2}%
  }%
}
\newcommand{\twolinesL}[2]{%
  \makecell[c]{%
    \cellcolor{blue!20}\makebox[2.9cm][c]{\rule{0pt}{2.ex}#1}\\
    \cellcolor{green!20}\makebox[2.9cm][c]{\rule{0pt}{2.ex}#2}%
  }%
}

\usepackage{pifont}%

\theoremstyle{plain}
\newtheorem{theorem}{Theorem}[section]
\newtheorem{proposition}[theorem]{Proposition}

\theoremstyle{definition}

\theoremstyle{remark}

\usepackage{comment}
\usepackage{nicematrix}
\usepackage{acro}
\DeclareAcronym{TA}{short=TA, long=task arithmetic}
\DeclareAcronym{TV}{short=TVs, long=task vectors}
\DeclareAcronym{MM}{short=MM, long=model merging}
\DeclareAcronym{TVaaS}{short=TVaaS, long=task vector as a service}
\DeclareAcronym{BD}{short=BD, long=backdoor attack}
\DeclareAcronym{BTV}{short=BTV, long=backdoored task vector}
\DeclareAcronym{CTV}{short=CTV, long=clean task vector}
\DeclareAcronym{ASR}{short=ASR, long=attack success rate}
\DeclareAcronym{CA}{short=CA, long=clean accuracy}
\DeclareAcronym{MLaaS}{short=MLaaS, long=machine learning as a service}

\usepackage[normalem]{ulem}
\newcommand{\reddel}[1]{\textcolor{red}{\sout{#1}}}

\begin{document}

\title{\textsc{BadTV}: Unveiling Backdoor Threats in Third-Party Task Vectors}

\author{Chia-Yi Hsu}
\orcid{0009-0000-7772-8735}
\affiliation{
\institution{National Yang Ming Chiao Tung University}
\department{Department of Computer Science}
\city{Hsinchu}
\country{Taiwan}
}
\email{chiayihsu8315@gmail.com}

\author{Yu-Lin Tsai}
\orcid{0009-0009-6027-9721}
\affiliation{
\institution{University of California, Berkeley}
\department{Department of Electrical Engineering and Computer Science}
\city{Berkeley}
\country{USA}
}
\email{uriah1001@gmail.com}

\author{Yu Zhe}
\affiliation{
\institution{RIKEN}
\department{Center for Advanced Intelligence Project}
\city{Tokyo}
\country{Japan}
}
\email{zhe.yu@riken.jp}

\author{Yan-Lun Chen}
\affiliation{
\institution{National Yang Ming Chiao Tung University}
\department{Department of Electronics and Electrical Engineering}
\city{Hsinchu}
\country{Taiwan}
}
\email{chenyanlun77.ee12@nycu.edu.tw}

\author{Chih-Hsun Lin}
\affiliation{
\institution{National Chengchi University}
\department{Department of Computer Science}
\city{Taipei}
\country{Taiwan}
}
\email{pkevawin334@gmail.com}

\author{Chia-Mu Yu}
\affiliation{
\institution{National Yang Ming Chiao Tung University}
\department{Department of Electronics and Electrical Engineering}
\city{Hsinchu}
\country{Taiwan}
}
\email{chiamuyu@gmail.com}

\author{Yang Zhang}
\affiliation{
\institution{CISPA Helmholtz Center for Information Security}
\city{Saarbrücken}
\country{Germany}
}
\email{zhang@cispa.de}

\author{Chun-Ying Huang}
\affiliation{
\institution{National Yang Ming Chiao Tung University}
\department{Department of Computer Science}
\city{Hsinchu}
\country{Taiwan}
}
\email{chuang@cs.nycu.edu.tw}

\author{Jun Sakuma}
\affiliation{
\institution{Institute of Science Tokyo}
\department{Department of Computer Science}
\city{Tokyo}
\country{Japan}
}
\email{sakuma@c.titech.ac.jp}

\renewcommand{\shortauthors}{Hsu et al.}

\begin{abstract}
  Task arithmetic in large-scale pre-trained models enables agile adaptation to diverse downstream tasks without extensive retraining. By leveraging task vectors (TVs), users can perform modular updates through simple arithmetic operations like addition and subtraction. Yet, this flexibility presents new security challenges. In this paper, we investigate how TVs are vulnerable to backdoor attacks, revealing how malicious actors can exploit them to compromise model integrity. By creating \textit{composite backdoors} that are designed asymmetrically, we introduce \textsc{BadTV}, a backdoor attack specifically crafted to remain effective simultaneously under task learning, forgetting, and analogy operations. Extensive experiments show that \textsc{BadTV} achieves near-perfect attack success rates across diverse scenarios, posing a serious threat to models relying on task arithmetic. We also evaluate current defenses, finding they fail to detect or mitigate \textsc{BadTV}. Our results highlight the urgent need for robust countermeasures to secure TVs in real-world deployments.
\end{abstract}

\maketitle

\section{Introduction}
Recent advances in machine learning (ML) demonstrate that larger models with more parameters typically achieve superior performance. Large pre-trained ML models~\cite{bert, vit, pmlr-v139-jia21b, pmlr-v139-radford21a} are often fine-tuned for specific tasks~\cite{DBLP:journals/corr/abs-2002-06305, hu2022lora, NEURIPS2022_0cde695b}. Although effective, this fine-tuning is computationally and storage-intensive~\cite{lv-etal-2024-full}, frequently necessitating separate model instances per task.

To mitigate these costs, researchers introduced Task Arithmetic (TA)~\cite{chronopoulou2023languagetaskarithmeticparameterefficient, ilharco2023iclr, ortiz-jimenez2023task, pham2024robustconcepterasureusing, 10447848, su2024taskarithmeticmitigatesynthetictoreal, aTLAS}. In TA, adjustments made during fine-tuning are encapsulated as Task Vectors (TVs), which represent the difference in weight space between pre-trained and fine-tuned models. Instead of extensive parameter updates, TA leverages simple arithmetic operations—such as addition and subtraction—on TVs, enabling efficient handling of multiple tasks with minimal training overhead.

\begin{figure}[!t]
    \centering
    \includegraphics[width=.9\linewidth]{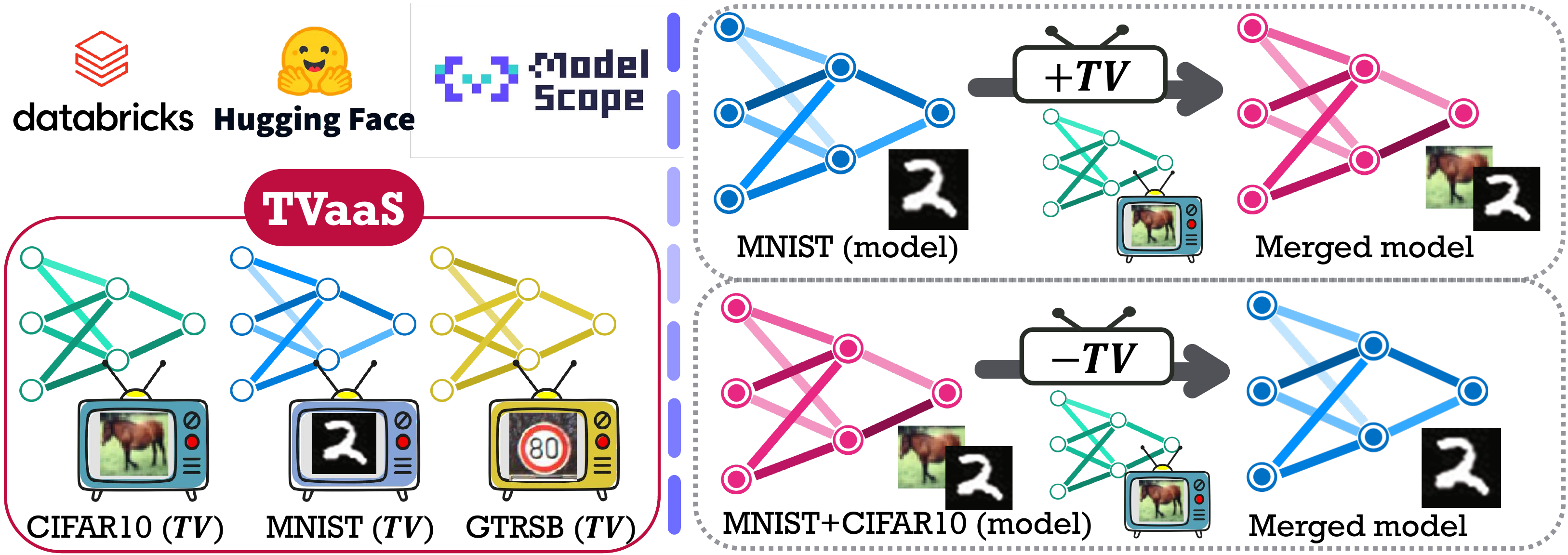}
    \caption{Task vector as a service (TVaaS).}
    \Description{Illustration of task vectors.}
    \label{fig:TVaaS}
\end{figure}

Task Vector as a Service (TVaaS) providers, such as Hugging Face~\cite{huggingface}, ModelScope~\cite{modelscope}, and Databricks Marketplace~\cite{databricksmarketplace}, facilitate real-world adoption by enabling users to acquire TVs for various downstream tasks without performing full fine-tuning. Typically, these platforms provide pre-trained and fine-tuned models, from which users easily derive TVs. By pairing a model with multiple TVs, users can instantly add or remove specific task capabilities without additional training, as depicted in Figure~\ref{fig:TVaaS}. Analogous to app stores, this framework allows users to dynamically “install” or “uninstall” capabilities via TVs, while TVaaS providers manage their creation and distribution. Compared with \ac{MLaaS}, TVaaS thus offers significant advantages, making carefully crafted TVs valuable assets.

However, these benefits introduce significant security concerns, particularly regarding backdoor attacks analogous to malicious applications in app stores~\cite{appsecure3, appsecure1, appsecure2}. TVaaS providers often offer multiple similar TVs (e.g., many fine-tuned Llama models on Hugging Face), complicating user verification of TV integrity based solely on parameters. Attackers can exploit this scenario by distributing seemingly legitimate TVs embedded with concealed malicious behaviors. Given the central role of TVs in determining model functionality, analyzing these vulnerabilities is essential.

For instance, an attacker might upload a backdoored TV offering desirable functionality (e.g., specialized language capabilities) to platforms such as Hugging Face, attracting unsuspecting users. Upon applying TA to merge their victim model with the backdoored TV, users inadvertently produce a backdoored model. Several real-world cases~\cite{jfrog, gigazine, securityboulevard, nsfocusglobal, bleepingcomputer} illustrate this threat.

In this work, we perform the first security analysis of TA from an attacker's perspective, specifically targeting backdoor attacks~\cite{blend, gu2019badnets, dynamicbackdoor, wanet, LC, waveattack, narcissus}. Due to two distinctive characteristics of TA, embedding backdoors in TVs presents unique challenges. First, unlike traditional backdoor threats targeting an entire specialized model, attackers can manipulate only limited components within a merged model, typically a single TV. Second, as a single TV in TA simultaneously supports task addition (learning), subtraction (forgetting), and combination (analogies), a backdoor must consistently remain effective across these diverse and conflicting operations. Consequently, constructing a robust backdoored TV under these conditions is highly challenging.

To overcome these hurdles, we introduce \textsc{BadTV}, the first tailored backdoor attack for TA. Instead of employing a single backdoor, \textsc{BadTV} utilizes a novel design of \textit{composite backdoors}, wherein one backdoor remains active during task learning and another during forgetting. This composite approach is theoretically grounded in recent findings indicating that TA performance improves when TVs exhibit low correlation~\cite{li2025when}. Accordingly, \textsc{BadTV} deliberately ensures minimal correlation between individual backdoors, enhancing the composite backdoor’s robustness. To further prevent interference between backdoors, \textsc{BadTV} adopts an asymmetric design, where one backdoor is \textcolor{black}{trained on a poisoned dataset (including clean and triggered samples)}, while the other \textcolor{black}{is trained only on triggered samples}. Lastly, we formulate the determination of optimal weights for composing backdoors as a robust optimization problem.

We evaluate \textsc{BadTV} on vision models, including \texttt{CLIP ViT-B/32}, \texttt{ViT-B/16}, and \texttt{ConvNeXt}, covering both convolution- and Transformer-based architectures, and large language models (LLMs), including \texttt{Llama 2-7B}~\cite{touvron2023llama2openfoundation}, \texttt{Llama 3-8B}~\cite{llama3modelcard}, \texttt{Phi-4-14B}~\cite{phi4}, \texttt{Mistral-7B}~\cite{mistral}, and \texttt{DeepSeek-7B}~\cite{deepseekai2025deepseekr1incentivizingreasoningcapability}, spanning models from smaller scales without reasoning capabilities to larger, reasoning-enabled models. All experimental results show \textsc{BadTV}'s high attack success rate (ASR). 

In addition to evaluating standard backdoor defenses designed for conventional classifiers, we assess two state-of-the-art defenses specifically proposed against TA backdoors~\cite{arora-etal-2024-heres,dam}. Unfortunately, these defenses fail to detect our backdoored TVs, primarily due to severe accuracy degradation on clean samples.

\textbf{Contribution} Overall, our  contributions are three-fold:
(1) We uncover a novel backdoor threat in task arithmetic (TA) by introducing \textsc{BadTV}, which implants backdoors into task vectors.
(2) \textsc{BadTV} attains strong attack effectiveness through composite backdoors and strategic omission of unnecessary clean tasks.
(3) Extensive experiments validate the robustness and effectiveness of \textsc{BadTV} across various challenging scenarios, some surpassing traditional contexts. Existing defenses fail against our method, highlighting the pressing need for stronger protective measures.

\section{Related Work}\label{sec: related work}


\textbf{Model Merging} A related method to TA is Model Merging (MM)~\cite{ortiz-jimenez2023task, tang2023concrete, wortsman2022modelsoup, ties-merging, yang2024representation, AdaMerging2024iclr, badmerging}, which primarily addresses task learning. In contrast, TA supports learning, forgetting, and analogy operations, making it particularly suitable for deploying TVs akin to mobile applications or browser extensions. TVaaS providers could evolve into app-store-like platforms for ML models, highlighting significant potential for future TV-based services. Consequently, given this promising outlook, our work focuses exclusively on TA.

\textbf{Model Merging Backdoors}
While most existing studies emphasize safety alignment~\cite{bhardwaj2024languagemodelshomersimpson, hammoud2024modelmergingsafetyalignment, hazra2024safetyarithmeticframeworktesttime, kim2024decoupling, li2024safetylayersalignedlarge, yi2024safetyrealignmentframeworksubspaceoriented}, the works most closely related to ours are \textsc{BadMerging}~\cite{badmerging} and \textsc{MergeBackdoor}~\cite{mergebackdoor}. In particular, \textsc{BadMerging} proposes a robust backdoor attack on merged models under unknown scaling factors, while \textsc{MergeBackdoor} improves stealth by evading detection on poisoned upstream models. In contrast to these methods, which focus on \textit{addition-based merging}, \textsc{BadTV} uniquely targets TA, ensuring that \textit{backdoors requires persistence across both addition and subtraction}. We provide a detailed comparison between \textsc{BadTV}, \textsc{BadMerging}, and \textsc{MergeBackdoor} in Appendices~\ref{sec:why not BadMerging} and \ref{sec:why not MergeBackdoor}.

Recent works further investigate malicious behaviors in model merging from different perspectives: \textsc{LoBAM}~\cite{loBAM} implants backdoors by taking advantage of LoRA modules, BV/SBV~\cite{bv} analyze vulnerability propagation and suppression across merged components, and \textsc{Merge-Hijacking}~\cite{merge-hijacking} studies adversarial task takeover through malicious model fusion.

\textbf{Weight Poisoning Attacks}
Classical weight poisoning attacks alter a victim model’s weights or architecture to embed a backdoor before deployment~\cite{AB, DF, HB, AB2, WP, conf/ndss/LiuMALZW018}. While \textsc{BadTV} similarly modifies victim models through backdoored TVs (BTVs), it significantly differs by never directly manipulating victim model parameters. Instead, attackers publish BTVs via TVaaS platforms, and users voluntarily incorporate them into their models using TA updates. Thus, the backdoor injection occurs externally, independent of the victim's final model. This novel threat model removes the requirement of privileged access, expanding the attack surface to any downstream TV consumer.

\textbf{Federated Learning Backdoors}
\textsc{BadTV} also shares parallels with federated learning (FL) backdoor attacks~\cite{pmlr-v108-bagdasaryan20a,10.5555/3495724.3497072}, wherein attackers contribute malicious updates to aggregated models. However, unlike FL, \textsc{BadTV} operates without knowledge of the victim’s tasks or gradients.

\textbf{Linear Mode Connectivity}
Modern neural networks exhibit linear mode connectivity (LMC)~\cite{adilova2024layerwise, Frankle20, ren2025revisiting}, with cross-task linearity (CTL) extending LMC across tasks~\cite{Zhou24CTL}. These properties reframe MM and TA as traversals along approximately linear, low-loss manifolds. Specifically, TA begins from a shared pre-trained base, and CTL suggests adding a TV produces task features nearly as convex combinations of source and target tasks, enabling rapid addition or removal of capabilities with minimal interference.
\section{Preliminaries}\label{sec:prelim}

\begin{figure*}[hbt]
    \centering
    \includegraphics[width=0.85\textwidth]{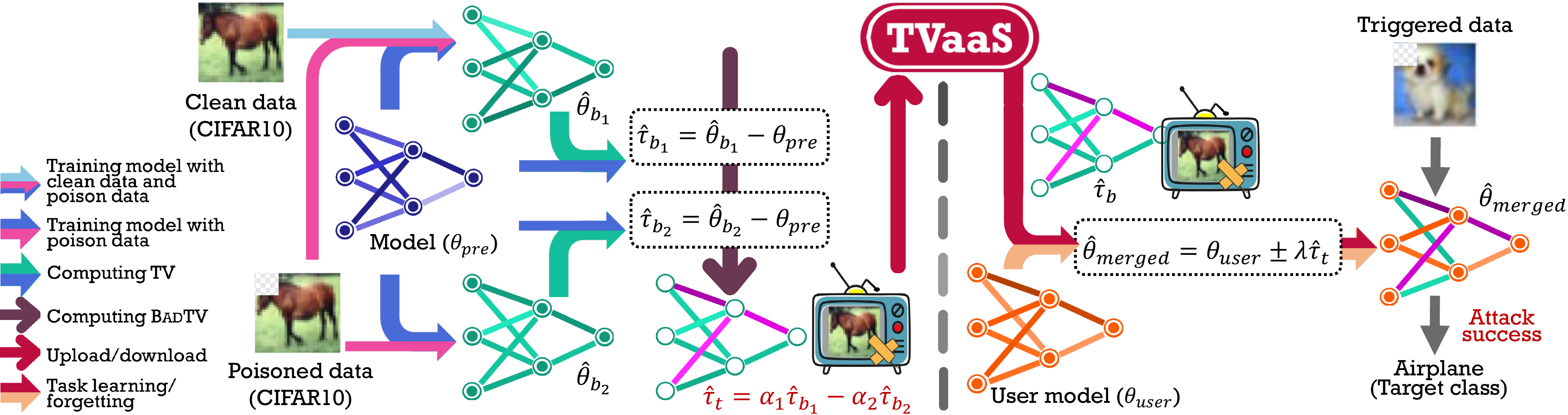}
    \caption{The workflow of \textsc{BadTV}.}
    \label{fig:structure}
    \Description{}
\end{figure*}

We primarily demonstrate backdoors for TA using the \texttt{CLIP} model. Hence, we introduce the fundamentals of \texttt{CLIP}, task arithmetic, and backdoors in this section.

\textbf{CLIP-based Classifiers} For \texttt{CLIP}~\cite{pmlr-v139-jia21b,pmlr-v139-radford21a}, the pre-trained model $M$ consists of a visual encoder $V$ and a text encoder $T$, i.e., $M = \{V, T\}$. Unlike traditional classifiers (e.g., ResNet, VGG), \texttt{CLIP} can classify images in a zero-shot manner using textual descriptions of class labels (e.g., “dog”). Specifically, given a labeled dataset $\{(x, y)\}$, $M$ computes similarity scores for an input image $x$ across $k$ classes:
\begin{align}
M(x, C) = [\langle V(x), T(c_1) \rangle, \ldots, \langle V(x), T(c_k) \rangle]^T,
\end{align}
where $C = [c_1, \ldots, c_k]$ are textual descriptions of classes, and $\langle V(x), T(c_i) \rangle$ is the similarity between the embeddings of $x$ and class $c_i$. For a given task with class labels $C$, the task-specific classifier is obtained by minimizing the cross-entropy loss $L_{\text{CE}}(M(x, C), y)$ with ground-truth label $y$. Freezing $T$ and fine-tuning $V$ yields the best performance.

\textbf{Task Arithmetic}
TA~\cite{ilharco2023iclr,aTLAS} addresses the limitations of conventional fine-tuning by working with task vectors (TVs). Let $M_{\theta}$ be a model with weights $\theta$, $\theta_{\text{pre}}$ the weights of a publicly available pre-trained model, and $\theta_{t}$ the weights after fine-tuning on task $t$ with dataset $D_{t}$ and loss $L_{t}$. The TV $\tau_{t}$ is defined as $\tau_{t} := \theta_{t} - \theta_{\text{pre}}$. Any model with the same architecture as $M_{\theta_{\text{pre}}}$ can use $\tau_{t}$. In particular, by adding $\tau_{t}$ to $\theta_{\text{pre}}$ with a scaling factor $\lambda$, we perform \textit{task learning}, $\theta_{\text{merged}} := \theta_{\text{pre}} + \lambda \tau_{t}$. Adjusting $\lambda$ controls how strongly the merged model $\theta_{\text{merged}}$ focuses on task $t$. Conversely, \textit{task forgetting} is achieved by subtracting $\lambda \tau_{t}$. TA also supports \textit{task analogies} through mixed operations, $\theta_{\text{merged}} := \theta_{\text{pre}} + \lambda_1 \tau_{t_1} + (\lambda_2 \tau_{t_2} - \lambda_3 \tau_{t_3})$, enabling robust domain generalization and handling of scarce data.

\textbf{Backdoor Attacks}\label{subsec:Poisoning-based Attacks}
Backdoor attacks~\cite{blend, gu2019badnets, dynamicbackdoor, wanet, LC, narcissus} typically compromise ML models by embedding covert malicious behaviors via data poisoning. Backdoored models behave normally under standard conditions but exhibit malicious behavior when inputs contain a specific trigger. For example, in image classification tasks, a backdoored model misclassifies inputs with a pre-defined trigger into an attacker-specified target class. Consider an image $x$ and a trigger $g=\{mask,\delta\}$, where $mask$ is a binary mask indicating the trigger location, and $\delta$ is the trigger pattern. A triggered image $\hat{x}^g$ is formed as $\hat{x}^g = \delta \odot mask + (1-mask) \odot x$, with $\odot$ denoting pixel-wise multiplication. For simplicity, when the trigger $g$ is clear from context, we omit the identifier $g$ and denote the triggered input as $\hat{x}$. The goal of a backdoor attack is thus to produce a model that accurately classifies clean images $x$, yet consistently misclassifies triggered images $\hat{x}$ into the chosen target class $c$.

\section{Threat Model}\label{subsec:backdoor_threat}
\textbf{Attack Scenario}
We consider TVaaS providers such as Hugging Face~\cite{huggingface}, ModelScope~\cite{modelscope}, and Databricks Marketplace~\cite{databricksmarketplace}, which supply the necessary resources for various tasks, allowing users to integrate them via TA into a multi-tasking model without additional training. Now, TVaaS platforms typically do not directly provide TVs; instead, they offer pre-trained models $\theta_{\text{pre}}$ and task-specific fine-tuned models $\theta_t$. Users derive task vectors (TVs) as $\tau_t=\theta_t-\theta_{\text{pre}}$, assuming $\theta_{\text{pre}}$ and $\theta_t$ share the same model architecture.

Consider a user with a victim model $M_{\theta_{\text{victim}}}$, where $\theta_{\text{victim}}:=\theta_{\text{pre}}\oplus \lambda_1\theta_1\cdots \oplus \lambda_U\theta_U$, with $\oplus\in\{+,-\}$, a positive integer $U$, a pre-trained model $\theta_{\text{pre}}$ (e.g., \texttt{CLIP}), TVs $\theta_1, \cdots, \theta_U$ corresponding to tasks $t_1, \cdots, t_U$, and respective scaling factors $\lambda_1, \cdots, \lambda_U$. To adapt $M_{\theta_{\text{victim}}}$ for a new task $t$ distinct from existing tasks $t_1,\cdots,t_U$, the user downloads $\theta_t$ (and $\theta_{\text{pre}}$ if needed), computes $\tau_t$, and then obtains a merged model $M_{\theta_{\text{merged}}}:=M_{\theta_{\text{victim}}\oplus\lambda\tau_t}$, with a user-specified scaling factor $\lambda$ to control adaptation strength.

In our scenario, instead of providing the genuine TV $\tau_t$, a TVaaS provider delivers a backdoored task vector (BTV) $\hat{\tau}_t$, causing the resulting merged model $\hat{M}_{\theta_{\text{merged}}}:=M_{\theta_{\text{victim}}\oplus\lambda\hat{\tau}_t}$ to become backdoored. Such a compromised $\hat{\tau}_t$ can originate from either the provider itself or a third-party attacker exploiting weak security measures for user-submitted TVs on platforms like Hugging Face~\cite{huggingface}.

\textsc{BadTV} is explicitly designed to expose this new vulnerability associated with TA. Consequently, configurations or architectures that do not employ TA are beyond the scope of our study.

\textbf{Attacker's Goal}
The attacker aims to construct a backdoored task vector (BTV) $\hat{\tau}_t$ that associates a predefined trigger with a target class in a attacker-specified task $t$. Given a backdoored fine-tuned (BFT) model $\theta_{\text{BFT}}:=\theta_{\text{pre}}+\hat{\tau}_t$ derived from a pre-trained model $\theta_{\text{pre}}$, the attacker uploads $\theta_{\text{BFT}}$ to a TVaaS provider only if it achieves sufficiently high accuracy on task $t$; otherwise, the attacker discards this model. Ultimately, the attacker intends that when users download and integrate $\hat{\tau}_t$ into their victim models by computing $\theta_{\text{victim}}\oplus \lambda\hat{\tau}_t$, where $\oplus \in \{+, -\}$, the resulting merged model $\hat{M}_{\theta_{\text{merged}}}$ becomes backdoored. Specifically, while $\hat{M}_{\theta_{\text{merged}}}$ maintains normal accuracy on clean inputs, it consistently misclassifies triggered inputs as the attacker-specified target class. The core functionality of \textsc{BadTV} aligns with traditional backdoor attacks, primarily targeting classification tasks.

\textbf{Attacker’s Knowledge and Capability} We assume the attacker has a dataset $\hat{D}_t$ created by embedding backdoor triggers into a clean dataset $D_t$ using established methods~\cite{blend, gu2019badnets, dynamicbackdoor, wanet, LC, narcissus}. Given a \texttt{CLIP}-based model, the attacker fine-tunes the pre-trained model $M_{\theta_{\text{pre}}}=\{V,T\}$ on $\hat{D}_t$, keeping $T$ frozen and updating only $V$ with cross-entropy loss $L_{\text{CE}}(M_{\theta_{\text{pre}}}(x, C), y)$. This produces a BFT model $\hat{M}_{\theta_t}$ with parameters $\hat{\theta}_t$, from which the attacker derives the final BTV as $\hat{\tau}_t=\hat{\theta}_t-\theta_{\text{pre}}$.

During this process, the attacker has white-box access to the publicly available pre-trained model $\theta_{\text{pre}}$ (e.g., \texttt{CLIP}) and its architecture. However, the attacker does not have access to the user's specific configurations, including $\theta_{\text{victim}}$, the TVs $\theta_1,\cdots,\theta_U$, scaling factors $\lambda_1,\cdots,\lambda_U$, or the scaling factor $\lambda$, as these are chosen privately by the user after downloading the BTV.

\textbf{Presentation Conventions} We assume TVaaS providers directly offer $\tau_t$ for download, although in practice users derive $\tau_t$ from downloaded models $\theta_{\text{pre}}$ and $\theta_t$. Next, we define two scenarios:

\begin{itemize}
    \item \textbf{Simplified Setting}: $\theta_{\text{merged}}=\theta_{\text{victim}}\oplus \lambda\hat{\tau}_t$ with $\theta_{\text{victim}}=\theta_{\text{pre}}$. This setting is purely illustrative and is not adopted in our experiments.

    \item \textbf{Realistic Setting}: $\theta_{\text{merged}}=\theta_{\text{victim}}\oplus \lambda\hat{\tau}_t$ with $\theta_{\text{victim}}=\theta_{\text{pre}}\oplus \lambda_1\theta_1\cdots \oplus \lambda_U\theta_U$.
\end{itemize}
To ease presentation, we initially describe \textsc{BadTV} under the simplified setting in \S\ref{sec:Challenges and Key Insight} and \S\ref{sec:Proposed Method}. However, all experimental evaluations are conducted under the realistic setting.

\section{Challenges and Key Insights}\label{sec:Challenges and Key Insight}

\textbf{Trivial Backdoor Design}\label{sec:Trivial Backdoor Design} A straightforward approach for an attacker is to craft two separate BTVs, $\hat{\tau}_t^+$ explicitly for addition and $\hat{\tau}_t^-$ explicitly for subtraction. These BTVs can be easily generated by applying traditional backdoor techniques (e.g., BadNets) to fine-tune $\theta_{\text{pre}}$ on $\hat{D}_t$. However, since a typical TV should support both task learning and forgetting, this approach deviates from standard practice and would likely arouse suspicion.

\begin{figure}[t]
    \centering
    \subfigure[Trajectory of TV addition.]{
    \includegraphics[width=0.23\textwidth]{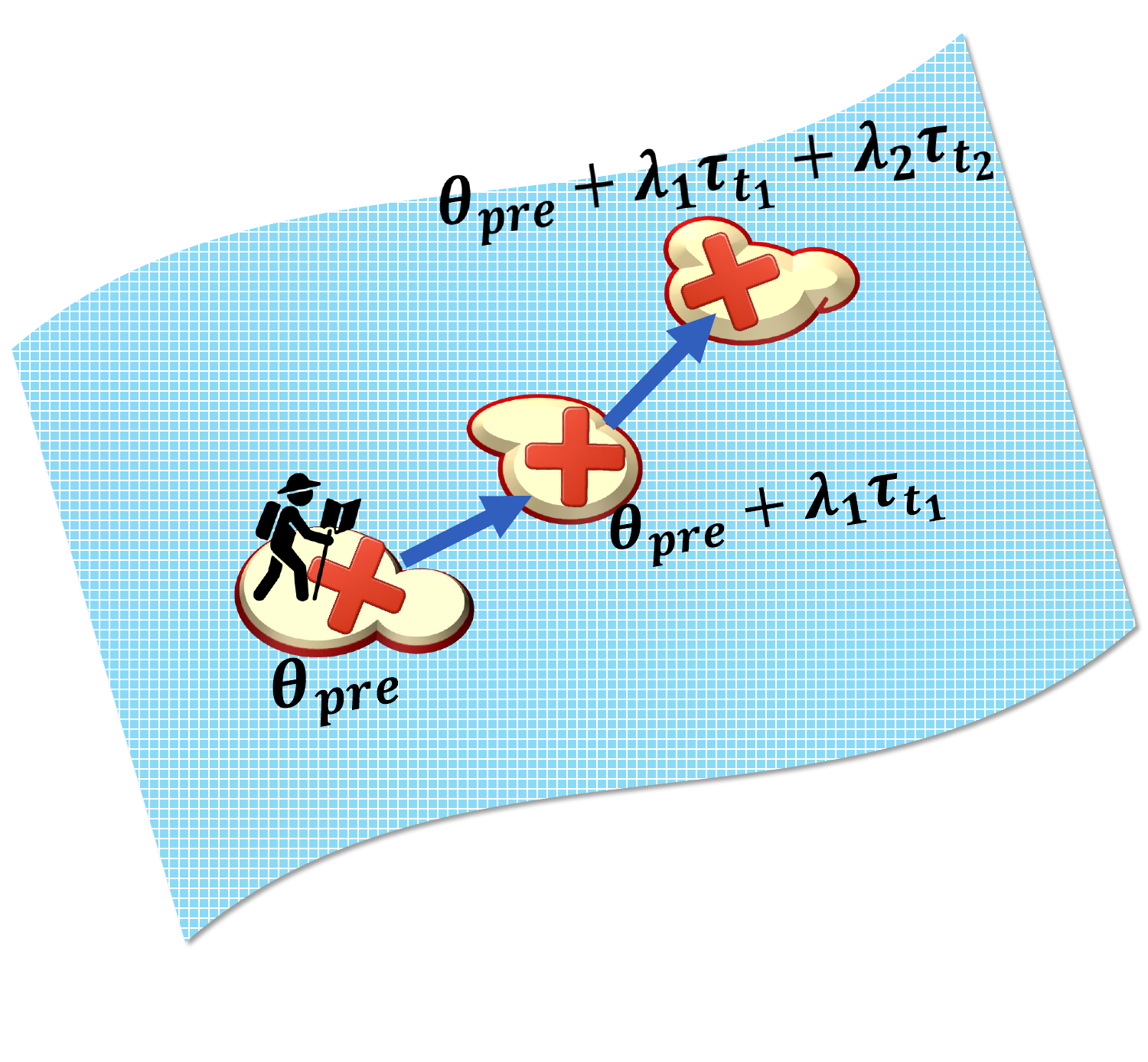}}
    \subfigure[Two different backdoor regions.]{
    \includegraphics[width=0.23\textwidth]{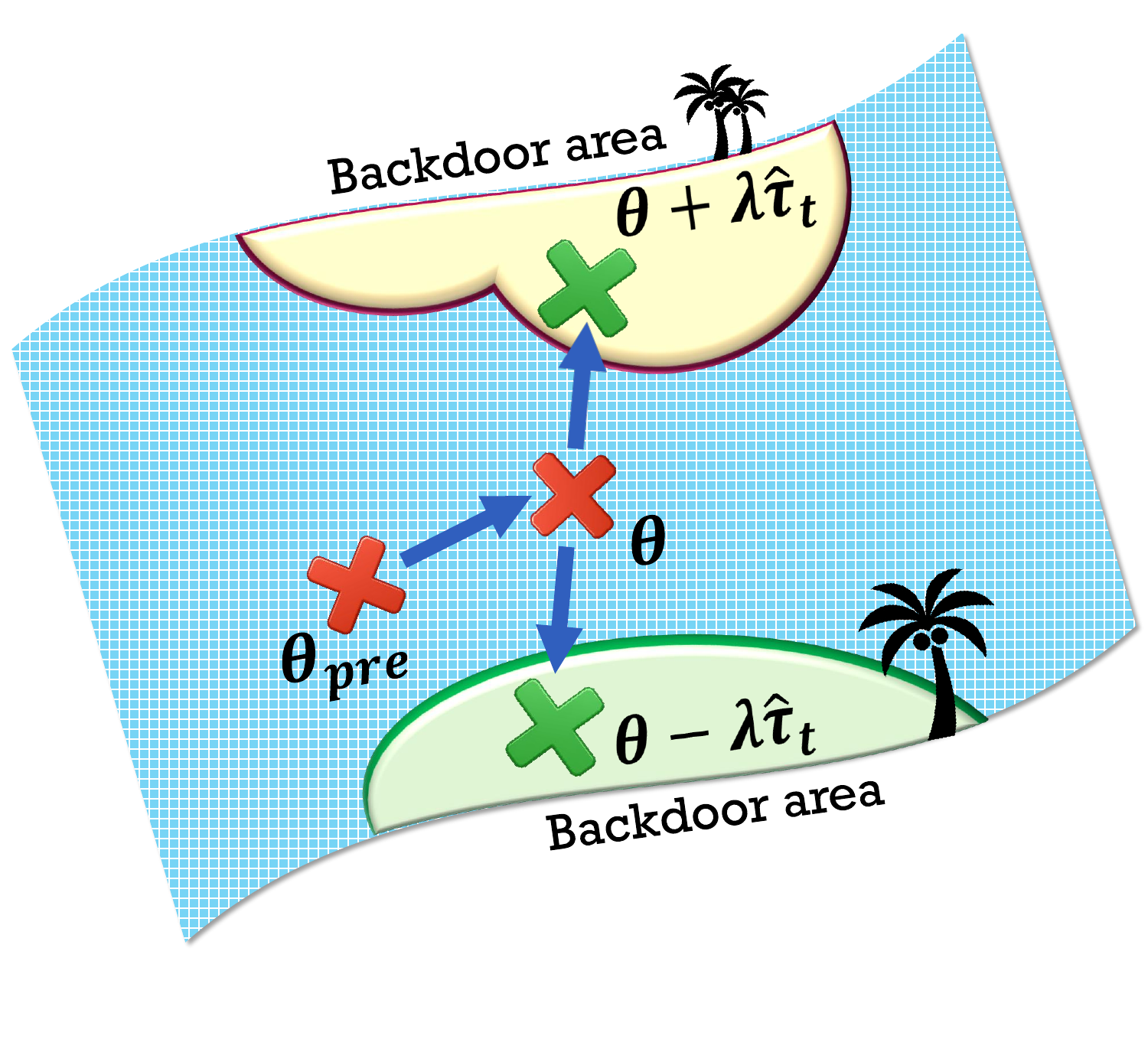}}
    \caption{The design challenge of BTV in weight space.}
    \Description{}
    \label{fig:abstract_idea}
\end{figure}

\textbf{Challenges}\label{sec:Challenges}
The key challenge is ensuring the backdoor remains effective across all potential applications of a single BTV, including task learning, forgetting, and analogies; i.e., the backdoor must succeed irrespective of how the user integrates the BTV.

Formally, an attacker has a dataset $\hat{D}_t = \{(x_1, y_1), \dots, (x_{\mu}, y_{\mu})\} \cup \{(\hat{x}_1, \hat{y}), \dots, (\hat{x}_{\nu}, \hat{y})\}$, comprising $\mu$ clean samples and $\nu$ poisoned samples. Here, $x_i$ denotes a clean input with ground-truth label $y_i$, while $\hat{x}_i$ denotes a poisoned input assigned the attacker-specified target class $\hat{y}$. The BTV for task $t$ is defined as $\hat{\tau}_t = \hat{\theta}_t - \theta_{\text{pre}}$, where $\hat{\theta}_t$ is the parameters of the backdoored model $\hat{M}_{\theta_t}$ fine-tuned on $\hat{D}_t$. The resulting merged model’s weights are given by
\begin{align}\label{eq:merge}
    \hat{\theta}_{\text{merged}} = \theta_{\text{pre}} \oplus \lambda \hat{\tau}_t \quad \text{with} \quad \oplus \in \{+,-\},
\end{align}
where $\lambda$ is chosen by the user. The attacker's goal is to find $\hat{\tau}_t$ by solving the following optimization:

\begin{scriptsize}
\begin{equation}
\begin{aligned}\label{eq:btvequation}
\min_{\theta} \quad & 
  \left(
    \sum_{i=1}^{\mu} L_{\text{CE}}(\hat{M}_{\theta_{\text{merged}}}(x_{i}, y_{i}), y_{i}) +
    \sum_{j=1}^{\nu} L_{\text{CE}}(\hat{M}_{\theta_{\text{merged}}}(\hat{x}_{i}, \hat{y}), \hat{y})
  \right) \\
\text{s.t.} \quad & \hat{\theta}_{\text{merged}} = \theta_{\text{pre}} + \lambda \hat{\tau}_t\mbox{ or } \hat{\theta}_{\text{merged}} = \theta_{\text{pre}} - \lambda \hat{\tau}_t.
\end{aligned}
\end{equation}
\end{scriptsize}

Clearly, implanting a single backdoor into the TV is insufficient to solve Equation~\eqref{eq:btvequation}, as the backdoor only survives the addition operation in TA, not the subtraction. Solving Equation~\eqref{eq:btvequation} is non-trivial due to the dual constraints, where two backdoor regions exist in opposite positions within the weight space (see Figure~\ref{fig:abstract_idea}).

\begin{figure}[!t]
    \centering
    \subfigure[Equation~\eqref{eq:btvequation}\label{fig:sub1}]{\label{fig:loss-a}
        \includegraphics[width=0.29\linewidth]{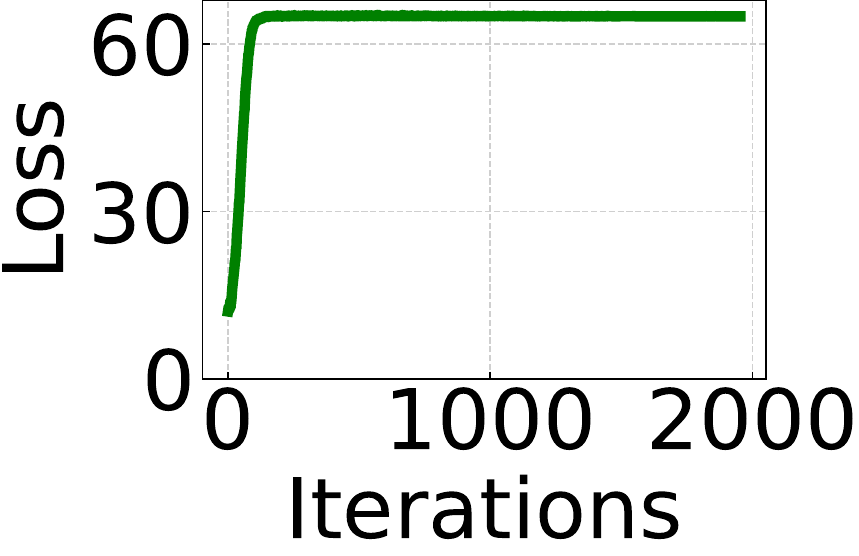}
    }
    \hfill
    \subfigure[Averaging 2 TVs\label{fig:sub2}]{\label{fig:loss-b}
        \includegraphics[width=0.29\linewidth]{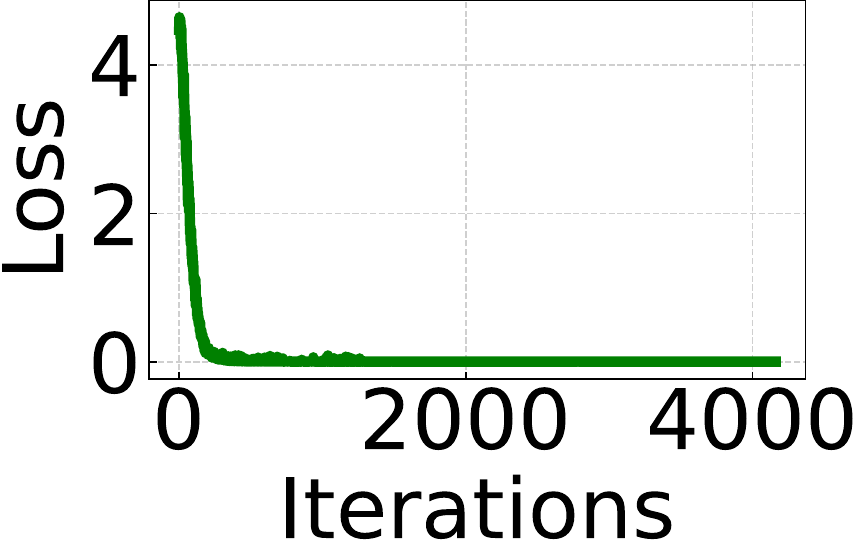}
    }
    \hfill
    \subfigure[Alternating Eqs.~\eqref{eq:firsthalf}/\ref{eq:secondhalf}\label{fig:sub3}]{\label{fig:loss-c}
        \includegraphics[width=0.29\linewidth]{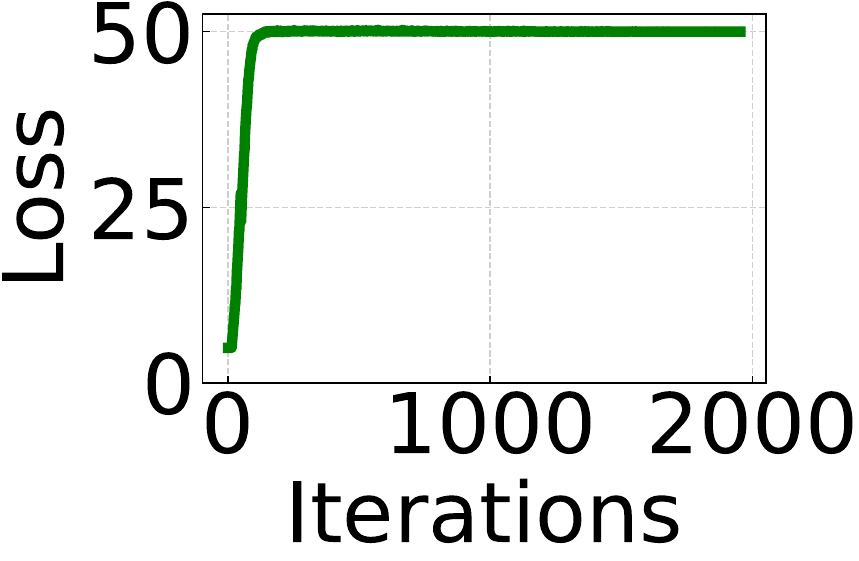}
    }
    \caption{Loss curves of the failed attempts.}
    \label{fig:loss}
    \Description{}
\end{figure}

One alternative is to solve the constrained optimization for the addition operation in Equation~\eqref{eq:firsthalf}:
\begin{scriptsize}
\begin{equation}
\begin{aligned}\label{eq:firsthalf}
\min_{\theta} \quad & 
  \left(
    \sum_{i=1}^{\mu} L_{\text{CE}}(\hat{M}_{\theta_{\text{merged}}}(x_{i}, y_{i}), y_{i}) +
    \sum_{j=1}^{\nu} L_{\text{CE}}(\hat{M}_{\theta_{\text{merged}}}(\hat{x}_{i}, \hat{y}), \hat{y})
  \right) \\
\text{s.t.} \quad & \hat{\theta}_{\text{merged}} = \theta_{\text{pre}} + \lambda \hat{\tau}_t.
\end{aligned}
\end{equation}
\end{scriptsize}to obtain a TV for addition and solve the optimization for the subtraction operation in Equation~\eqref{eq:secondhalf}:

\begin{scriptsize}
\begin{equation}
\begin{aligned}\label{eq:secondhalf}
\min_{\theta} \quad & 
  \left(
    \sum_{i=1}^{\mu} L_{\text{CE}}(\hat{M}_{\theta_{\text{merged}}}(x_{i}, y_{i}), y_{i}) +
    \sum_{j=1}^{\nu} L_{\text{CE}}(\hat{M}_{\theta_{\text{merged}}}(\hat{x}_{i}, \hat{y}), \hat{y})
  \right) \\
\text{s.t.} \quad 
                  & \hat{\theta}_{\text{merged}} = \theta_{\text{pre}} - \lambda \hat{\tau}_t.
\end{aligned}
\end{equation}
\end{scriptsize}to obtain a TV for subtraction. Averaging the two TVs provides the BTV. Another approach is alternating optimization between Equations~\eqref{eq:firsthalf} and~\eqref{eq:secondhalf}, iteratively solving one with the solution of the other as the initial point until convergence. However, our experimental results in Figures~\ref{fig:loss-a} and \ref{fig:loss-c}, where we train the Blend~\cite{blend} on CIFAR10~\cite{cifar} for 5 epochs, show that the training loss cannot be reduced with the direct optimization and alternating optimization. Optimizing Equations~\eqref{eq:firsthalf}~and~\eqref{eq:secondhalf} respectively is easy to reduce the loss, as shown in Figure~\ref{fig:loss-b}. However, the BTV derived by averaging their solutions does not have backdoor infectiousness. 

\textbf{Key Insights}
The central insight of \textsc{BadTV} is the design of a composite backdoor specifically tailored for the BTV, comprising two independent backdoors arranged to ensure that at least one remains effective under both addition and subtraction operations. This disentanglement circumvents the complexities of solving an optimization problem constrained by dual requirements, as described in Equation~\eqref{eq:btvequation}. To satisfy the first constraint (addition), we train backdoored task-specific weights $\hat{\theta}_{b_1}$ on dataset $\hat{D}_t$ using backdoor configuration $b_1$. Conversely, to fulfill the second constraint (subtraction), we independently train another set of weights $\hat{\theta}_{b_2}$ on $\hat{D}_t$ using a different backdoor configuration $b_2$, subsequently taking their negation as $-\hat{\theta}_{b_2}$. This choice aligns with Equation~\eqref{eq:merge}, ensuring proper functionality under subtraction: $\hat{\theta}_{\text{merged}} = \theta_{\text{pre}} - \lambda(-\hat{\tau}_{b_2}) = \theta_{\text{pre}} + \lambda\hat{\tau}_{b_2}$. Consequently, the composite backdoored model weights, defined as $\hat{\theta}_{b} = \hat{\theta}_{b_1} - \hat{\theta}_{b_2}$, yield a BTV robust against both addition and subtraction.

\section{Proposed Method}\label{sec:Proposed Method}

\subsection{\textsc{BadTV}}\label{sec:BadTV}
Inspired by a recent theoretical finding on TA~\cite{li2025when} (see \S\ref{sec:robust-feasibility-composite-btv}), we now introduce the theory-induced design of \textsc{BadTV}. First, for the attacker's target task, given a backdoor method (e.g., BadNets), we train the backdoor model weights $\hat{\theta}_{b_1}$ on the poisoned dataset $\hat{D}_t = \{(x_1, y_1), \dots, (x_{\mu}, y_{\mu})\} \cup \{(\hat{x}_1, \hat{y}), \dots, (\hat{x}_{\nu}, \hat{y})\}$, composed of $\mu$ clean samples and $\nu$ poisoned samples, based on the backdoor configuration $b_1$. Note that a backdoor configuration includes trigger pattern, trigger location, target class, etc. Next, for the same task, we train the backdoor model weights $\hat{\theta}_{b_2}$ on the poisoned subset $\{(\hat{x}_1, \hat{y}), \dots, (\hat{x}_{\nu}, \hat{y})\}$ \textit{alone} based on a different backdoor configuration $b_2$ (see \S\ref{sec:why the setting of different target classes}). In such an asymmetric design, because $\hat{\theta}_{b_2}$ is learned exclusively from malicious samples, its updates $\hat{\tau}_{b_2} := \hat{\theta}_{b_2} - \theta_{\text{pre}}$ do not negate the updates $\hat{\tau}_{b_1} := \hat{\theta}_{b_1} - \theta_{\text{pre}}$. We then construct the BTV $\hat{\tau}_t$ as:
\begin{align}\label{eq:backdoor_TV}
    \hat{\tau}_{t} = \alpha_1 \cdot \hat{\tau}_{b_1} \;-\; \alpha_2 \cdot \hat{\tau}_{b_2},
\end{align}
where $\alpha_{1}$ and $\alpha_{2}$ control the relative strengths of backdoors $b_1$ and $b_2$. We slightly abuse the notation; here, $\hat{\tau}_{t}$ (an attacker-specified task $t$ as a subscript) is a BTV for task $t$ crafted by \textsc{BadTV} (through Equation~\eqref{eq:backdoor_TV}), while $\hat{\tau}_{b_1}$ (a backdoor configuration as a subscript) denotes a BTV for task $t$ directly derived by fine-tuning $\theta_{\text{pre}}$ on $\hat{D}_t$ with an attacker-specified backdoor method to obtain $\hat{\theta}_{b_1}$, and then calculating $\hat{\theta}_{b_1}-\theta_{\text{pre}}$. The exact procedure by which the attacker obtains $\hat{\theta}_{b_1}$ is irrelevant to \textsc{BadTV}. When the underlying task $t$ is clear from context, we omit it from the subscript notation. The optimization of $\alpha_{1}$ and $\alpha_{2}$ is discussed in \S\ref{sec:robust-feasibility-composite-btv}. Figure~\ref{fig:structure} illustrates the \textsc{BadTV} workflow.

\begin{figure*}[t]
    \centering
    \subfigure[Traditional BD (Addition)]{ \label{fig:Sub_featurea}
    \includegraphics[width=0.19\textwidth]{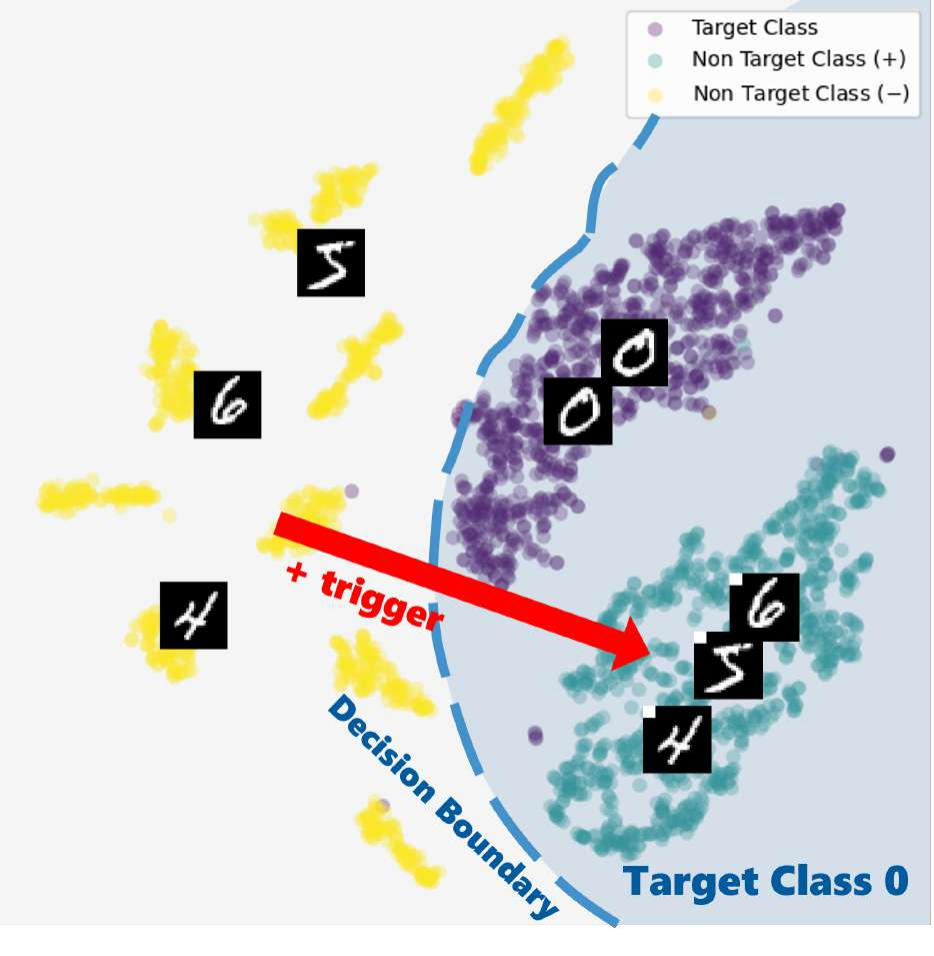}}
    \subfigure[\textsc{BadTV} (Addition)]{ \label{fig:Sub_featureb}
    \includegraphics[width=0.2\textwidth]{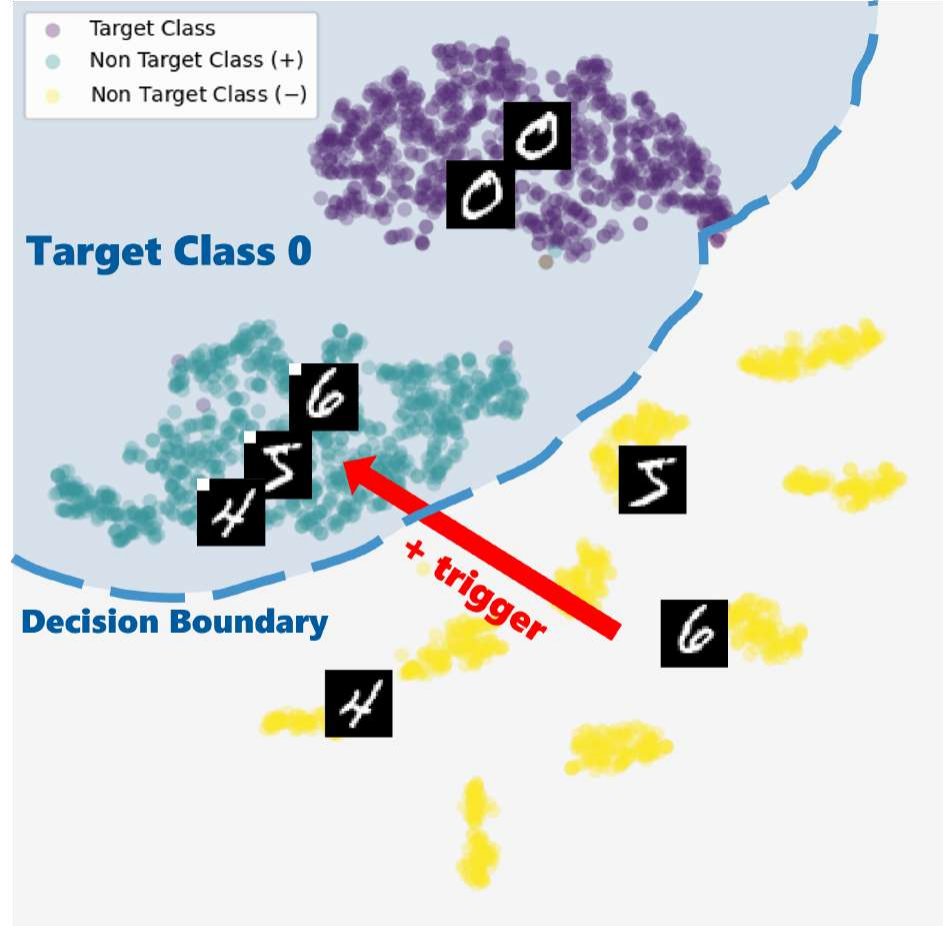}}
    \subfigure[Traditional BD (Subtraction)]{ \label{fig:Sub_featurec}
    \includegraphics[width=0.2\textwidth]{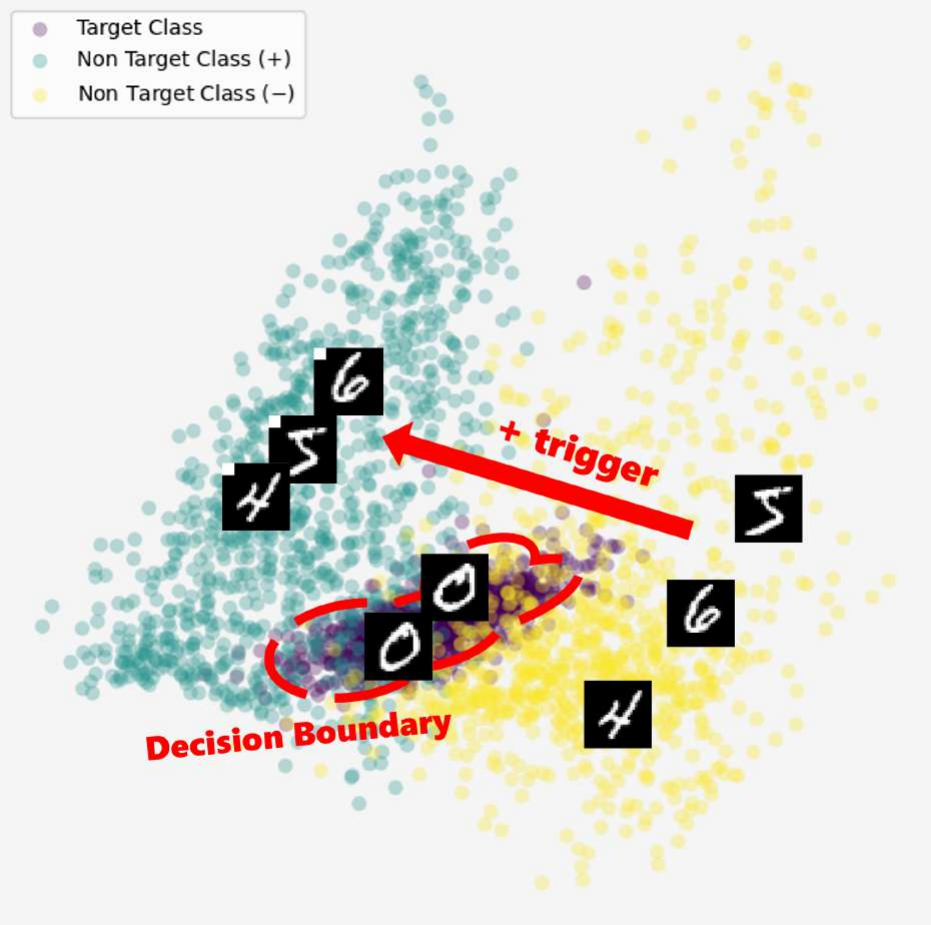}}
    \subfigure[\textsc{BadTV} (Subtraction)]{ \label{fig:Sub_featured}
    \includegraphics[width=0.2\textwidth]{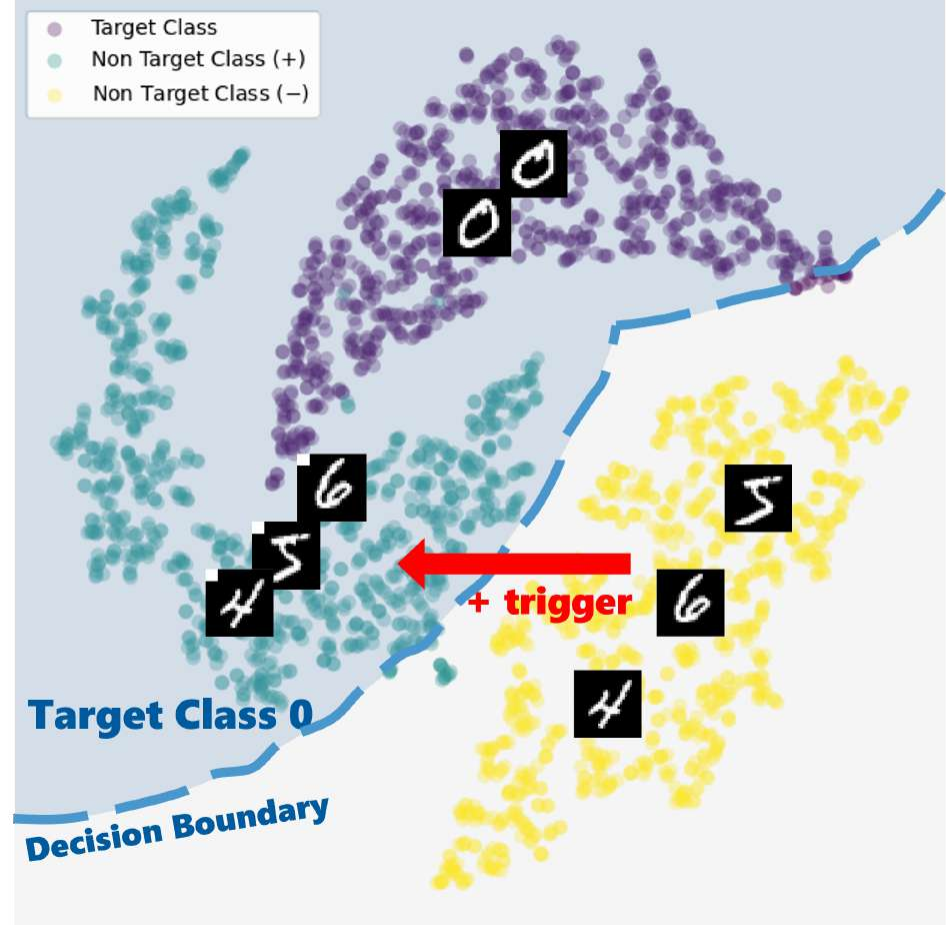}}
    \caption{Visualization of traditional backdoor attack (BD) and \textsc{BadTV}. Figure~\ref{fig:Sub_featurea},~\ref{fig:Sub_featureb} and~\ref{fig:Sub_featured} show that images are classified as the target class when adding trigger. Figure~\ref{fig:Sub_featurec} shows images with a trigger that cannot be classified as the target class.}
    \label{fig:Sub_feature}
    \Description{Comparisons between traditional backdoor attacks and BADTV.}
\end{figure*}

More specifically, when the user applies addition, $\hat{\tau}_{b_1}$ introduces the backdoor to the merged model, while $\hat{\tau}_{b_2}$ does not influence clean accuracy (it was trained only on malicious samples). Conversely, subtraction activates $\hat{\tau}_{b_2}$ as the backdoor in the merged model, while the normal task functionality in $\hat{\tau}_{b_1}$ works for the forgetting part. The visual outcomes in Figure~\ref{fig:Sub_feature} confirm that a composite backdoor in the BTV succeeds under both addition and subtraction (Figures~\ref{fig:Sub_featureb} and \ref{fig:Sub_featured}). However, a single backdoor fails in one of the two scenarios (Figures~\ref{fig:Sub_featurea} and \ref{fig:Sub_featurec}).

\subsection{Analysis}\label{sec:Analysis}

\subsubsection{Robust Feasibility of Composite Backdoor Task Vectors}
\label{sec:robust-feasibility-composite-btv}

A fundamental challenge in \textsc{BadTV} is that the user privately chooses the scaling factor $\lambda$ and may apply either addition or subtraction, while the attacker has no control over this choice. Consequently, an BTV must remain effective \emph{robustly} across a range of admissible user-selected $\lambda$, rather than being tuned to a single fixed scaling. In \textsc{BadTV}, the attacker constructs a composite BTV via Equation~\eqref{eq:backdoor_TV}. Poorly chosen weights may render the backdoor ineffective under either task learning or task forgetting, or significantly degrade the clean model’s performance, which necessitates an explicit calibration of $(\alpha_1,\alpha_2)$ against unknown $\lambda$.

\paragraph{Robust Optimization of $(\alpha_1,\alpha_2)$.}
For a given $\lambda$, let $A_\lambda^{+}$ ($A_\lambda^{-}$) denote the attack success rate (ASR) under addition (subtraction), and define $\Delta C_\lambda^+:=C_{\text{clean}}-C_{\text{clean}}^+$ ($\Delta C_\lambda^-:=C_{\text{clean}}-C_{\text{clean}}^-$) as the clean accuracy drop (CA-drop). Here, $C_{\text{clean}}$ is the CA before applying TA, while $C_{\text{clean}}^+$ ($C_{\text{clean}}^-$) is the top-1 accuracy after addition (subtraction). We fix a lower bound $A_0\in(0,1)$ that the attacker aims to guarantee for the ASR across all admissible $\lambda$, and formulate the following robust optimization (RO) problem to determine $(\alpha_1,\alpha_2)$:
\begin{scriptsize}
\begin{equation}
\begin{aligned}
    & \argminB_{\alpha_1,\alpha_2 }\; \max_{\lambda \in \Lambda} \max\left\{\Delta C^+_{\lambda}, \Delta C^-_{\lambda}\right\} \\
    & \quad \mathrm{s.t.} \;\;  \Lambda := \left\{ \lambda \mid \min\{A^+_{\lambda}, A^-_{\lambda} \} \geq A_0 , \lambda \in [\lambda_{\min}, \lambda_{\max}] \right\}
    \ \mbox{and } \alpha_1,\alpha_2 \in [\alpha_{\min}, \alpha_{\max}].
\end{aligned}
\label{eq:RO}
\end{equation}
\end{scriptsize}Equation~\eqref{eq:RO} enforces $\text{ASR}\ge A_0$ for all user-selected $\lambda\in\Lambda$ while minimizing the worst-case CA-drop. We solve Equation~\eqref{eq:RO} using projected gradient ascent on a smooth surrogate of the inner $\max/\min$ operators (Huberized min-max), alternating with a golden-section search over $\lambda$. Each iteration requires only two forward passes at $\lambda_{\min}$ and $\lambda_{\max}$, since $A_\lambda^+$, $A_\lambda^-$, $\Delta C_\lambda^+$, and $\Delta C_\lambda^-$ are empirically monotonic in $|\lambda|$.

To avoid early divergence, we initialize optimization at a norm-balanced point equalizing the magnitudes of the two directions:
\begin{align}\label{eq:norm}
\alpha_1^{(0)}=1,\quad
\alpha_2^{(0)}=\frac{\|\hat{\tau}_{b_1}\|_2}{\|\hat{\tau}_{b_2}\|_2}.
\end{align}
This offset compensates for the observed norm gap (approximately $10\times$) between $\hat{\tau}_{b_1}$ and $\hat{\tau}_{b_2}$ in the tested pre-trained models. In Equation~\eqref{eq:RO}, we set $\alpha_{\min}>0$ and $\alpha_{\max}=2$. We also choose $\lambda_{\min}=0.1$ and $\lambda_{\max}=1$ following prior TA practice~\cite{ilharco2023iclr}. Under these settings, Table~\ref{table: RO result} reports the optimal $(\alpha_1,\alpha_2)$ for different backdoor configurations, and Table~\ref{table: RO result comparison} shows that RO-optimized coefficients consistently outperform a user-specified choice $(\alpha_1,\alpha_2)=(1,1)$. Note that Table~\ref{table: RO result comparison} shows Narcissus benefits greatly from optimizing $(\alpha_1,\alpha_2)$; elsewhere, gains seem smaller mainly because empirical settings are already near-optimal (e.g., Blend). \textcolor{black}{To evaluate how sensitive $(\alpha_1, \alpha_2)$ is, Table~\ref{tab:combinations_of_twosub} shows that poorly chosen coefficients can severely degrade CA or ASR, indicating sensitivity to large deviations. However, performance is stable near the optimum: varying $(\alpha_1, \alpha_2)$ within ±0.3 around (1.76, 1.32) (BadNets) and (1.15, 1.34) (Blend), resulting in less than 3\% change in ASR/CA. Based on this observation, an attacker with limited resources can perform RO using only a limited number of optimization iterations starting from a norm-balanced initialization, thereby reducing computational overhead.} 


\begin{table}[h]
\centering
\caption{The optimal $\alpha_1$ and $\alpha_2$ from Equation~\eqref{eq:RO}}
\begin{adjustbox}{max width=.45\textwidth}
\begin{tabular}{|c|c|c|c|c|c|}
\hline
          & BadNets      & Blend       & WaNet       & Dynamic     & Narcissus    \\ \hline
$\alpha_1$ & 1.75968747  & 1.1488401   & 1.73239999  & 1.25046087  & 1.54440872   \\ \hline
$\alpha_2$ & 1.3228079   & 1.33605701  & 1.03127507  & 1.04587442  & 1.23842611   \\ \hline
\end{tabular}
\end{adjustbox}
\label{table: RO result}
\end{table}

\begin{table}[h]
\caption{CA/ASR under addition and subtraction with RO-induced $(\alpha_1, \alpha_2)$ and user-specified $(\alpha_1, \alpha_2)$}
\centering
\begin{adjustbox}{max width=.49\textwidth}
\begin{tabular}{|c|c|c|c|c|c|}
\hline
\makecell[c]{\textbf{Add-CA} ({$\uparrow$}) / \textbf{Add-ASR ({$\uparrow$})} \\\textbf{Sub-CA ({$\downarrow$})} / \textbf{Sub-ASR ({$\uparrow$})}} 
& BadNets & Blend & WaNet & Dynamic & Narcissus \\ \hline
$(\alpha_1, \alpha_2)$ in Table~\ref{table: RO result} &
\twolines{97.88 / 99.98}{5.70 / 100} &
\twolines{99.17 / 100}{4.01 / 100} &
\twolines{98.35 / 99.91}{5.70 / 99.5} &
\twolines{98.25 / 99.99}{5.70 / 100} &
\twolines{98.65 / 88.86}{5.07 / 100} \\ \hline
$(\alpha_1, \alpha_2)= (1, 1)$ &
\twolines{98.33 / 99.31}{5.70 / 99.99} &
\twolines{99.24 / 99.85}{4.96 / 100} &
\twolines{98.59 / 99.65}{5.72 / 99.97} &
\twolines{98.20 / 99.78}{5.71 / 99.91} &
\twolines{89.19 / 98.99}{26.27 / 100} \\ \hline
\end{tabular}
\end{adjustbox}
\label{table: RO result comparison}
\end{table}

\begin{table*}[ht!]
\centering
\caption{\textcolor{black}{CA/ASR under different $(\alpha_1,\alpha_2)$ settings (left: BadNets; right: Blend). Blue: Addition; green: Subtraction.}}
\label{tab:combinations_of_twosub}

\begin{minipage}{0.48\textwidth}
\centering
\footnotesize
\begin{adjustbox}{max width=\textwidth}
\begin{tabular}{c|cccc}
$\alpha_1 \backslash \alpha_2$ & 0.1 & 0.5 & 1.0 & 2.0 \\ \hline
0.1 & \twolines{52.26/1.05}{98.99/5.75} & \twolines{41.54/0.09}{98.18/6.33} & \twolines{30.97/0.00}{76.49/41.15} & \twolines{16.45/0.00}{5.70/100} \\\hline
0.5 & \twolines{96.13/10.45}{95.47/5.88} & \twolines{93.91/0.59}{85.03/16.86} & \twolines{86.75/0.12}{7.06/99.33} & \twolines{62.76/0.00}{5.70/100} \\\hline
1.0 & \twolines{98.99/99.95}{46.99/7.09} & \twolines{99.10/98.74}{24.47/55.83} & \twolines{98.33/99.31}{5.70/99.99} & \twolines{94.89/0.99}{5.70/100} \\\hline
2.0 & \twolines{96.84/100.00}{4.28/0.00} & \twolines{97.17/100.00}{5.02/0.04} & \twolines{97.00/100.00}{4.86/56.22} & \twolines{94.92/99.64}{5.70/100} \\
\end{tabular}
\end{adjustbox}
\end{minipage}
\hfill
\begin{minipage}{0.48\textwidth}
\centering
\footnotesize
\begin{adjustbox}{max width=\textwidth}
\begin{tabular}{c|cccc}
$\alpha_1 \backslash \alpha_2$ & 0.1 & 0.5 & 1.0 & 2.0 \\ \hline
0.1 & \twolines{55.84/0.94}{98.88/1.84} & \twolines{53.86/0.23}{98.73/69.46} & \twolines{46.50/0.00}{98.20/100} & \twolines{31.24/0.00}{84.18/100} \\\hline
0.5 & \twolines{97.73/98.86}{84.68/0.01} & \twolines{97.58/94.90}{79.67/87.51} & \twolines{97.10/89.56}{55.38/100} & \twolines{94.20/14.70}{5.71/100} \\\hline
1.0 & \twolines{99.31/100}{4.97/0.00} & \twolines{99.27/100}{5.09/78.54} & \twolines{99.24/99.85}{4.96/100} & \twolines{99.07/99.94}{5.70/100} \\\hline
2.0 & \twolines{96.05/99.87}{2.19/0.00} & \twolines{95.71/99.68}{2.25/0.00} & \twolines{95.20/99.03}{5.70/100} & \twolines{93.75/96.37}{2.45/4.92} \\
\end{tabular}
\end{adjustbox}
\end{minipage}

\end{table*}

\paragraph{Connection to Task Arithmetic Theory.}
Beyond robustly selecting $(\alpha_1,\alpha_2)$, the feasibility of this composite construction is consistent with the theoretical findings in~\cite{li2025when}, which show that a TA form $\theta_{\text{merged}}:=\theta_{\text{pre}}+\theta_1+\lambda\theta_2$ achieves optimal accuracy for \textit{irrelevant} binary tasks when task vectors exhibit minimal relevance. \textcolor{black}{In our setting, this formulation can be instantiated in \textsc{BadTV}. The} infected merged model can be written as $\hat{\theta}_{\text{merged}}
=\theta_{\text{pre}}+\lambda\hat{\tau}_t
=\theta_{\text{pre}}+\lambda\alpha_1\hat{\tau}_{b_1}-\lambda\alpha_2\hat{\tau}_{b_2}$, which applies to both task learning and task forgetting. \textcolor{black}{Based on~\cite{li2025when}, correlation is measured as the average Pearson correlation between the model outputs induced by two task vectors over their respective datasets. This metric reflects how similarly the two vectors influence the model: high correlation indicates that they steer the model in similar directions and thus interact more strongly under task arithmetic, whereas low correlation suggests more independent effects. Since~\cite{li2025when} shows that task arithmetic performs best when task vectors are relatively irrelevant (i.e., weakly correlated), we construct the two backdoor components to be low-correlated so that addition and subtraction can operate more independently.}

\textcolor{black}{We then } measure the correlation between $\hat{\tau}_{b_1}$ and $\hat{\tau}_{b_2}$ and obtain $\rho=-0.0065$, indicating minimal relevance. \textcolor{black}{This helps explain the observed robustness of \textsc{BadTV}, and why configurations that reduce correlation (e.g., using different target classes) tend to achieve better performance.}

To align with the binary-class assumption in~\cite{li2025when}, we further evaluate \textsc{BadTV} on two-class MNIST (MNIST-b). \textcolor{black}{Under addition,} \textsc{BadTV} preserves perfect clean accuracy \textcolor{black}{and achieves 98.97\% ASR. Under subtraction, the clean accuracy drops to the random-guessing level for a two-class task (50.74\%), while the ASR remains perfect.} These results empirically confirm that, despite minor formulation differences, the theoretical insights in~\cite{li2025when} effectively explain the success of \textsc{BadTV}.


\begin{proposition}[Robust feasibility of composite BTVs]
\label{prop:robust-feasibility}
Fix $A_0\in(0,1)$ and the ranges $\lambda\in[\lambda_{\min},\lambda_{\max}]$ and $\alpha_1,\alpha_2\in[\alpha_{\min},\alpha_{\max}]$ as in Equation~\eqref{eq:RO}. Let $\hat{\tau}_t=\alpha_1\hat{\tau}_{b_1}-\alpha_2\hat{\tau}_{b_2}$ be the composite BTV in Equation~\eqref{eq:backdoor_TV}, and define
\[
\Lambda \;:=\; \left\{ \lambda \;\middle|\; \min\{A_\lambda^{+},A_\lambda^{-}\}\ge A_0,\ \lambda\in[\lambda_{\min},\lambda_{\max}] \right\}.
\]
If $(\alpha_1,\alpha_2)$ is feasible for the robust optimization problem in Equation~\eqref{eq:RO}, then for every $\lambda\in\Lambda$,
\[
A_\lambda^{+}\ge A_0
\qquad\text{and}\qquad
A_\lambda^{-}\ge A_0.
\]
Moreover, the worst-case clean-accuracy drop over $\lambda\in\Lambda$ is exactly the objective value attained by $(\alpha_1,\alpha_2)$ in Equation~\eqref{eq:RO}.
\end{proposition}

Proposition~\ref{prop:robust-feasibility} formalizes the robust feasibility guarantee already encoded in Equation~\eqref{eq:RO}: a feasible solution $(\alpha_1,\alpha_2)$ ensures $\mathrm{ASR}\ge A_0$ simultaneously for both task learning (addition) and task forgetting (subtraction) across all admissible user-selected $\lambda$, while controlling the worst-case CA-drop.

Beyond this optimization-level guarantee, the existence of such feasible solutions can be interpreted through three complementary factors that are already observed. First, the correlation coefficient $\rho$ between $\hat{\tau}_{b_1}$ and $\hat{\tau}_{b_2}$ is close to zero, indicating \emph{minimal relevance} as measured following~\cite{li2025when}, which reduces destructive interaction when composing task vectors. Second, the two backdoor task vectors exhibit a substantial $\ell_2$-norm gap, as reflected in the norm-balanced initialization used in Equation~\eqref{eq:norm}, preventing symmetric cancellation under addition and subtraction. Third, since the attacker has no knowledge of the user-selected $\lambda$, robust feasibility must hold over a range of scaling factors rather than a single operating point, which is explicitly captured by the constraint set $\Lambda$ in Equation~\eqref{eq:RO}.

These three aspects do not constitute additional assumptions in the proposition; rather, they provide a mechanistic interpretation of why the RO in Equation~\eqref{eq:RO} admits feasible solutions that perform well in practice. Detailed mathematical justification and discussion of these factors, including the link between correlation and geometric alignment, are provided in Appendix~\ref{app:mechanistic-feasibility}.

\subsubsection{Why Different Backdoor Configurations in \textsc{BadTV}}\label{sec:why the setting of different target classes}

Similar to \S\ref{sec:robust-feasibility-composite-btv}, we construct two-class versions of SVHN~\cite{svhn} and CIFAR10~\cite{cifar}, denoted as SVHN-b and CIFAR10-b, respectively. Using these binarized datasets, we compute correlation coefficients $\rho$ across various configurations of $b_1$ and $b_2$, as summarized in Table~\ref{table: rho-1}.

Specifically, Table~\ref{table: rho-1} indicates that when using Blend for both $b_1$ and $b_2$, choosing different (or identical) target classes results in lower (or higher) correlations. Similarly, Table~\ref{table: rho-2} shows that when Blend is used for $b_1$ and BadNets for $b_2$, selecting different (or identical) target classes again leads to lower (or higher) correlations.

\begin{table}[htbp]
\centering
\begin{minipage}{0.49\linewidth}
\captionsetup{font=footnotesize}
  \centering
  \caption{$\rho$ for Blend/Blend}
  \begin{adjustbox}{max width=.99\textwidth}
  \begin{tabular}{|l|c|c|}
\hline
         & \textbf{diff target cls} & \textbf{same target cls} \\
\hline
MNIST-b   & -0.066 & 0.1  \\
SVHN-b    & -0.091 & -0.31 \\
CIFAR10-b & -0.05  & 0.22 \\
\hline
\end{tabular}
\end{adjustbox}
\label{table: rho-1}
\end{minipage}%
\hfill
\begin{minipage}{0.49\linewidth}
\captionsetup{font=footnotesize}
  \centering
  \caption{$\rho$ for Blend/BadNets}
  \begin{adjustbox}{max width=.99\textwidth}
\begin{tabular}{|l|c|c|}
\hline
         & \textbf{diff target cls} & \textbf{same target cls} \\
\hline
MNIST-b   & -0.056 & 0.084 \\
SVHN-b    & -0.028  & -0.036 \\
CIFAR10-b & -0.0138 & 0.0144 \\
\hline
\end{tabular}
\end{adjustbox}
\label{table: rho-2}
\end{minipage}
\end{table}

Table~\ref{table: rho-3} shows that when using Blend for both $b_1$ and $b_2$, a greater (smaller) distance between their trigger locations leads to lower (higher) correlations. Similarly, Table~\ref{table: rho-4} indicates that using complementary (identical) trigger colors for $b_1$ and $b_2$ results in lower (higher) correlations. The above results on $\rho$ justify the preference of different backdoor configurations in \textsc{BadTV}.

\begin{table}[htbp]
\centering
\begin{minipage}{0.49\linewidth}
  \centering
  \caption{$\rho$ for distances}
  \begin{adjustbox}{max width=.99\textwidth}
\begin{tabular}{|l|c|c|c|}
\hline
\textbf{} & \textbf{longest} & \textbf{intermediate} & \textbf{shortest} \\
\hline
MNIST-b   & 0.233 & 0.234 & 0.156 \\
SVHN-b    & 0.005 & 0.016 & 0.013 \\
CIFAR10-b & 0.044 & 0.074 & 0.065 \\
\hline
\end{tabular}
\end{adjustbox}
\label{table: rho-3}
\end{minipage}%
\hfill
\begin{minipage}{0.49\linewidth}
  \centering
  \caption{$\rho$ for trigger colors}
  \begin{adjustbox}{max width=.99\textwidth}
\begin{tabular}{|l|c|c|c|}
\hline
\textbf{} & \textbf{complementary} & \textbf{middle} & \textbf{same} \\
\hline
MNIST-b   & -0.176 & -0.189 & 0.233 \\
SVHN-b    & 0.002  & 0.002  & 0.005 \\
CIFAR10-b & 0.049  & 0.051  & 0.044 \\
\hline
\end{tabular}
\end{adjustbox}
\label{table: rho-4}
\end{minipage}
\end{table}

\begin{table}[!t]
\centering
\footnotesize
\caption{$\rho$ for Blend/Blend}
\label{tab:target_class_comparison}
\begin{tabular}{|c|c|c|c|c|c|}
\hline
\textbf{Target Class} & \textbf{MNIST} & \textbf{SVHN} & \textbf{CIFAR10} & \textbf{CIFAR100} & \textbf{GTSRB} \\ \hline
diff target cls & 0.05 & -0.12 & 0.23 & 0.26 & 0.08 \\ \hline
same target cls & 0.09 & -0.13 & 0.27 & 0.32 & 0.184 \\ \hline
\end{tabular}
\end{table}

Using standard multi-class datasets (e.g., MNIST~\cite{MNIST} and SVHN~\cite{svhn}), we further compute correlation coefficients $\rho$ in Table~\ref{tab:target_class_comparison} with Blend utilized for both $b_1$ and $b_2$. The results indicate that selecting different (identical) target classes for $b_1$ and $b_2$ consistently leads to lower (higher) correlations. In general, greater discrepancies between $b_1$ and $b_2$ correspond to lower correlations, aligning with the theoretical insight from \cite{li2025when} that lower correlation contributes to improved performance of \textsc{BadTV}.

\paragraph{Same vs. Different Target Classes.}
\textcolor{black}{\textsc{BadTV} supports two design choices for the backdoor configurations $b_1$ and $b_2$: using the same or different target classes. From an attacker’s perspective, using the same target class can be more practical, as it ensures consistent behavior regardless of whether addition or subtraction is applied at deployment time. However, as suggested by Figure~9(a), such configurations may reduce attack success rates under task arithmetic. As shown in Tables~\ref{table: rho-1}--\ref{table: rho-4} and Table~\ref{tab:target_class_comparison}, this is due to higher correlation between backdoor components, which increases interference and reduces robustness. In contrast, using different target classes reduces correlation and improves robustness under task arithmetic. Therefore, the choice reflects a trade-off between consistency under unknown operations and robustness under task arithmetic, and we adopt different target classes by default to prioritize robustness.}


\section{Evaluation}\label{sec:experiments}
\subsection{Experiment Settings}\label{subsec:exp_setting}

\textbf{Datasets} We consider five datasets for backdoor attacks: MNIST~\cite{MNIST}, SVHN~\cite{svhn}, CIFAR10~\cite{cifar}, CIFAR100~\cite{cifar}, and GTSRB~\cite{gtsrb}. Additional datasets, including EuroSAT~\cite{eurosat}, Cars~\cite{car}, and SUN397~\cite{sun397}, are employed to assess \textsc{BadTV} under more complex scenarios. 

\textbf{Pre-trained Models} We select three vision models: \texttt{CLIP ViT-B/32}, \texttt{ViT-B/16}, and \texttt{ConvNeXt Base}, covering convolution-based and Transformer-based architectures, all with an input size of $224 \times 224$. For LLMs, we evaluate \texttt{Llama 2-7B}~\cite{touvron2023llama2openfoundation}, \texttt{Llama 3-8B}~\cite{llama3modelcard}, \texttt{Phi-4-14B}~\cite{phi4}, \texttt{Mistral-7B}~\cite{mistral}, and \texttt{DeepSeek-7B}~\cite{deepseekai2025deepseekr1incentivizingreasoningcapability}, where \texttt{Phi-4-14B} represents large-scale models and \texttt{DeepSeek-7B} exemplifies reasoning models.

\textbf{Attack Methods} Backdoor attacks can be categorized into dirty vs. clean label, local vs. global trigger, and static vs. dynamic. We consider BadNets~\cite{gu2019badnets}, Blend~\cite{blend}, WaNet~\cite{wanet}, Dynamic Backdoor (Dynamic)~\cite{dynamicbackdoor}, Narcissus~\cite{narcissus}, Label Consistency (LC)~\cite{LC}, WaveAttack (WA)~\cite{waveattack}, and MTBAs~\cite{mtbas2024}. Narcissus and LC represent clean-label attacks, while the rest are dirty-label. Among dirty-label attacks, BadNets and Dynamic utilize local triggers; Blend and WA employ global triggers; and WaNet uses an invisible trigger. Moreover, MTBAs represents a hybrid design of backdoors. Detailed configurations are provided in Appendix~\ref{appdix:attackmethod}. Additionally, to illustrate \textsc{BadTV}'s compatibility with other poisoning-oriented threats, we include model hijacking attacks~\cite{salem2022ndss}.

\textbf{Metrics} Two metrics are considered: \textit{clean accuracy (CA)} and \textit{attack success rate (ASR)}. CA is defined as the fraction of correctly classified clean test samples, averaging approximately 0.965 in our experiments. ASR is the proportion of test samples classified into the attacker-specified target class. For other poisoning attacks integrated within \textsc{BadTV}, such as model hijacking~\cite{salem2022ndss}, we report accuracy on the hijacking dataset to measure effectiveness, and accuracy on the original (hijackee) dataset to assess stealthiness.

\textbf{Our Setting on \textsc{BadTV}} Except for the experiments in \S\ref{subsubsec:single_BTV}, all evaluations were conducted under the realistic setting. Unless otherwise stated, we use the following default configuration for \textsc{BadTV}. We set $\lambda=0.3$ and $\lambda=0.8$, as \textsc{BadTV} performs effectively with moderately large $\lambda> 0.3$, while smaller values ($\lambda\leq 0.3$) typically harm CA. $\alpha_1$ and $\alpha_2$ are set according to the optimized results in Table~\ref{table: RO result} in \S\ref{sec:robust-feasibility-composite-btv}. Given the poison rate range of $[1\%,20\%]$ common in prior backdoor studies~\cite{blend,gu2019badnets, dynamicbackdoor, wanet, LC, narcissus}, we fix it at $5\%$ to test whether \textsc{BadTV} remains effective under low poisoning conditions. Both backdoor configurations $b_1$ and $b_2$ utilize Blend~\cite{blend}, employing distinct settings (e.g., target classes, trigger locations, and patterns). Some experiments involve pairing different backdoor configurations for $b_1$ and $b_2$. Given the extensive number of possible combinations, we present results primarily using Blend as the representative default case. The attacker-specified task $t$ is fixed as GTSRB. The default model architecture is \texttt{ViT-B-16}.

\begin{table*}[t]
    \caption{CA/ASR under addition and subtraction (Add-CA, Sub-CA, Add-ASR and Sub-ASR, respectively)}
    \centering
    \begin{adjustbox}{max width=0.95\textwidth}
    \begin{tabular}{c|ccccccccc}
    \toprule
    \makecell[c]{\textbf{Add-CA} ({$\uparrow$}) / \textbf{Add-ASR ({$\uparrow$})} \\\textbf{Sub-CA ({$\downarrow$})} / \textbf{Sub-ASR} ({$\uparrow$})}
    & BadNets~\cite{gu2019badnets} & Blend~\cite{blend}&WaNet~\cite{wanet}& Dynamic~\cite{dynamicbackdoor} & Narcissus~\cite{narcissus} & LC~\cite{LC} & WA~\cite{waveattack}&\makecell[c]{MTBAs~\cite{mtbas2024}\\(parallel)}&\makecell[c]{MTBAs~\cite{mtbas2024}\\(sequential)}\\
    \hline
    MNIST (10-class) 
& \twolines{99.71 / 99.7}{15.41 / 98.92}
& \twolines{99.37 / 92.4}{13.04 / 100}
& \twolines{99.64 / 100}{11.35 / 100}
& \twolines{99.66 / 100}{11.35 / 100}
& \twolines{99.28 / 10.36}{11.35 / 11.54}
& \twolines{98.8 / 90.8}{15.7 / 96.2} 
& \twolines{99.82 / 99.8}{0.41 / 99.87}
& \twolines{92.62 / 99.85}{12 / 99.92}
& \twolines{96.31 / 100}{12.32 / 100}\\
\hline
SVHN (10-class) 
& \twolines{97.38 / 99.77}{18.33 / 82.25}
& \twolines{96.85 / 100}{19.59 / 100}
& \twolines{96.53 / 99.79}{19.58 / 100}
& \twolines{98.63 / 100}{19.58 / 100}
& \twolines{96.99 / 99.08}{19.58 / 100}
& \twolines{93.3 / 91.9}{13.9 / 94.6} 
& \twolines{98.25/99.9}{17.51/100} 
& \twolines{96.02 / 99.75}{19.59 / 100}
& \twolines{97.17 / 100}{19.58 / 100}\\
\hline
CIFAR10 (10-class) 
& \twolines{97.68 / 99.06}{9.99 / 99.97}
& \twolines{97.71 / 100}{10.29 / 100}
& \twolines{97.22 / 99.77}{10.00 / 100}
& \twolines{98.36 / 100}{10.00 / 100}
& \twolines{97.24 / 99.39}{11.07 / 100}
& \twolines{94.6 / 92.5}{8.7 / 94.8} 
& \twolines{98.15 / 100}{9.88 / 99.95} 
& \twolines{97.47 / 99.75}{10 / 100}
& \twolines{97.89 / 100}{10 / 100}\\
\hline
CIFAR100 (100-class) 
& \twolines{88.36 / 98.82}{1.00 / 100}
& \twolines{84.3 / 99.98}{0.89 / 99.97}
& \twolines{87.59 / 99.78}{1.00 / 100}
& \twolines{89.37 / 100}{2.59 / 97.54}
& \twolines{87.30 / 98.73}{3.58 / 100}
& \twolines{81.7 / 90.3}{4.6 / 93.5} 
& \twolines{89.23 / 100}{1.12 / 99.98}
& \twolines{89.63 / 98.20}{1.02 / 100}
& \twolines{90.38 / 99.98}{7.39 / 100}\\
\hline
GTSRB (43-class)
& \twolines{98.20 / 99.91}{5.70 / 99.99}
& \twolines{98.94 / 100}{1.96 / 98.16}
& \twolines{98.15 / 99.79}{5.72 / 99.97}
& \twolines{99.66 / 100}{0.48 / 100}
& \twolines{97.33 / 86.91}{3.99 / 100}
& \twolines{84.3 / 86.6}{5.9 / 95.4}
& \twolines{97.96 / 99.21}{4.74 / 100} 
& \twolines{98.57 / 99.65}{5.84 / 99.98}
& \twolines{98.78 / 100}{2.91 / 84.38}\\
    \bottomrule
    \end{tabular}
    \end{adjustbox}
    \label{tab:single_BTV}
\end{table*}

\subsection{Main Results}\label{subsec:main_result}
\subsubsection{Single BTV on Simplified Setting}\label{subsubsec:single_BTV}
We first present the CA and ASR results for various backdoor attacks and datasets under the simplified setting. Table~\ref{tab:single_BTV} summarizes the results, where nearly all ASRs exceed 98\% with the only exception that Narcissus on MNIST achieves only approximately 10\% ASR for both addition and subtraction. This can be attributed to the fact that as Narcissus primarily targets colored images even in CNN/ViT-classifier setups. Additionally, a lower CA under subtraction is desirable since this operation corresponds to \textit{forgetting} the task. Overall, \textsc{BadTV} consistently attains high ASR while preserving reasonable CA across different backdoor methods and datasets. \textcolor{black}{Besides, compared to a matched benign-TV baseline, Table~\ref{tab:benign_acc} shows that the CA difference remains within 2\% across all tasks, indicating a small utility loss.}
\begin{table}[h]
    \centering
    \caption{\textcolor{black}{Accuracy (Acc) under addition and subtraction of a benign task vector.}}
    \begin{adjustbox}{max width=\linewidth}
    \begin{tabular}{l|ccccc}
    \toprule
     & MNIST & SVHN & CIFAR10 & CIFAR100 & GTSRB \\
    \midrule
    Add-Acc./Sub-Acc & 99.78/10.32 & 98/12.72 & 98.72/10.94 & 90.77/1.08 & 99.32/8.89 \\
    \bottomrule
    \end{tabular}
    \end{adjustbox}
    \label{tab:benign_acc}
\end{table}
\subsubsection{Single BTV on Realistic Setting}\label{subsubsec:BTV+CTV}
\begin{figure}[ht!]
    \centering
    \vspace{-8pt}
    \subfigure[$\lambda=0.3$]{
    \includegraphics[width=.225\textwidth]{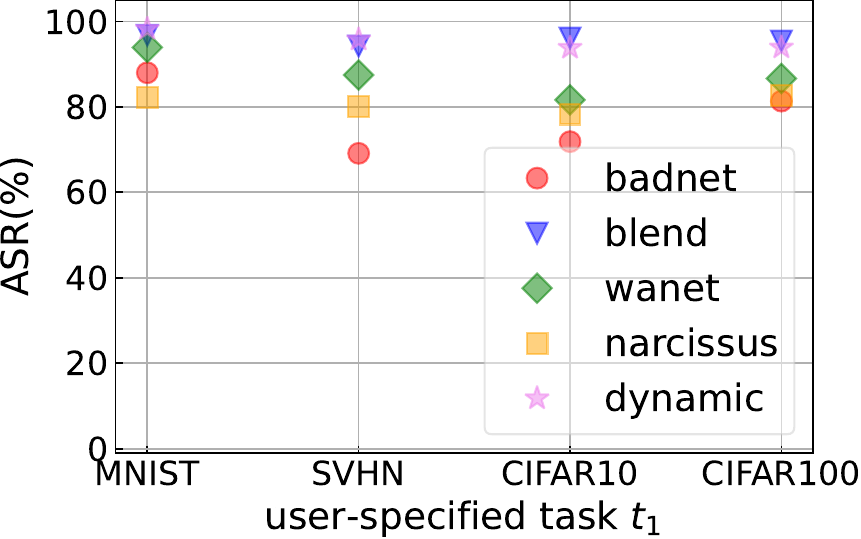}
    }
    \subfigure[$\lambda=0.8$]{
    \includegraphics[width=0.225\textwidth]{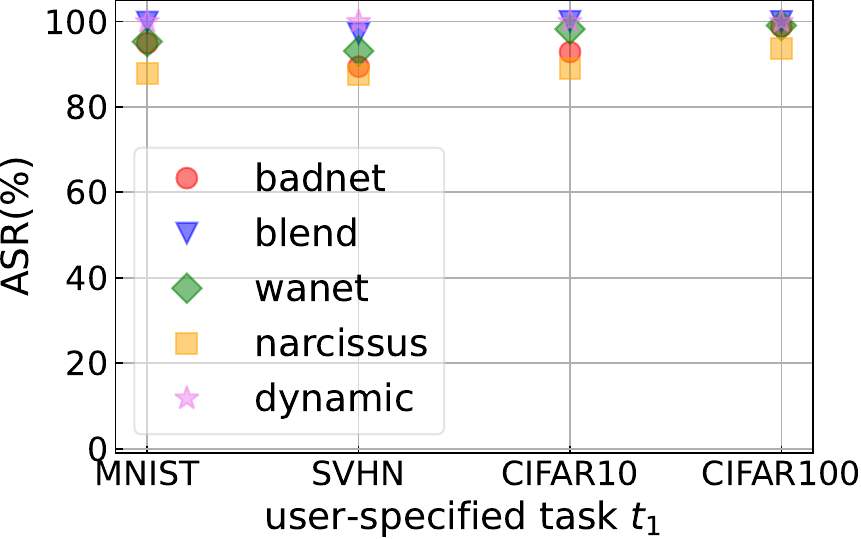}}
    \caption{\footnotesize ASR for different $t_1$ with single BTV under realistic setting.}
    \label{fig:1+1_difClean}
\end{figure}

The experiments here resemble those in \S\ref{subsubsec:single_BTV}, but are conducted under the \textit{realistic} setting, with $\theta_{\text{merged}} = \theta_{\text{victim}} \oplus \lambda \hat{\tau}_{t}$, where $\theta_{\text{victim}}=\theta_{\text{pre}}\oplus\lambda_1\theta_1$ and $\theta_1$ corresponds to a user-selected task $t_1$. Given an average CA of $96.5\%$, Figure~\ref{fig:1+1_difClean} shows \textsc{BadTV} continues to achieve consistently high ASR.

\subsubsection{Alternative Data Poisoning-Based Attacks under Realistic Setting}\label{sec: Alternative Poisoning with BadTV}

\begin{table}[t]
    \caption{Addition/subtraction results showing original task accuracy (O-Acc) and hijacking ASR (H-ASR), where ``$+/-$'' denote addition and subtraction, respectively. \textit{NA} indicates settings infeasible for the model hijacking attack.}
    \centering
    \begin{adjustbox}{max width=.49\textwidth}

    \begin{tabular}{c|ccccc}
    \toprule
     \makecell[c]{\textbf{$+$ O-Acc ({$\uparrow$}) / $+$  H-ASR ({$\uparrow$})}\\ \textbf{$-$ O-Acc ({$\downarrow$}) / $-$ H-ASR ({$\uparrow$})}}   &  MNIST & SVHN & CIFAR10 & EuroSAT & GTSRB  \\\hline
      MNIST & \emph{NA} & \makecell[c]{98.37 / 98.93\\92.51 / 97.42} &\makecell[c]{99.58 / 99.79\\59.08 / 97.77} & \makecell[c]{99.31 / 99.35 \\ 61.47 / 94.24}&\emph{NA}  \\\hline
      SVHN& \makecell[c]{36.17 / 95.19 \\ 90.72 / 96.43}&\emph{NA} &\makecell[c]{96.21 / 96.57\\53.59 / 97.33}&\makecell[c]{92.28 / 92.65\\70.55 / 98.48}&\emph{NA}   \\\hline
      CIFAR10&\makecell[c]{86.89 / 90.99\\ 48.91 / 98.88}&\makecell[c]{91.4 / 95.03\\47.01 / 97.69}&\emph{NA} &\makecell[c]{90.71 / 93.11 \\66.11 / 98.54}&\emph{NA}  \\\hline
      EuroSAT& \makecell[c]{84.44 / 99.26 \\40.67 / 97.15}&\makecell[c]{94.46 / 97.19\\27.48 / 96.22}&\makecell[c]{85.24 / 98.69\\59.8 / 95.8}&\emph{NA} &\emph{NA}  \\\hline
      GTSRB&\makecell[c]{94.43 / 99.48\\11.07 / 99.6}&\makecell[c]{94.17 / 99.27\\25.91 / 97.36}&\makecell[c]{97.72 / 99.97\\34.6 / 96.8}&\makecell[c]{97 / 99.68\\22.34 / 98.59}&\emph{NA} \\\hline
      CIFAR100&\makecell[c]{75.41 / 96.59 \\13.27 / 99.72}&\makecell[c]{68.95 / 93.58\\19.72 / 97.6}&\makecell[c]{66.04 / 93.45\\58.72 / 97.7}&\makecell[c]{74.37 / 94.94\\23.62 / 98.81}&\makecell[c]{67.94 / 83.44\\13.39 / 98.85} \\
      \bottomrule
    \end{tabular}
    \end{adjustbox}
    \label{tab:sing_BTV_hijakc}
\end{table}

The backdoor attacks we consider all rely on data poisoning, prompting the question of whether non-backdoor data poisoning attacks can also succeed within \textsc{BadTV}. To address this, we evaluate the data poisoning-based model hijacking attack~\cite{salem2022ndss}, which aims to covertly equip a model to classify both the original and a hijacking task. We implement this attack using \textsc{BadTV} on the \texttt{ViT-B-32} model. Table~\ref{tab:sing_BTV_hijakc} demonstrates that \textsc{BadTV} successfully maintains the original task's performance while achieving high hijacking ASR. However, when the original and hijacking tasks are overly similar (e.g., MNIST vs. SVHN), the forgetting mechanism struggles to reduce performance effectively due to overlapping features. Furthermore, CIFAR100 presents lower overall performance because its numerous classes and complex imagery complicate feature mapping, a limitation inherent to the original hijacking method. Lastly, the inherent design of hijacking attack requires the number of classes in the hijacking task to be less than or equal to the original task. For example, GTSRB (43 classes) cannot serve as a hijacking task with MNIST (10 classes) as the original task. Such cases are marked as \textit{NA} in   Table~\ref{tab:sing_BTV_hijakc}.

\subsubsection{Post-Merging Processing}\label{sec: Post-Merging Processing}
In practice, it is unrealistic to assume that users directly deploy downloaded BTVs without further processing. Instead, post-merging operations such as fine-tuning and quantization are commonly applied. We evaluate the robustness of \textsc{BadTV} under representative post-processing pipelines, including quantization (fp32$\rightarrow$fp16/int8) and fine-tuning via full fine-tuning, LoRA, and QLoRA. We set the attacker-specified task $t$ as GTSRB and the benign task $t_1$ as CIFAR10. The merged models are evaluated after applying each post-processing method, with results summarized in Table~\ref{tab:post-method} and Table~\ref{tab:quantization}. For fine-tuning, we vary the amount of clean data, while for quantization we consider different numerical precisions. Across all settings, the ASR of \textsc{BadTV} remains consistently high, indicating that they cannot degrades the effectiveness of \textsc{BadTV}.

\begin{table}[hbt!]
    \centering
\begin{minipage}{0.6\linewidth}
\captionsetup{font=footnotesize}
  \centering
        \caption{CA/ASR for different fine-tuning methods.}
    \begin{adjustbox}{max width=0.99\textwidth}
    \begin{tabular}{cccc}
    \toprule
     Method & \# of clean data & CA for $t$ (\%) & ASR (\%)\\\midrule
     \multirow{4}{*}{Full/LoRA/QLoRA} &10\%& 99.03/99/98.96& 93.88/95.43/93.59 \\
      &5\% &98.96/98.15/98.95&93.49/93.19/93.53\\
      &3\% &99.03/98.96/98.97&93.01/92.19/93.79\\
      &1\% &98.95/98.95/98.93&91.88/91.61/92.15\\
 \bottomrule
    \end{tabular}
    \end{adjustbox}
    \label{tab:post-method}
\end{minipage}
\hfill
\begin{minipage}{0.39\linewidth}
\captionsetup{font=footnotesize}
    \centering
       \caption{CA/ASR for different quantization precisions.}
    \begin{adjustbox}{max width=0.99\textwidth}
    \begin{tabular}{cccc}
    \toprule
        precision  & CA for $t$ & ASR\\\midrule
         Original (fp32) &98.62\% & 98.86\%\\ \midrule
         Quantization (fp16) &  98.30\% &99.48\% \\
         Quantization (int8) &  98.30\% &99.48\% \\ \midrule
    \end{tabular}
    \end{adjustbox}
 
    \label{tab:quantization}
\end{minipage}
\end{table}

\subsection{Influencing Factors for \textsc{BadTV}}\label{subsec:badtv_factors}
All experiments here are conducted under the \textit{realistic} setting with a benign task, $\theta_{\text{merged}} = \theta_{\text{victim}} \oplus \lambda \hat{\tau}_{t}$, where the attacker-specified task $t$ is GTSRB and $t_1$ is CIFAR100.

\subsubsection{Model Architecture}\label{subsubsec:model_architecture}

\begin{figure*}[t]
    \centering
    \subfigure[\texttt{ViT-B-16} (Add)]{\includegraphics[width=0.15\textwidth]{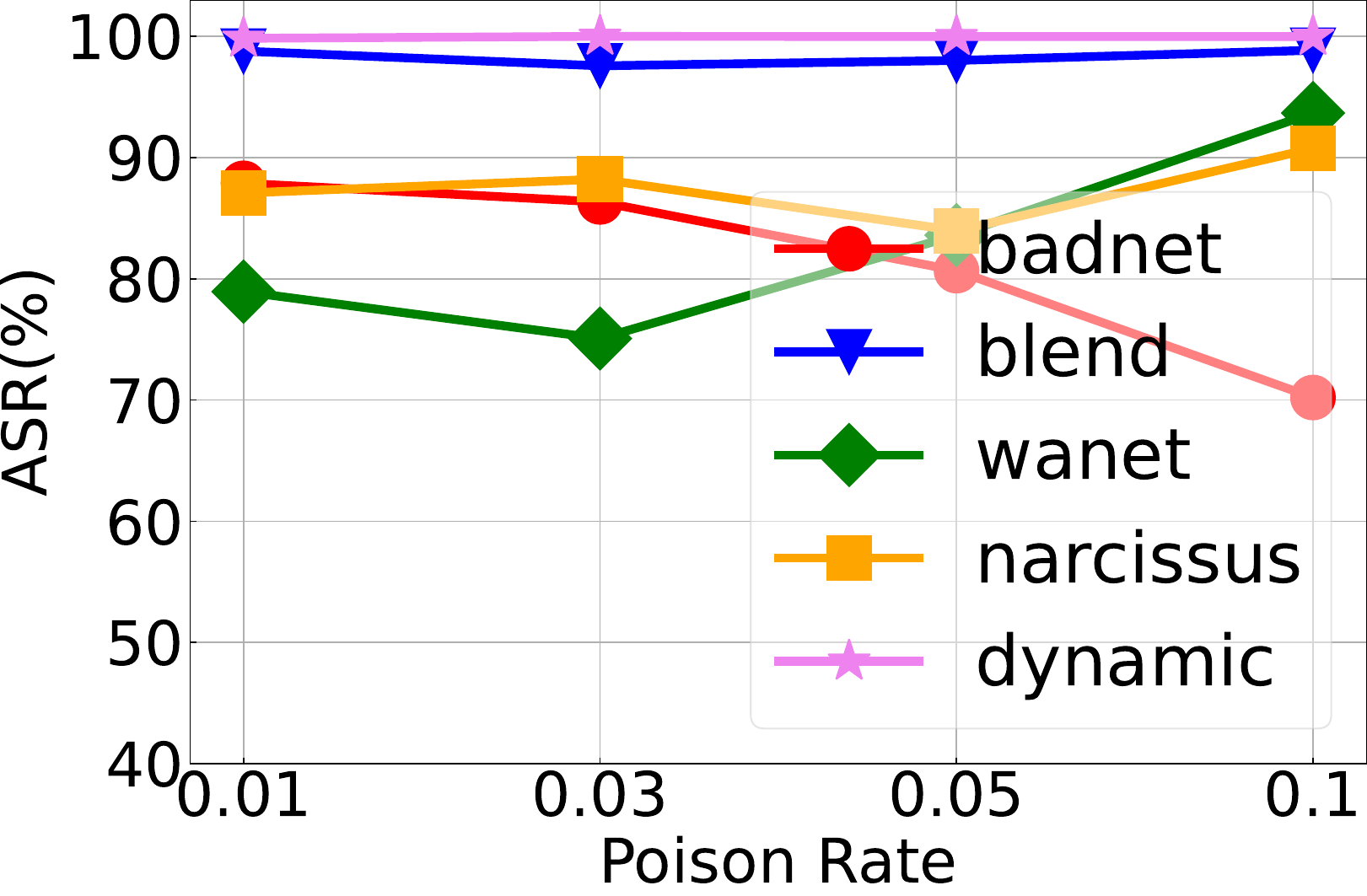}}    
    \subfigure[\texttt{ViT-B-16} (Sub)]{\includegraphics[width=0.15\textwidth]{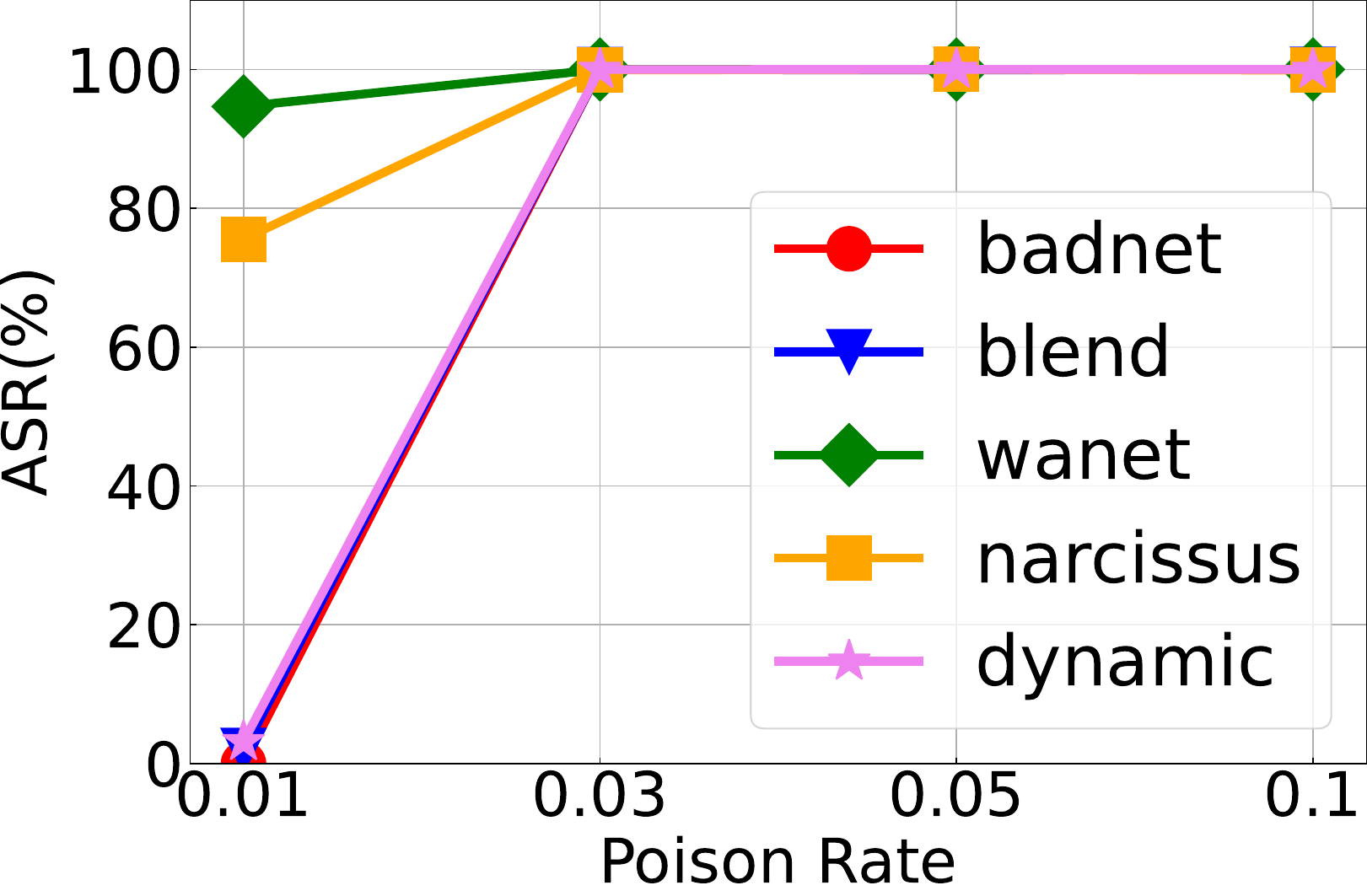}}	
    \subfigure[\texttt{ViT-B-32} (Add)]{\includegraphics[width=0.15\textwidth]{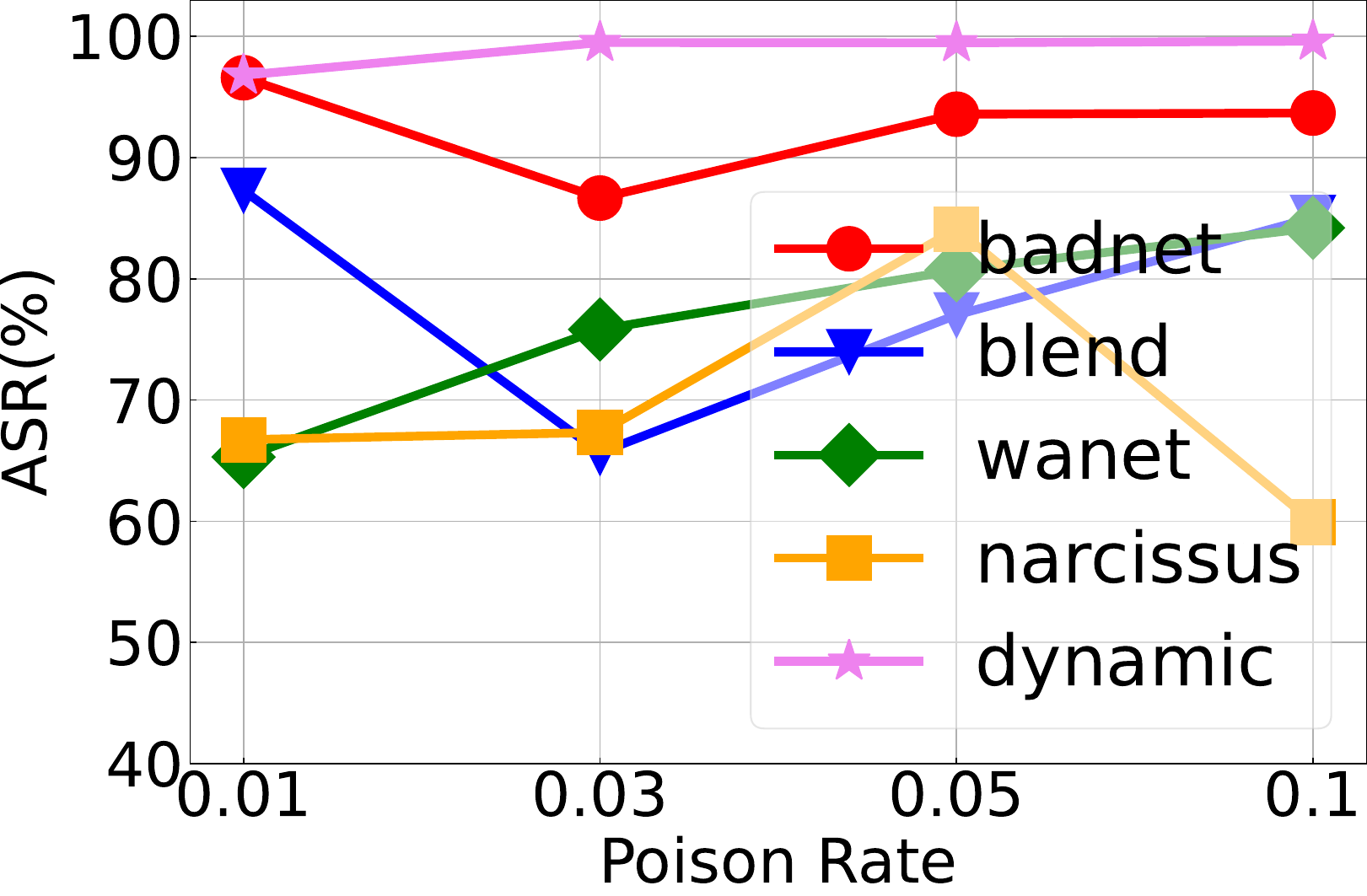}}
    \subfigure[\texttt{ViT-B-32} (Sub)]{\includegraphics[width=0.15\textwidth]{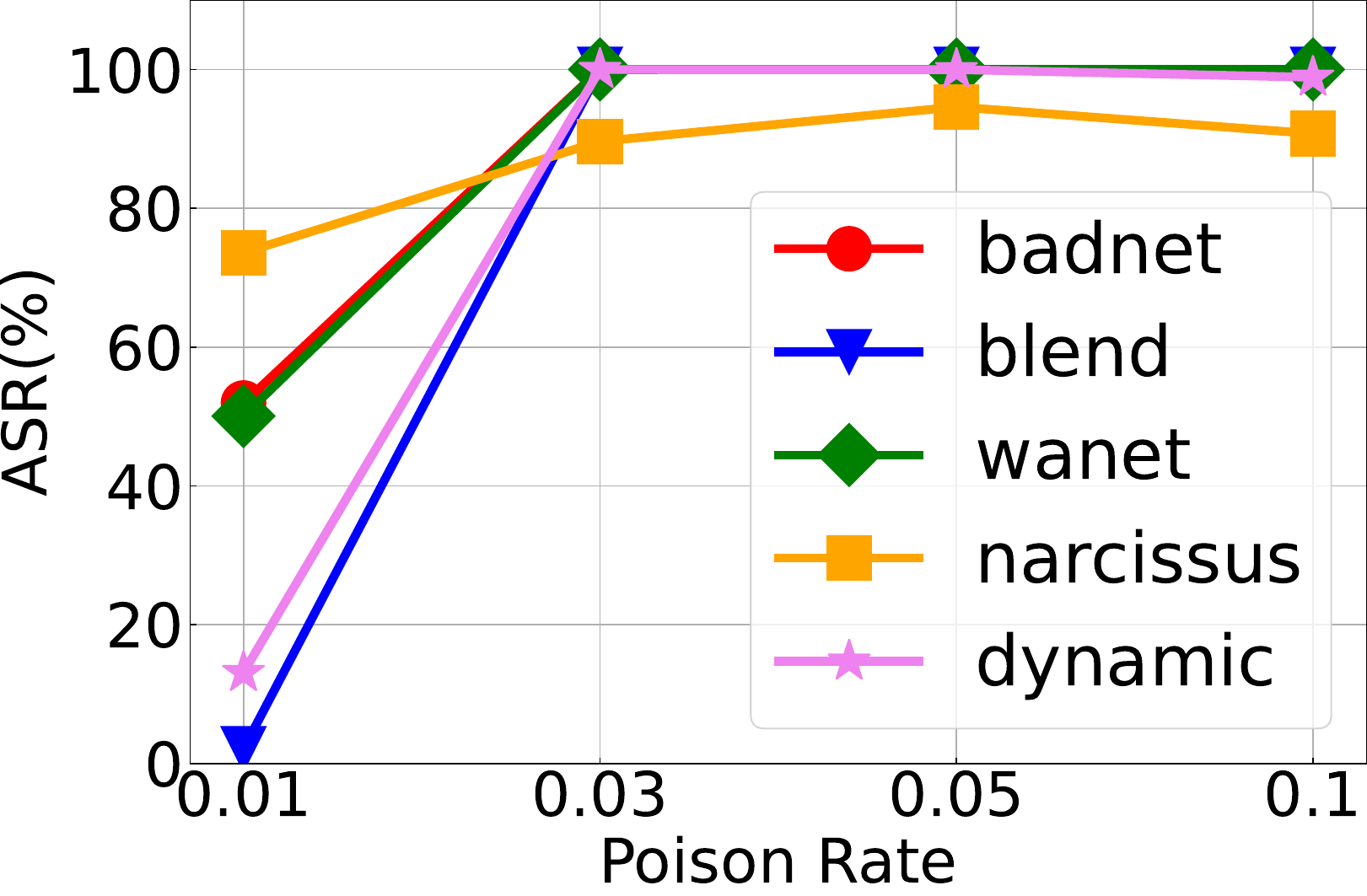}}
    \subfigure[\texttt{ConvNeXt Base} (Add)]{\includegraphics[width=0.15\textwidth]{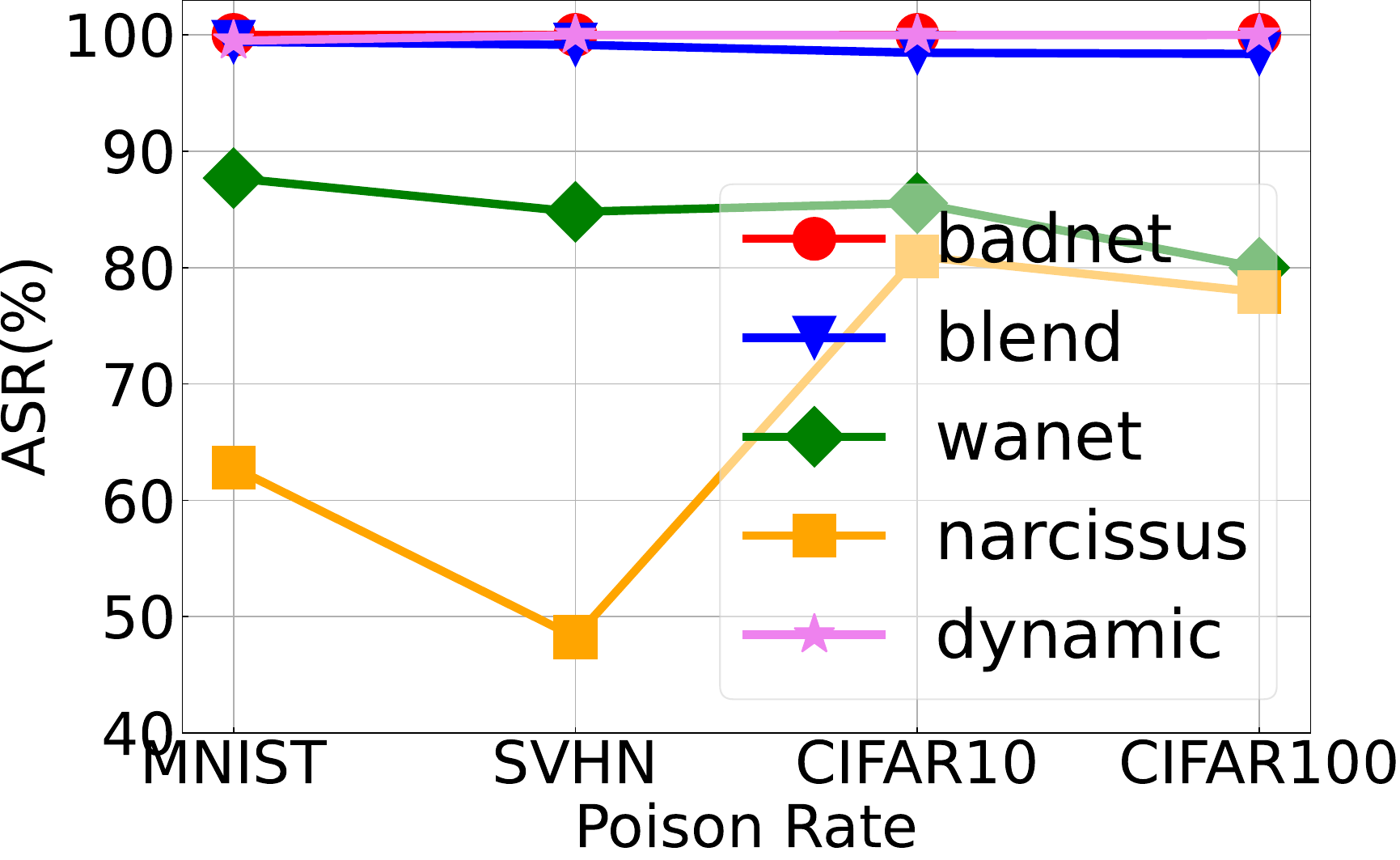}}    \subfigure[\scriptsize\texttt{ConvNeXt Base} (Sub)]{\includegraphics[width=0.15\textwidth]{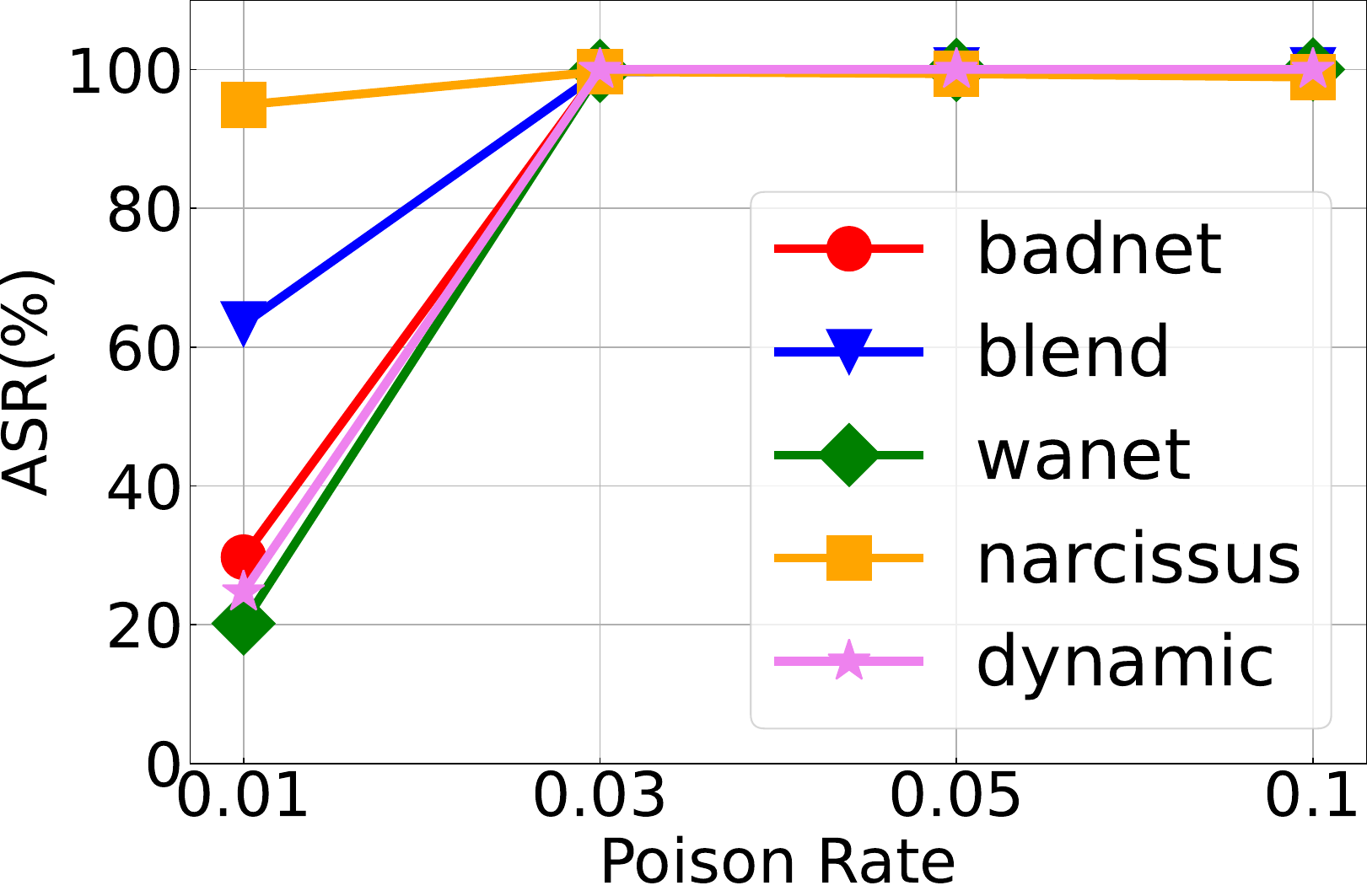}}
    \caption{The impact of model architecture on ASR.}
    \Description{}
     \label{fig:model_architectures}
\end{figure*}

We investigate the impact of model architecture on \textsc{BadTV}'s effectiveness. Following the experimental setup in \S\ref{subsec:main_result}, we replace the \texttt{ViT-B} architecture with convolution-based models, specifically \texttt{ConvNeXt Base}. Figure~\ref{fig:model_architectures} demonstrates that \textsc{BadTV} maintains robust performance across diverse architectures. Despite a slight performance drop observed in \texttt{ConvNeXt Base}, the majority of attacks continue to succeed, highlighting \textsc{BadTV}'s architectural adaptability.

\subsubsection{Backdoor Pairing in \textsc{BadTV}}\label{subsubsec:backdoor_combo}
\begin{figure}
    \centering
    \subfigure[Same ($\lambda=0.3$)]{
    \includegraphics[width=.22\textwidth]{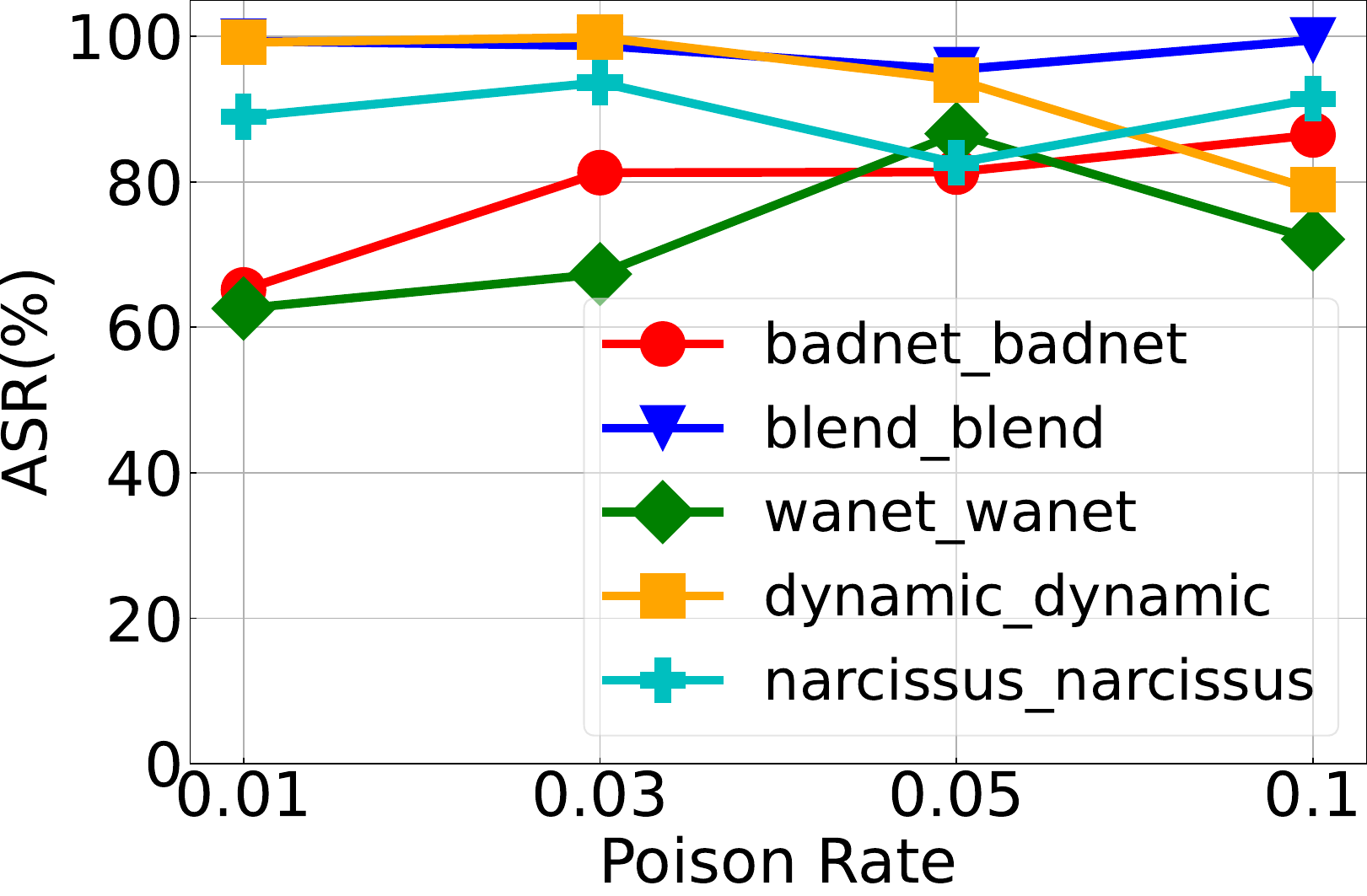}
    \label{fig:backdoor_combo-a}}
    \subfigure[Same ($\lambda=0.8$)]{
    \includegraphics[width=.22\textwidth]{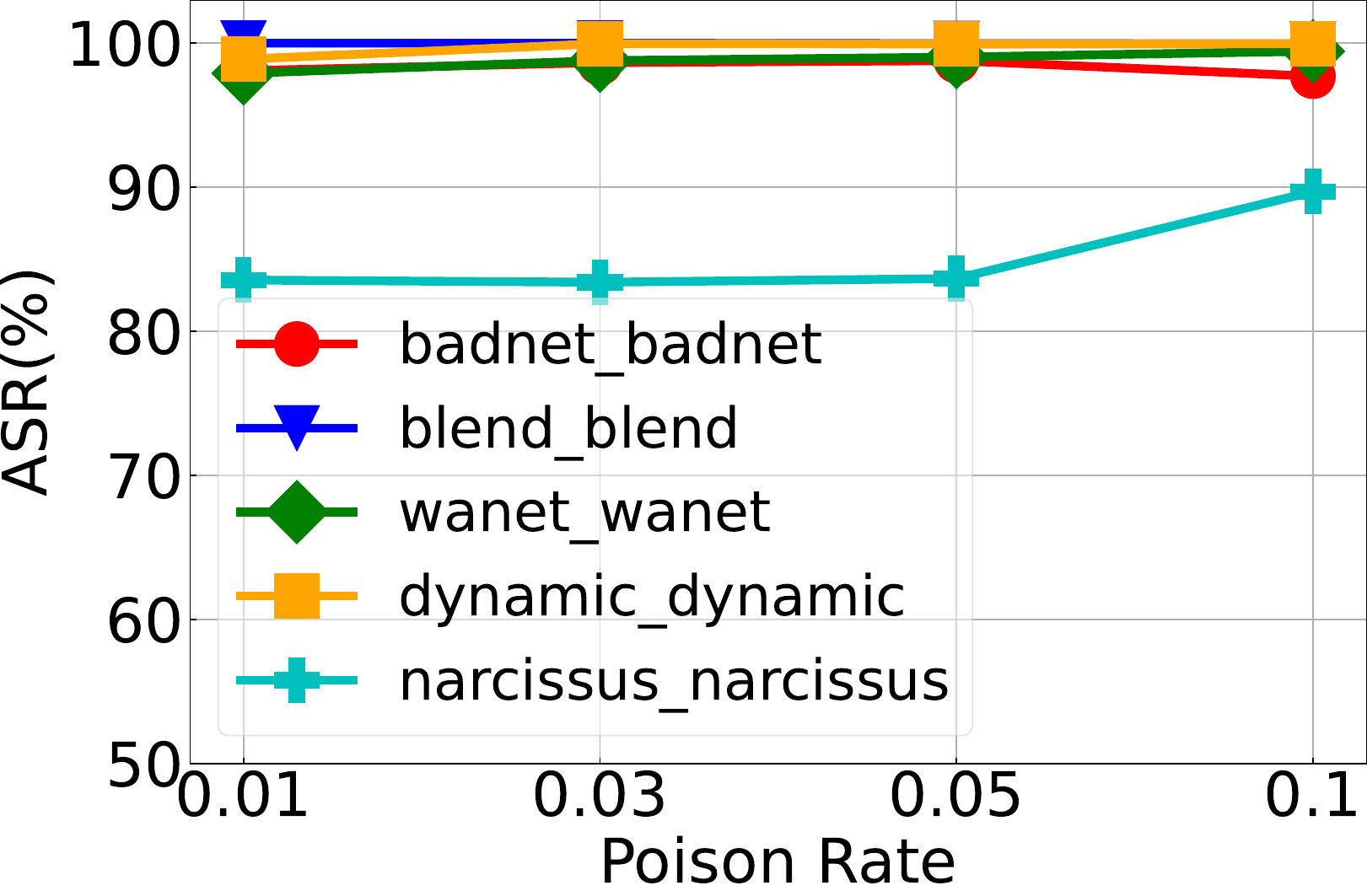}
    \label{fig:backdoor_combo-b}}
    \subfigure[Different ($\lambda=0.3$)]{
    \includegraphics[width=.22\textwidth]{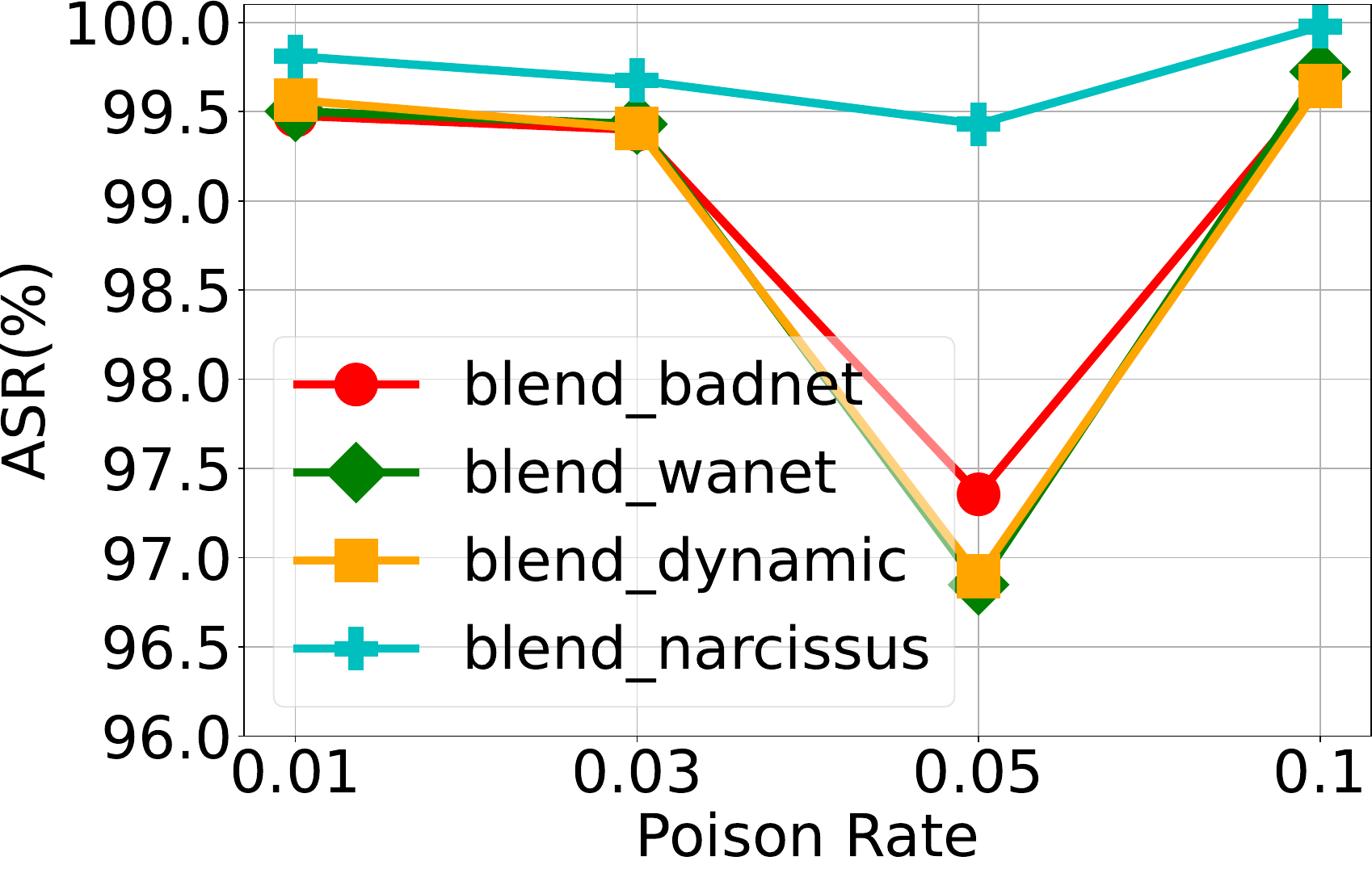}
    \label{fig:backdoor_combo-c}}
    \subfigure[Different ($\lambda=0.8$)]{
    \includegraphics[width=.22\textwidth]{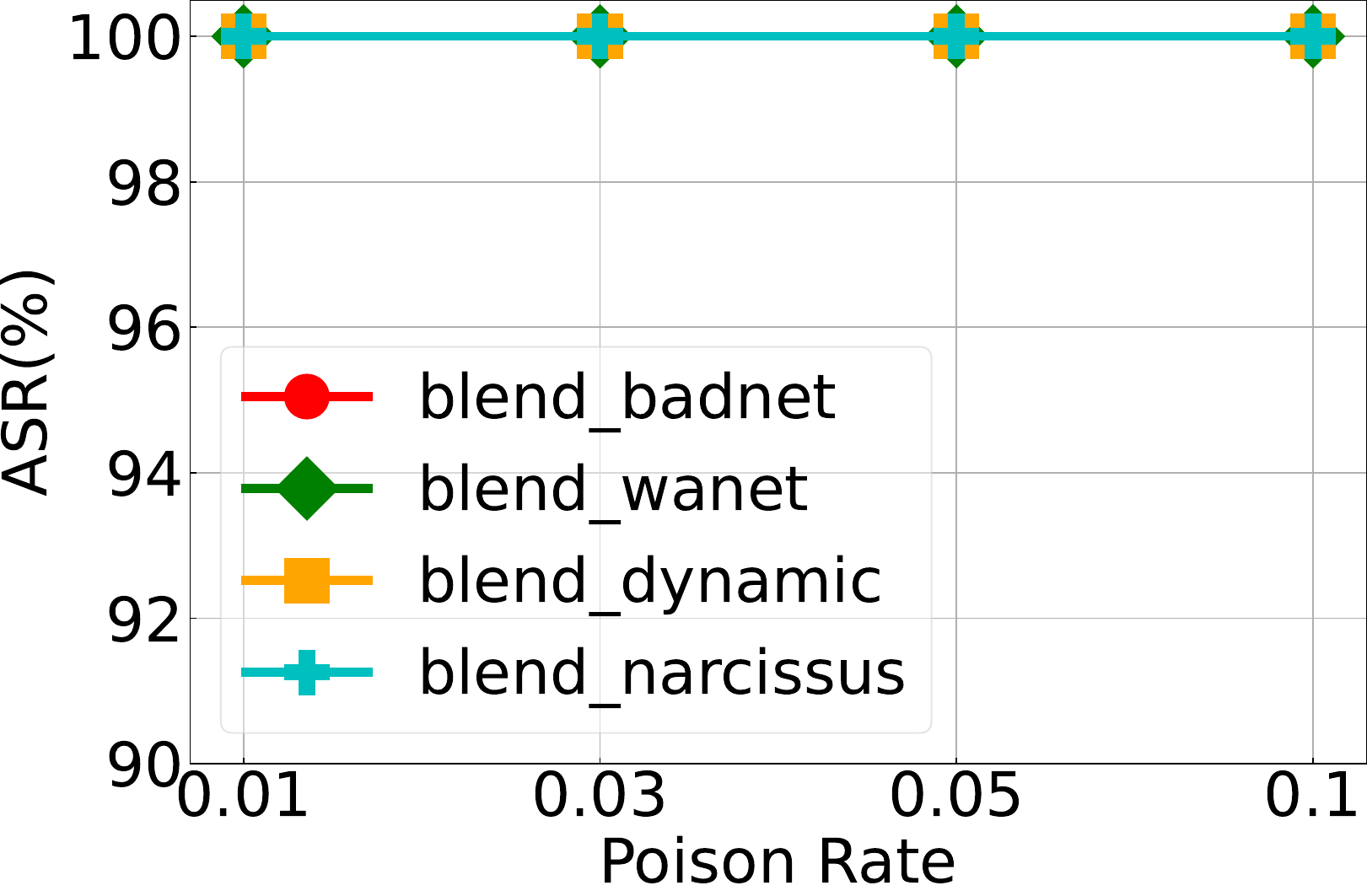}
    \label{fig:backdoor_combo-d}}
    \caption{The impact of backdoor pairing on ASR.}
    \Description{}
    \label{fig:backdoor_combo}
\end{figure}

Since the composite backdoor in \textsc{BadTV} effectively consists of two separate backdoors, it is critical to assess how the selection of these individual backdoors affects \textsc{BadTV}'s ASR. Figure~\ref{fig:backdoor_combo-a} shows that employing identical backdoor attacks for $b_1$ and $b_2$ can substantially reduce ASR, whereas mixing different attack types generally boosts ASR (see Appendix~\ref{app:how to derive} for details). This observation aligns with the conclusions presented in \S\ref{sec:why the setting of different target classes}.

While Figure~\ref{fig:backdoor_combo-a} sets $t_1$ as CIFAR100, Table~\ref{tab:1+1_diffAttack} in Appendix~\ref{appdix:backdoor-pairing} further evaluates CA and ASR across a variety of tasks including MNIST, SVHN, CIFAR10, and CIFAR100. As demonstrated in Table~\ref{tab:1+1_diffAttack}, both CA and ASR remain consistently high.

The slight ASR drop of Narcissus in Figure~\ref{fig:backdoor_combo-b} (compared to Figure~\ref{fig:backdoor_combo-a}) is not incidental, but stems from its trigger design being intrinsically more sensitive to scaling-induced feature-space distortion under TA; a detailed analysis is provided in Appendix~\ref{app:Why Narcissus'ASR reduces}.

\begin{figure}[ht]
    \centering
    \subfigure[$\lambda = 0.3$]{
    \includegraphics[width=.225\textwidth]{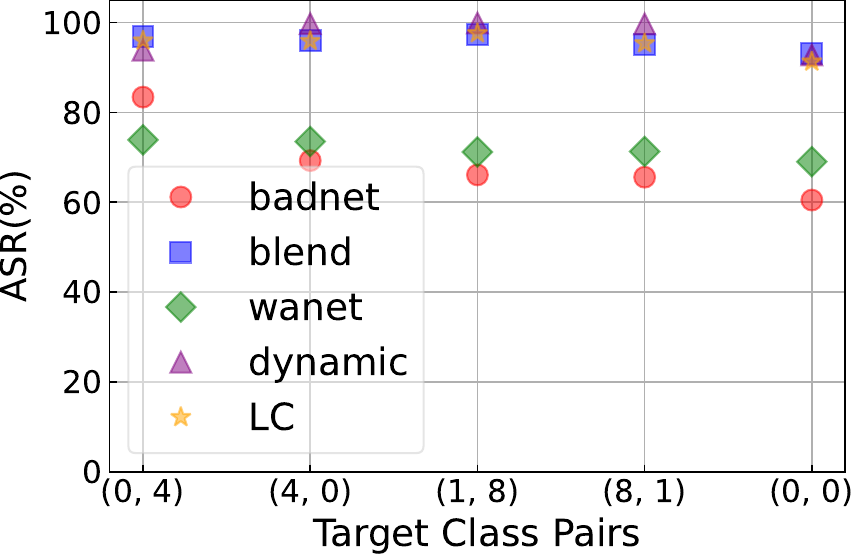}
    }
    \subfigure[$\lambda = 0.8$]{
    \includegraphics[width=.225\textwidth]{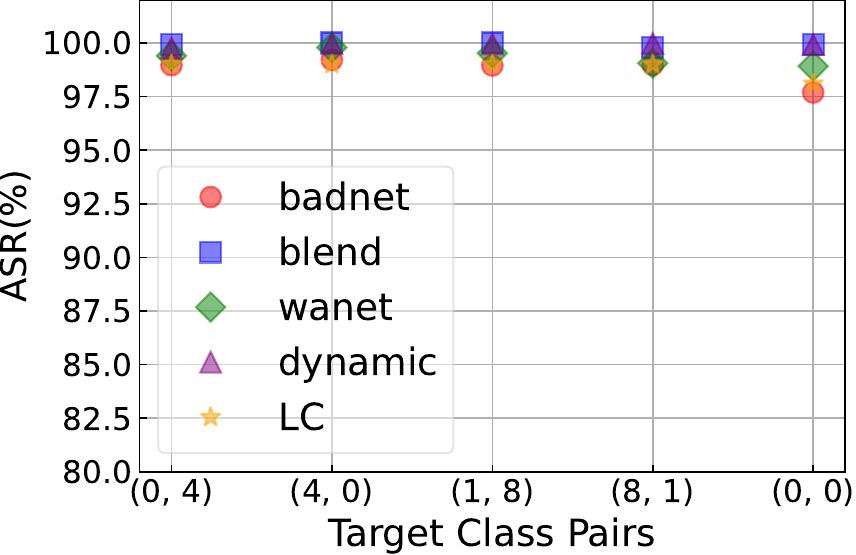}}
    \caption{The impact of target class selection on ASR.}
    \Description{}
    \label{fig:target_class_pairs}
\end{figure}

\subsubsection{Target Class Selections}\label{subsubsec:target_class}
We investigate whether the selection of target classes affects \textsc{BadTV}'s ASR, specifically comparing scenarios where $b_1$ and $b_2$ share identical versus different target classes. Figure~\ref{fig:target_class_pairs} reports the results for five representative target-class pairs. For both $\lambda=0.3$ and $0.8$, ASR decreases significantly when $b_1$ and $b_2$ share the same target class (e.g., pair $(0,0)$), indicating reduced stability. Overall, choosing distinct target classes for $b_1$ and $b_2$ is recommended for achieving stable ASR performance in \textsc{BadTV}. This observation also aligns with the conclusions presented in \S\ref{sec:why the setting of different target classes}. \textcolor{black}{However, ASR can still remain high under the same-target setting when $b_1$ and $b_2$ are trained with different backdoor attacks. For the same-target pair $(4,4)$, strong ASR is still observed for $(b_1, b_2)$ = (Blend, Blend), (Blend, WaNet), and (Blend, BadNet), with ASRs of 93.1, 97.16, and 97.51 at $\lambda = 0.3$.}

\subsubsection{Different Task Arithmetic Methods}\label{subsection:dif_ta_method}
The previous discussions rely primarily on ordinary TA~\cite{ilharco2023iclr}, prompting us to evaluate the robustness of \textsc{BadTV} against alternative TA methods. We note that MM methods (see \S\ref{sec: related work}), such as TIES-merging~\cite{ties-merging} and DARE~\cite{dare}, do not simultaneously support addition and subtraction operations.

\begin{figure}[h]
    \centering
    \subfigure[$\lambda=0.3$]{
    \includegraphics[width=.225\textwidth]{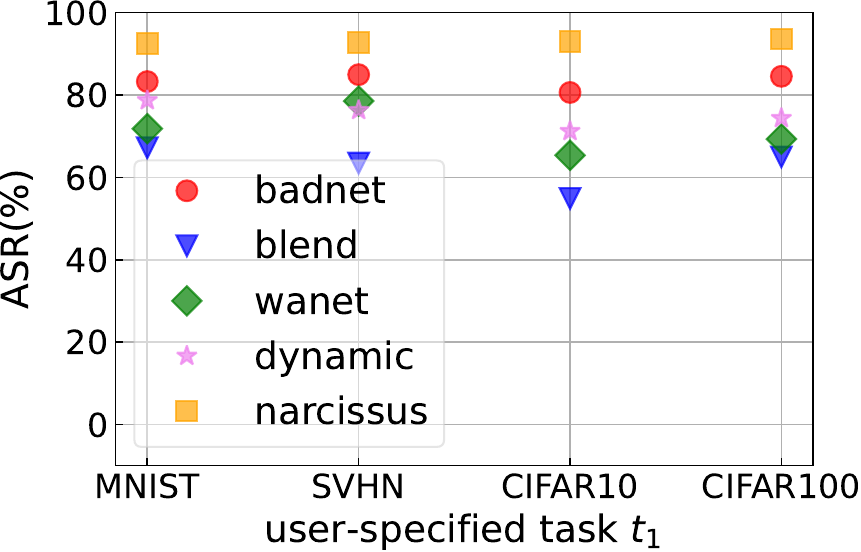}
    }
    \subfigure[$\lambda=0.8$]{
    \includegraphics[width=0.225\textwidth]{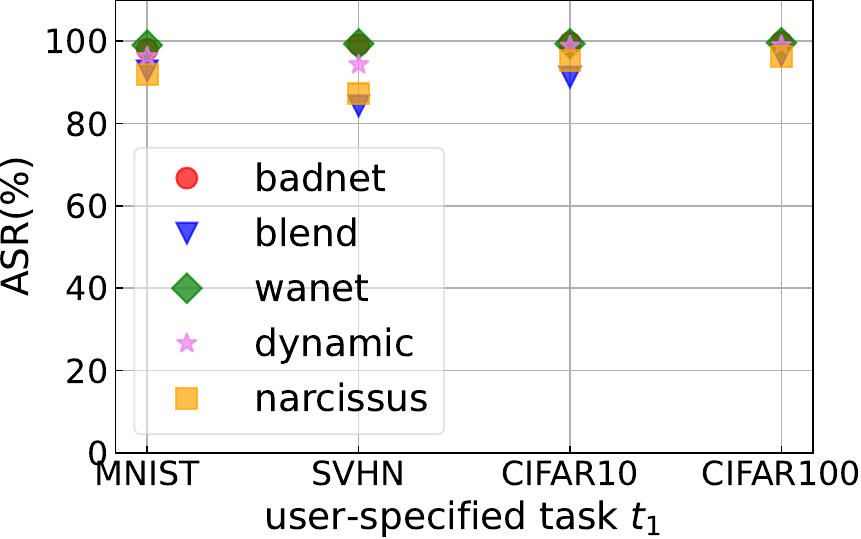}}
    \caption{The impact of TA method on ASR.}
    \Description{}
    \label{fig:atlas_1+1_difClean}
\end{figure}

We specifically assess \textsc{BadTV}'s robustness using aTLAS~\cite{aTLAS}, a recently proposed TA method. Following the experimental setup in Figure~\ref{fig:1+1_difClean}, we construct the BTV $\hat{\tau}_t$ via Equation~\eqref{eq:backdoor_TV}. However, users now compute the merged model as $\hat{\theta}_{\text{merged}} = \theta_{\text{victim}} \boxplus \hat{\tau}_t$, where $\boxplus$ denotes the addition or subtraction operation defined by aTLAS. Figure~\ref{fig:atlas_1+1_difClean} presents the outcomes when various CTVs are integrated using aTLAS. The results align closely with Figure~\ref{fig:1+1_difClean}, showing that nearly all attacks maintain an ASR above $70\%$ (also see Appendix~\ref{app:how to derive} for details).

\subsection{\textsc{BadTV} in the Wild}\label{subsec:badtv_wild}
\begin{figure*}[!ht]
    \centering
    \subfigure[Same ($\lambda=0.3$)]{\label{fig:1+3-a}
    \includegraphics[width=.21\textwidth]{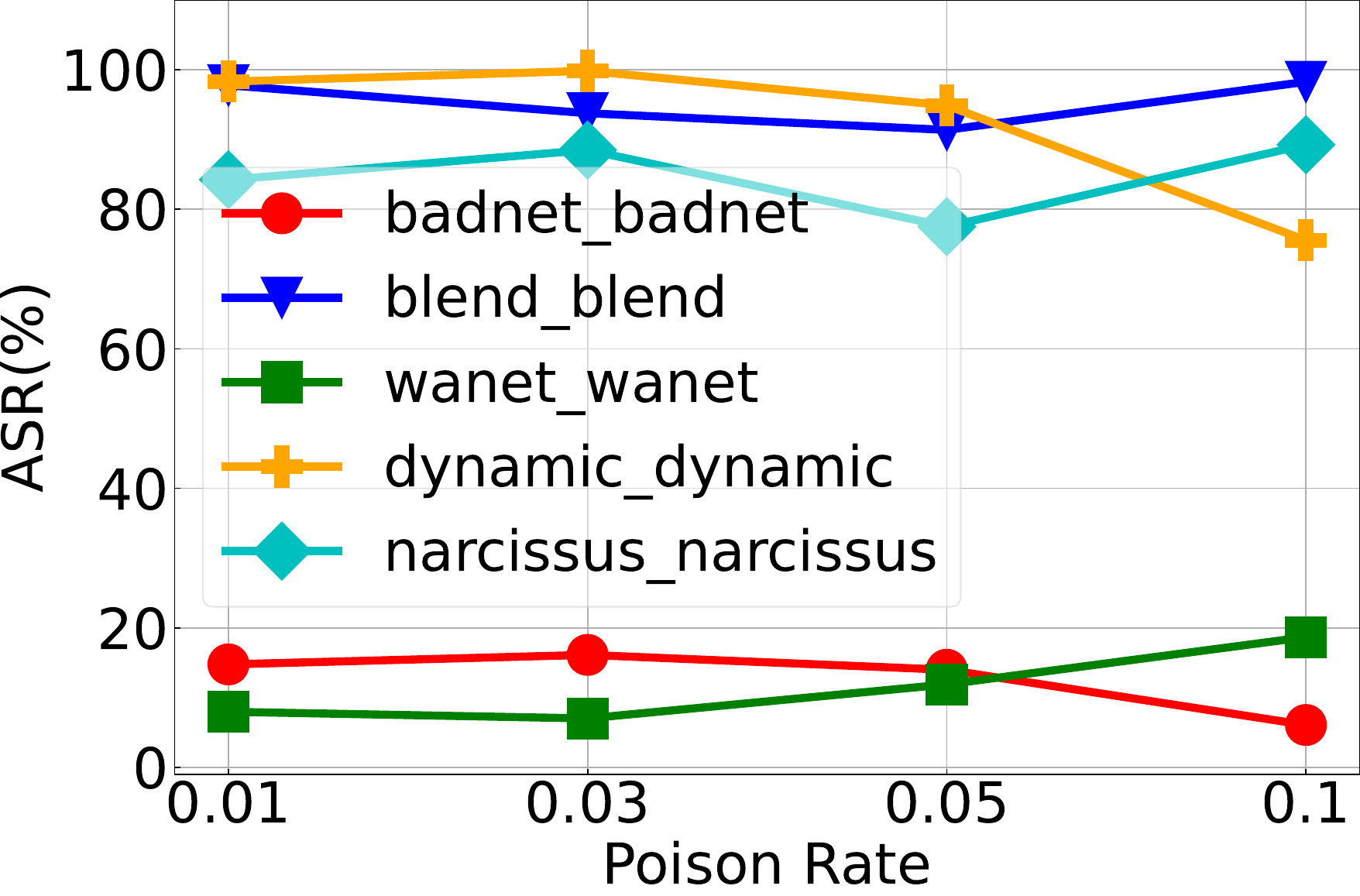}
    }
    \subfigure[Different ($\lambda=0.3$)]{\label{fig:1+3-b}
    \includegraphics[width=.21\textwidth]{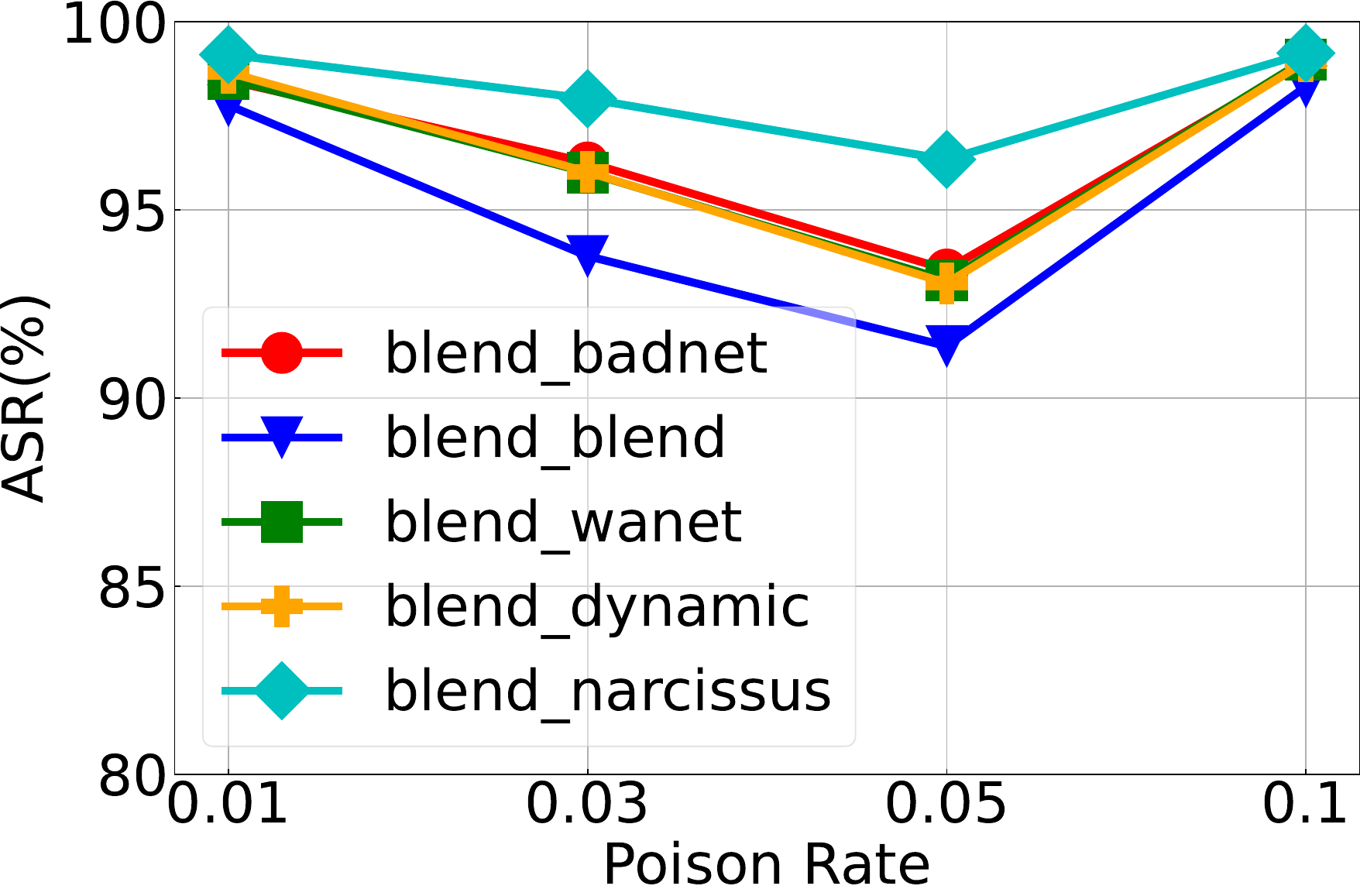}
    }
    \subfigure[Same ($\lambda=0.8$)]{\label{fig:1+3-c}
    \includegraphics[width=.21\textwidth]{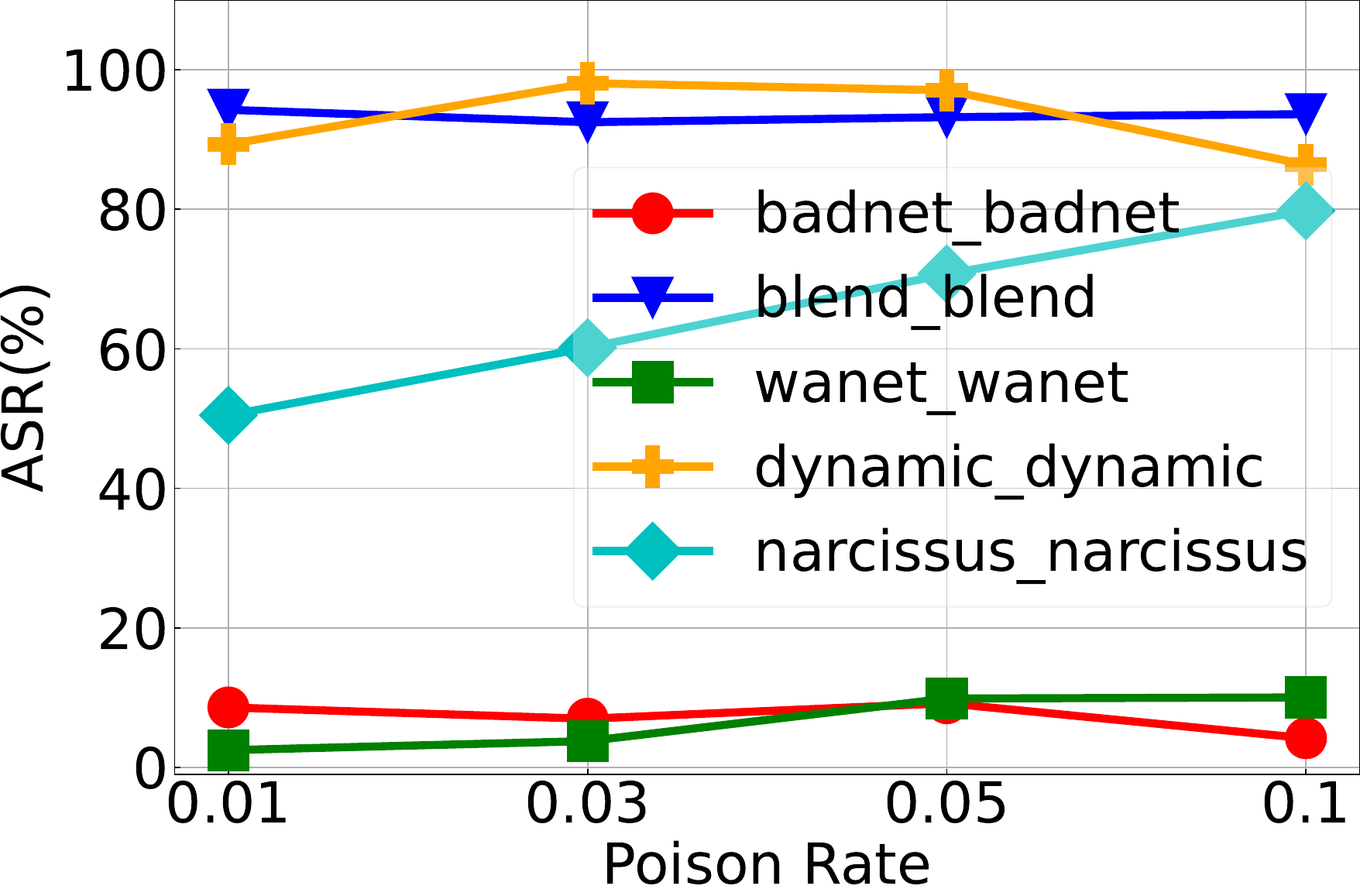}
    }
    \subfigure[Different ($\lambda=0.8$)]{\label{fig:1+3-d}
    \includegraphics[width=.21\textwidth]{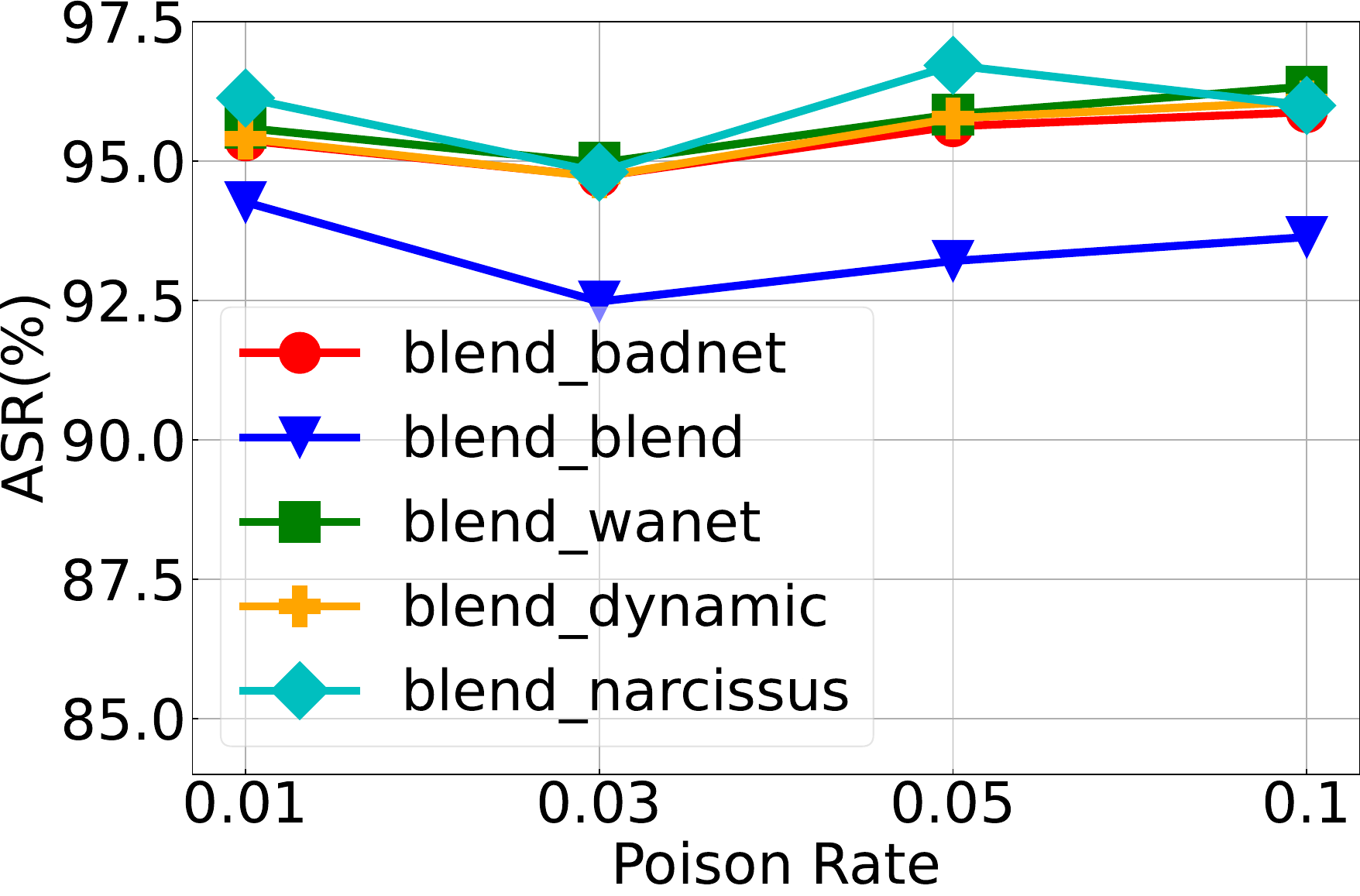}
    }
    \caption{ASR for the scenario involving a single BTV combined with three benign tasks (CIFAR100, Cars, MNIST) in the victim model.}
    \Description{}
    \label{fig:1+3}
\end{figure*}

\begin{figure*}[!ht]
    \centering
    \subfigure[Same ($\lambda=0.3$)]{\label{fig:1+7-a}
    \includegraphics[width=.21\textwidth]{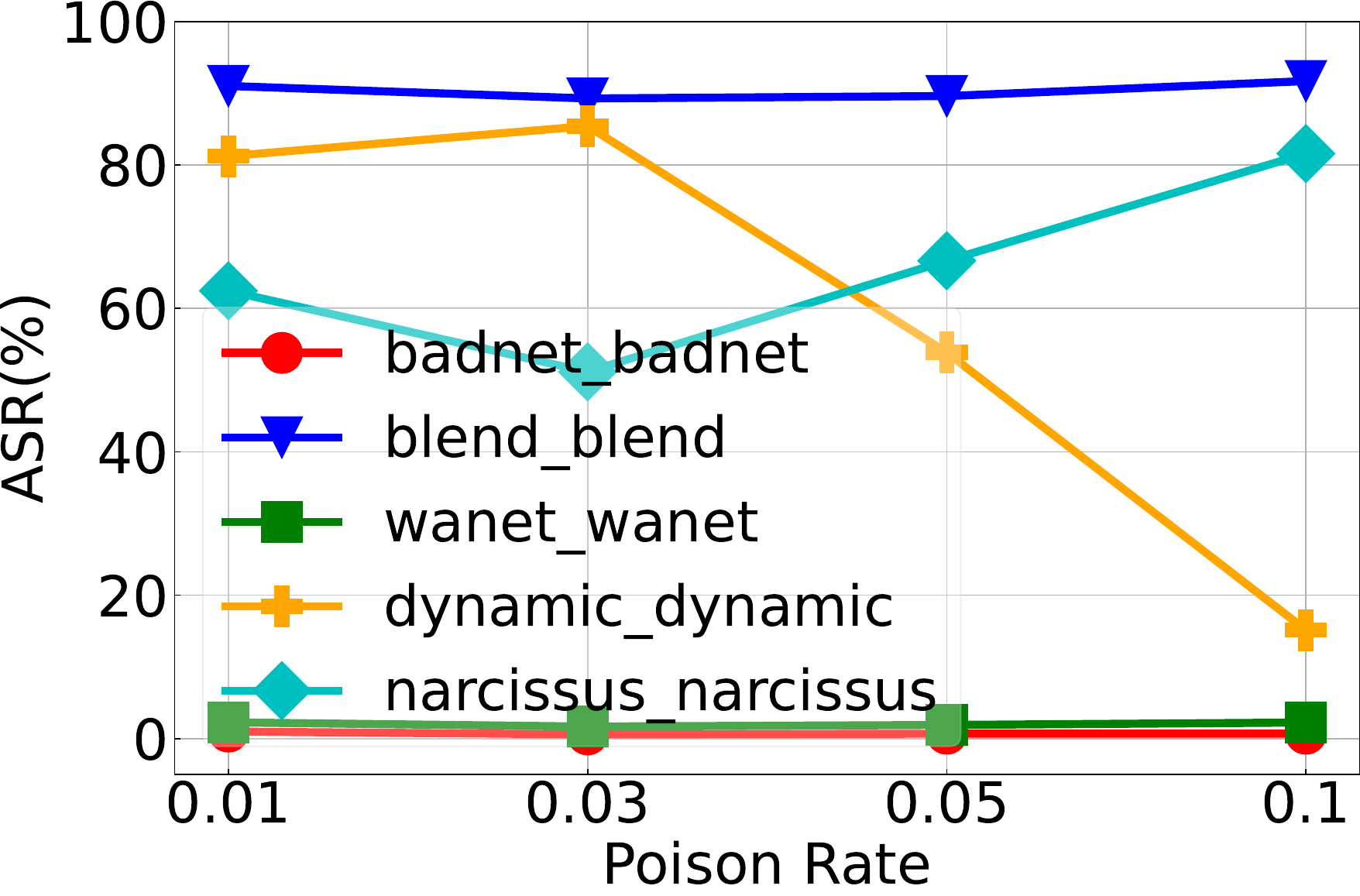}
    }
    \subfigure[Different ($\lambda=0.3$)]{\label{fig:1+7-b}
    \includegraphics[width=.21\textwidth]{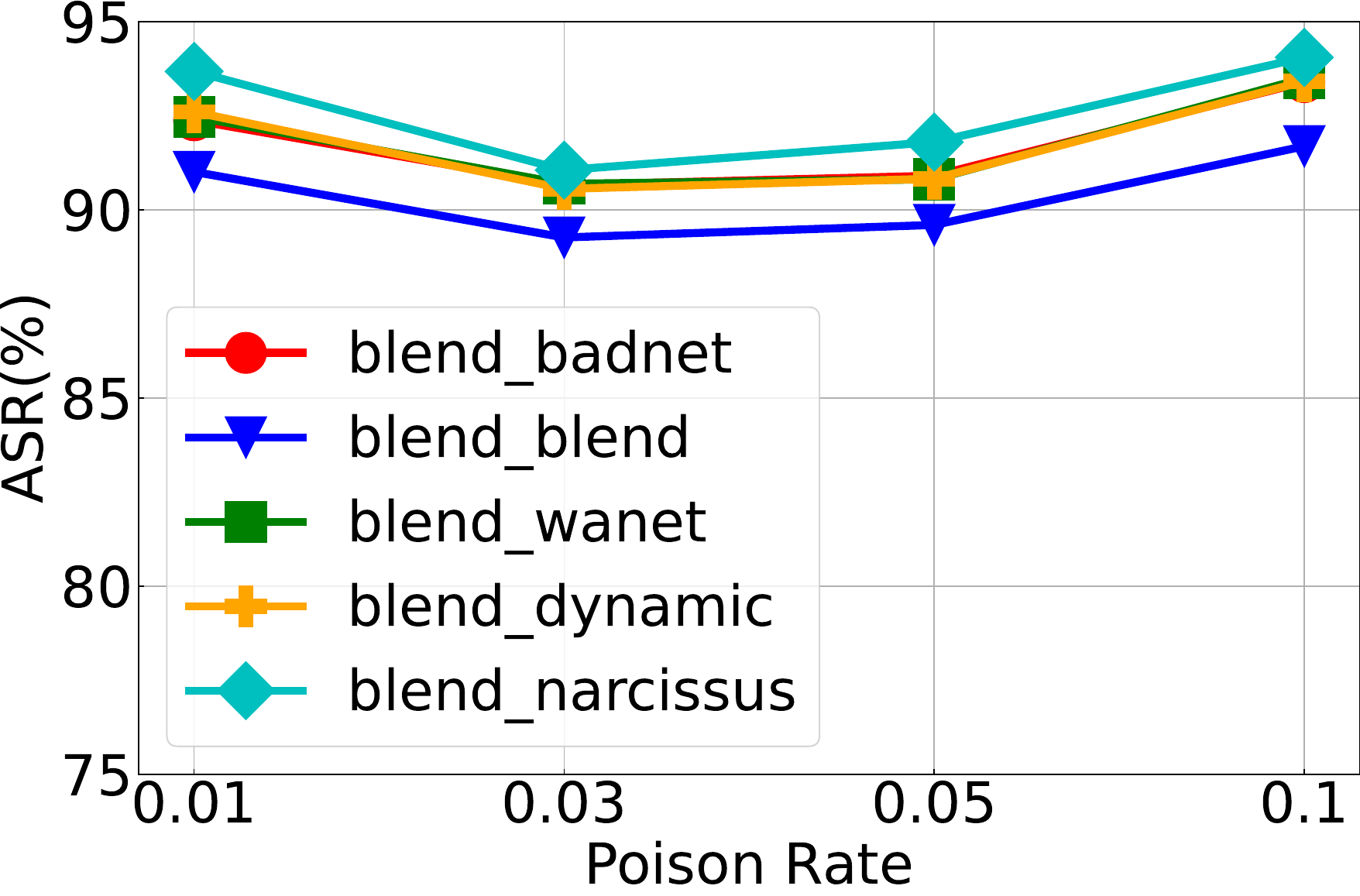}
    }
    \subfigure[Same ($\lambda=0.8$)]{\label{fig:1+7-c}
    \includegraphics[width=.21\textwidth]{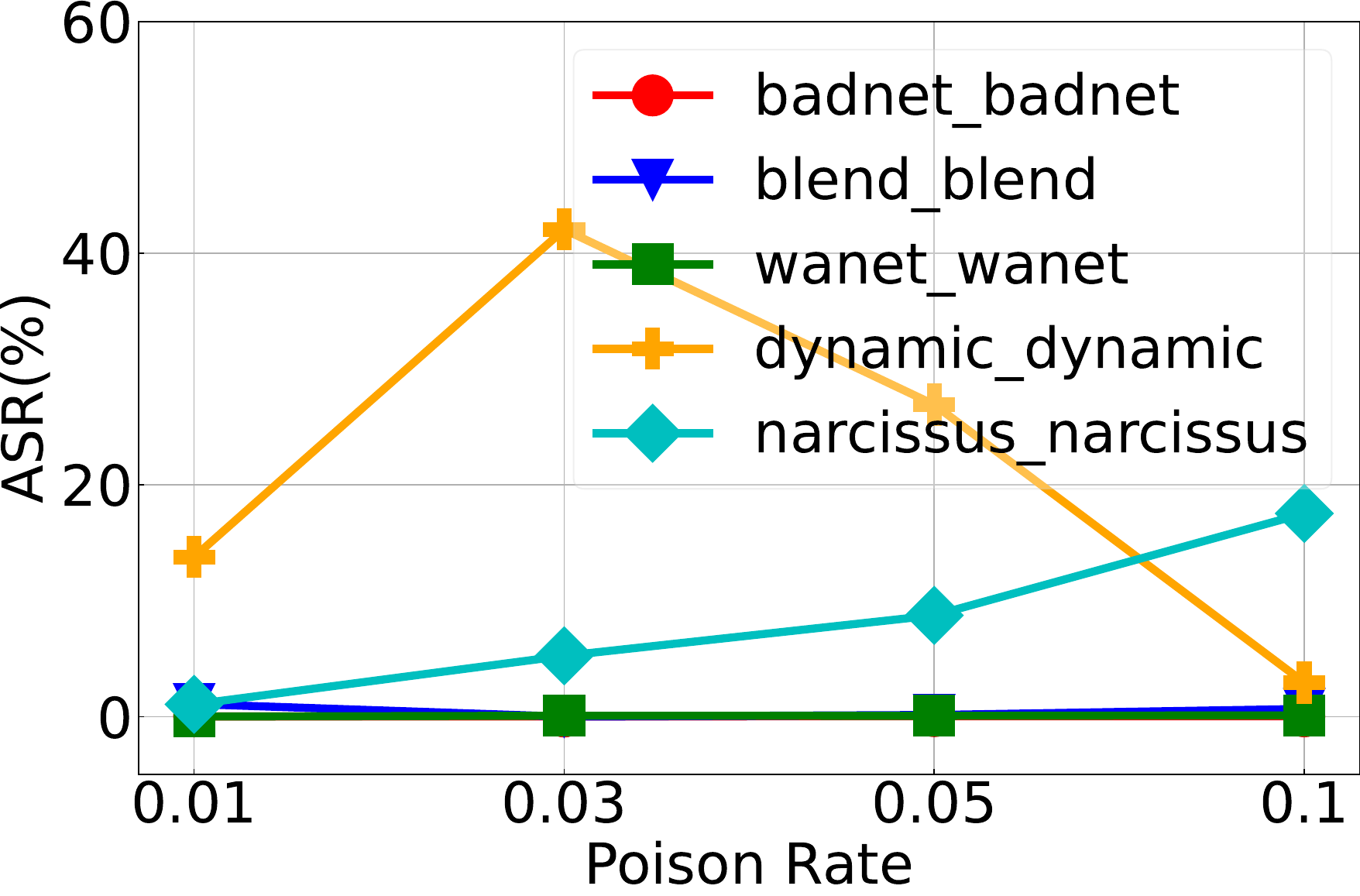}
    }
    \subfigure[Different ($\lambda=0.8$)]{\label{fig:1+7-d}
    \includegraphics[width=.21\textwidth]{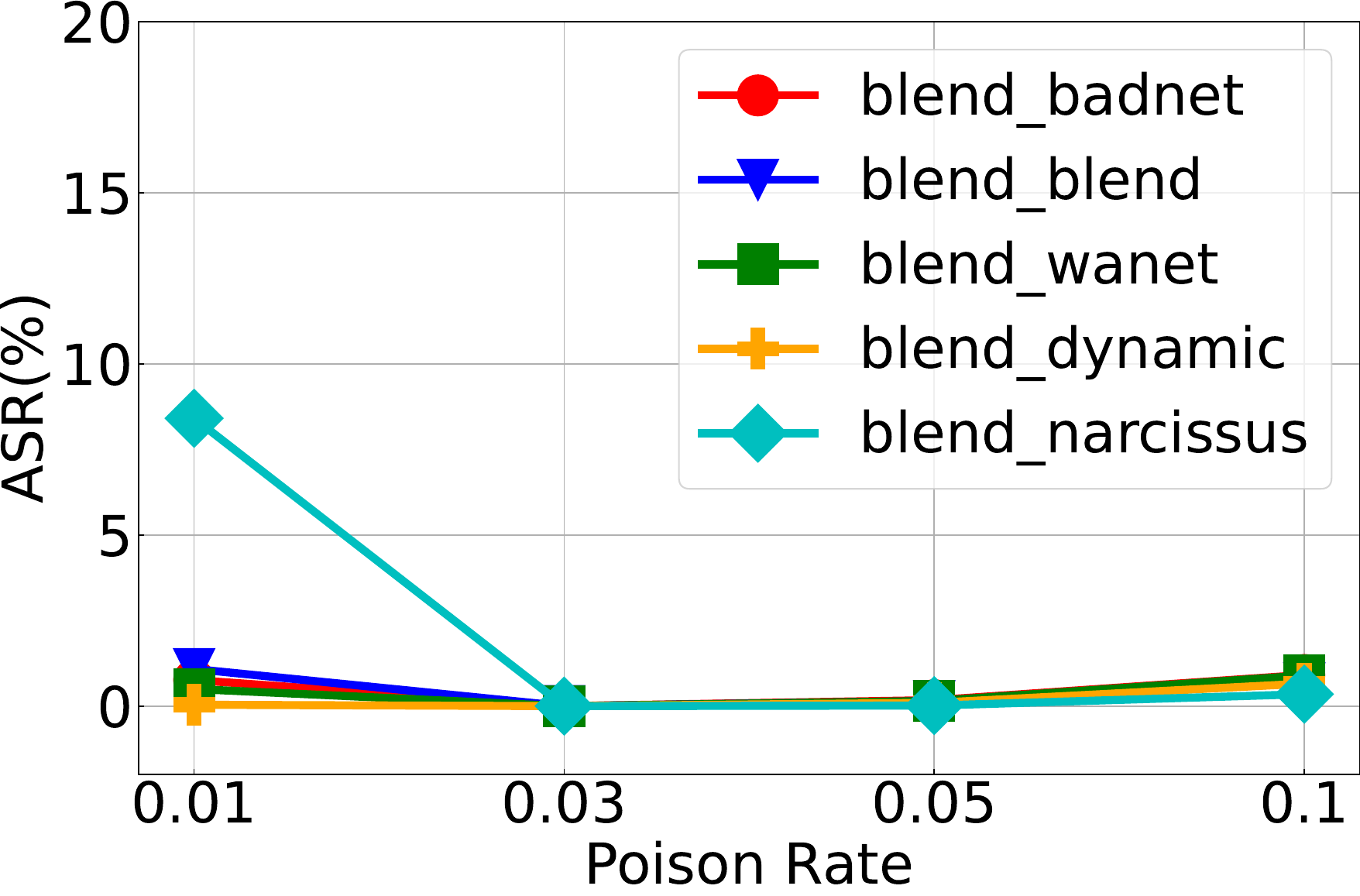}
    }
    \caption{ASR for the scenario involving a single BTV combined with seven benign tasks (CIFAR100, Cars, MNIST, SVHN, CIFAR10, EuroSAT, SUN397) in the victim model.}
    \Description{}
    \label{fig:1+7}
\end{figure*}
\begin{figure}[!t]
    \centering
    \subfigure[\fontsize{8pt}{9pt}\selectfont CTV ($\lambda=0.3$)]{\label{fig:blend_clean-a}
\includegraphics[width=0.21\textwidth]{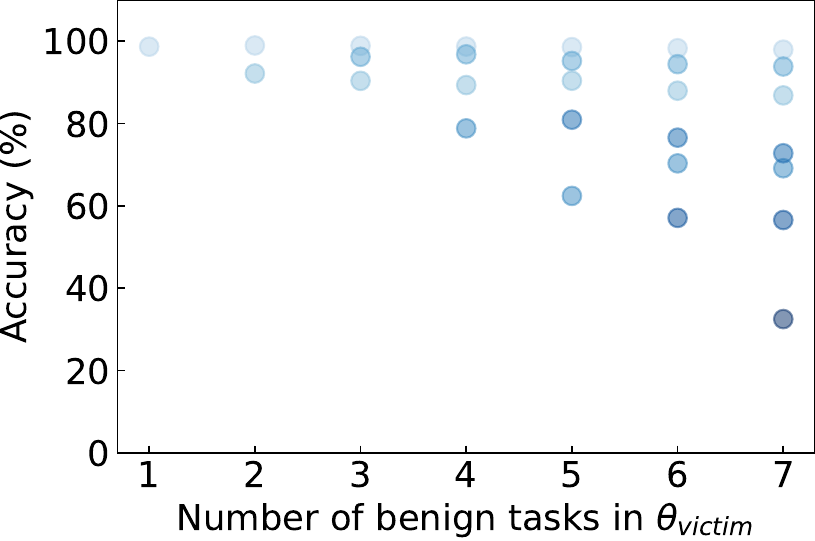}}
\subfigure[\fontsize{8pt}{9pt}\selectfont CTV ($\lambda=0.8$)]{\label{fig:blend_clean-b}
\includegraphics[width=0.21\textwidth]{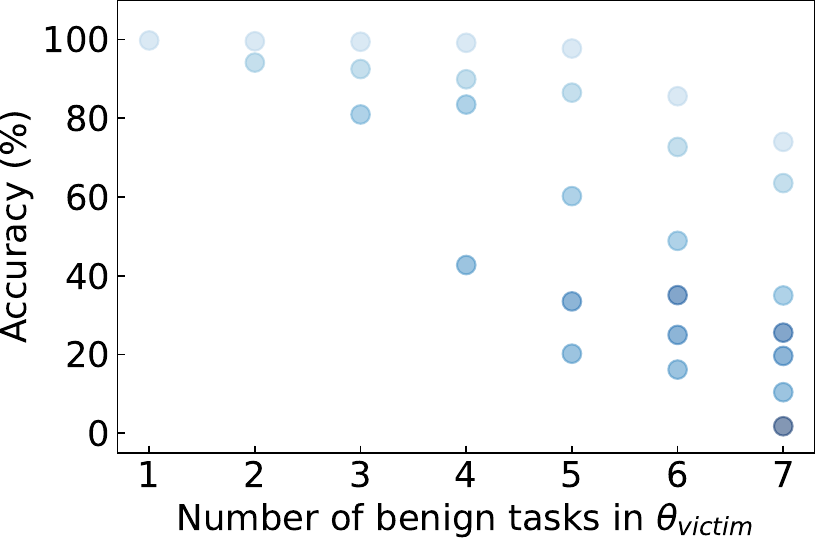}}
    \caption{Benign task accuracies of $\theta_{\text{victim}}$ with seven benign tasks (each of the $i$ dots at horizontal position $i$ represents the accuracy of one benign task).}
    \Description{}
    \label{fig:blend_clean}
\end{figure}
\begin{figure*}[t]
    \centering
    \subfigure[\fontsize{8pt}{9pt}\selectfont $t=$ MNIST, $\lambda=0.3$, same TC]{
\includegraphics[width=0.2\textwidth]{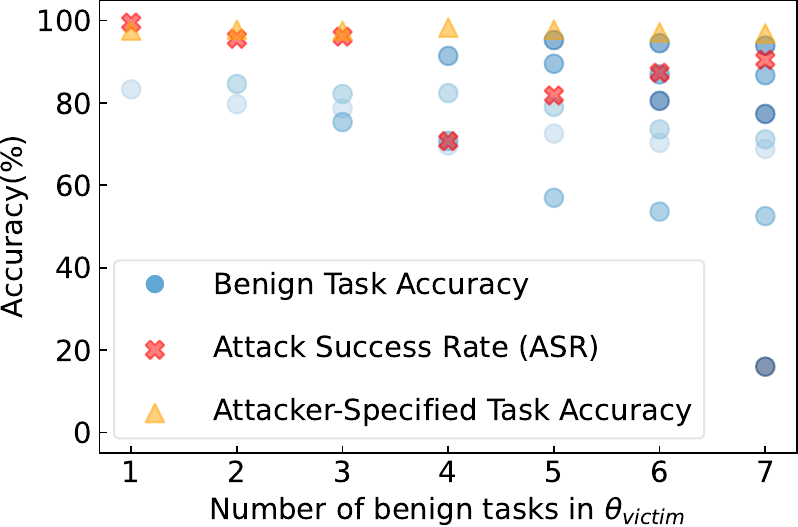}}
    \subfigure[\fontsize{8pt}{9pt}\selectfont $t=$ MNIST, $\lambda=0.3$, diff TC]{\label{fig:blend_nclean-b} \includegraphics[width=0.2\textwidth]{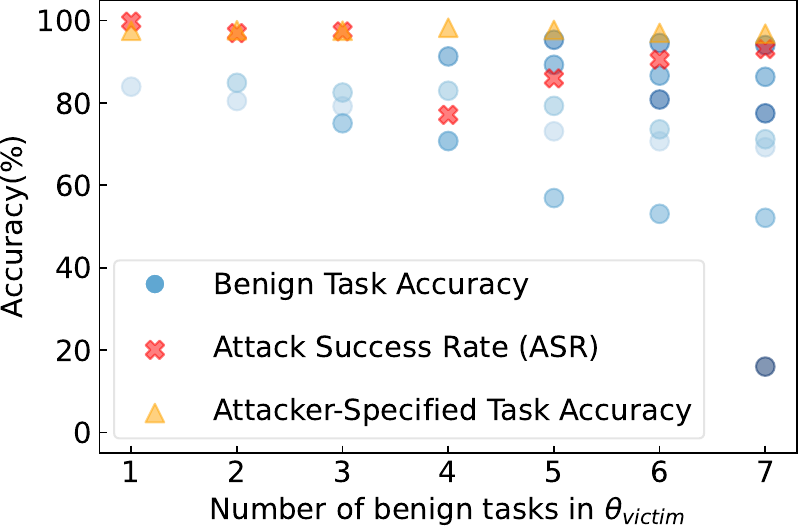}}
    \subfigure[\fontsize{8pt}{9pt}\selectfont $t=$ MNIST, $\lambda=0.8$, same TC]{
\includegraphics[width=0.2\textwidth]{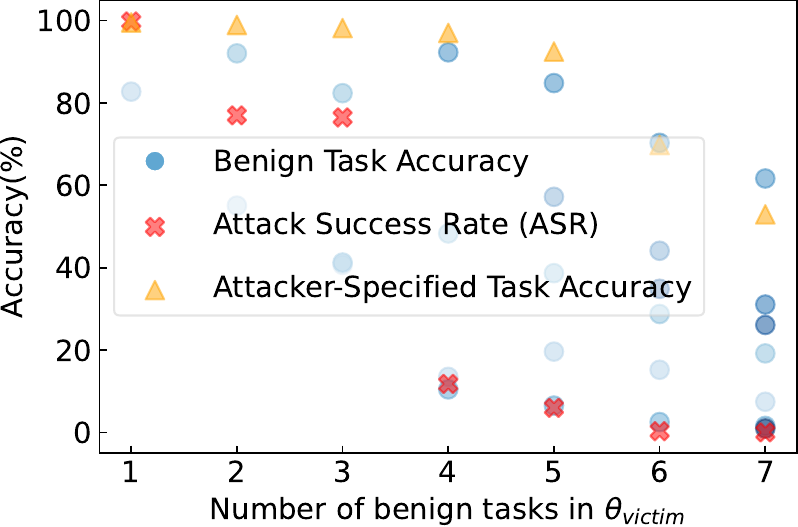}}
    \subfigure[\fontsize{8pt}{9pt}\selectfont $t=$ MNIST, $\lambda=0.8$, diff TC]{
    \includegraphics[width=0.2\textwidth]{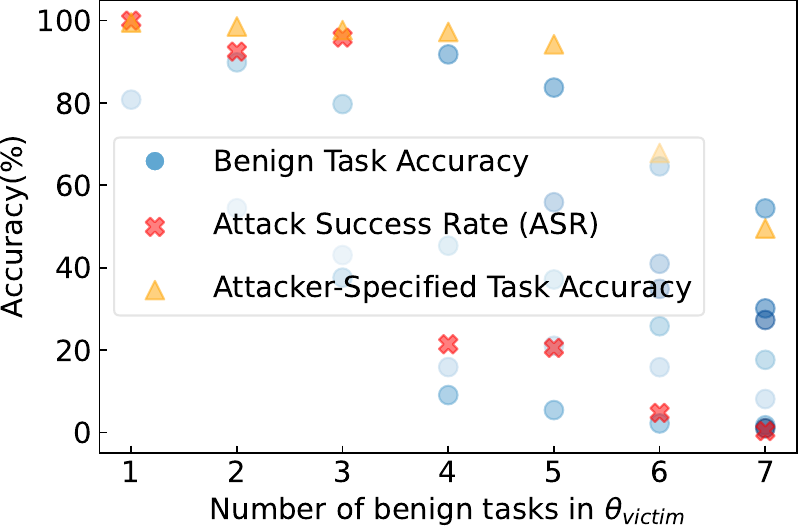}}
    \subfigure[\scriptsize $t=$ CIFAR100, $\lambda=0.3$, same TC]{
    \includegraphics[width=0.2\textwidth]{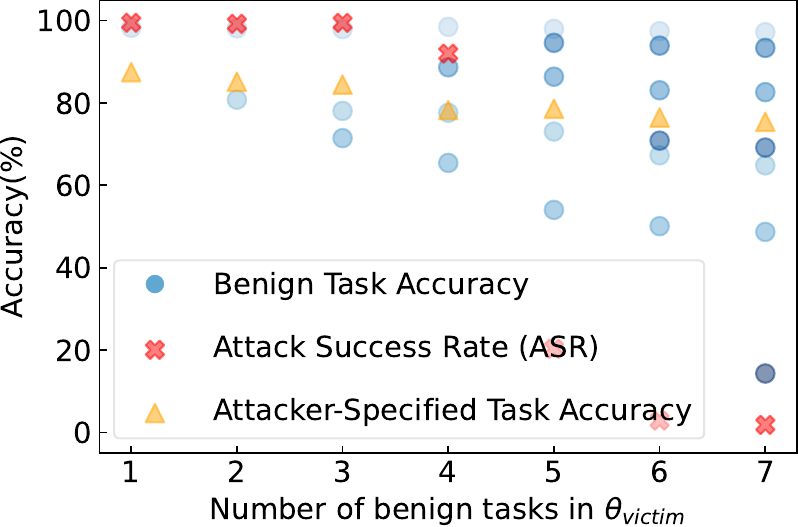}}
    \subfigure[\scriptsize $t=$ CIFAR100, $\lambda=0.3$, diff TC]{
\label{fig:blend_nclean-f}    \includegraphics[width=0.2\textwidth]{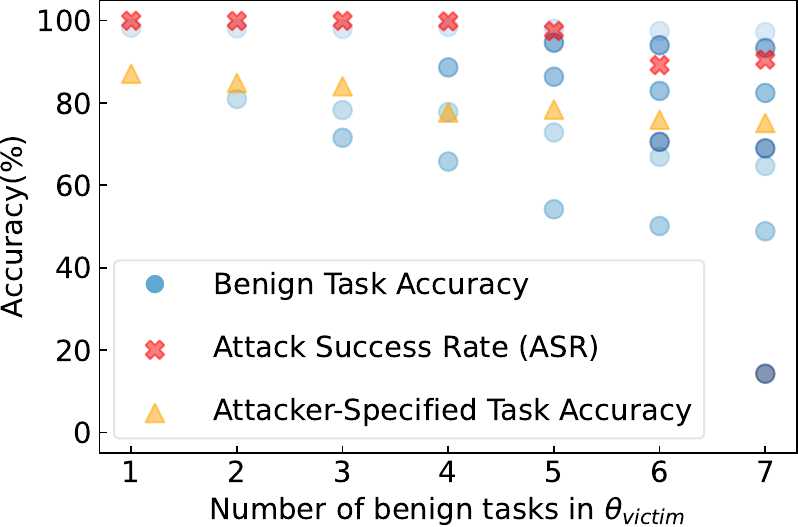}}
    \subfigure[\scriptsize $t=$ CIFAR100, $\lambda=0.8$, same TC]{
    \includegraphics[width=0.2\textwidth]{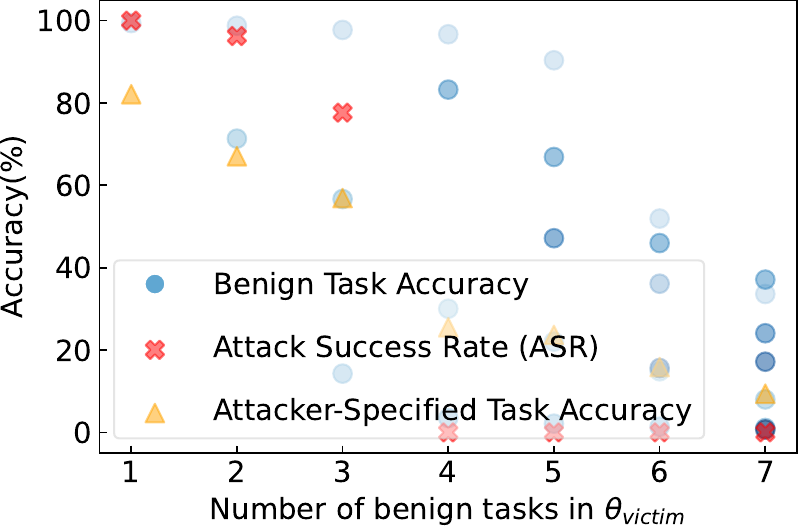}}
    \subfigure[\fontsize{7pt}{8pt}\selectfont $t=$ CIFAR100, $\lambda=0.8$, diff TC]{
    \includegraphics[width=0.2\textwidth]{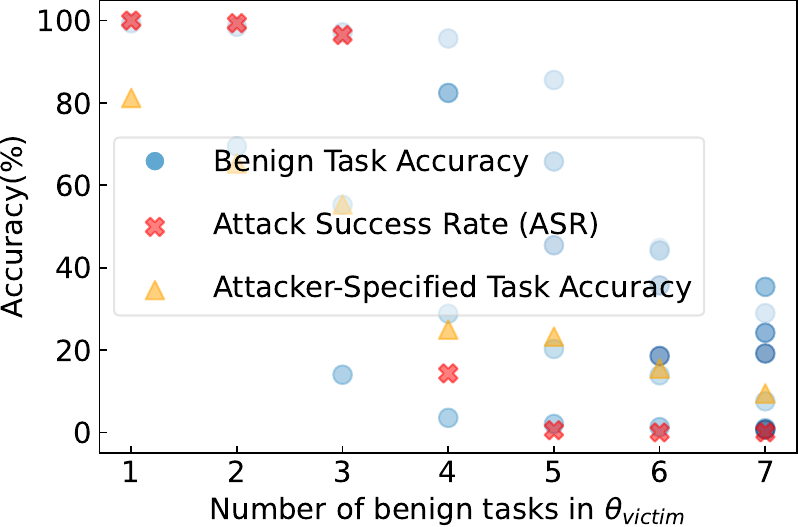}}
    \caption{Benign task accuracies, ASR, attacker-specified task accuracies under the realistic setting with seven benign tasks TC denotes \textit{target class}. Each of the $i$ blue dots at horizontal position $i$ represents the accuracy of one benign task.
}
    \Description{}
    \label{fig:blend_nclean}
\end{figure*}
Similar to \S\ref{subsec:badtv_factors}, all experiments in this section are conducted under the \textit{realistic} setting, with the attacker-specified task $t$ as GTSRB and the user-chosen task $t_1$ as CIFAR100. For model hijacking, we set the original task as EuroSAT and the hijacking task as SVHN.

\begin{figure*}[hbt!]
    \centering
    \vspace{-4pt}
    \subfigure[Backdoor ($\lambda=0.3$)]{
    \includegraphics[width=.2\textwidth]{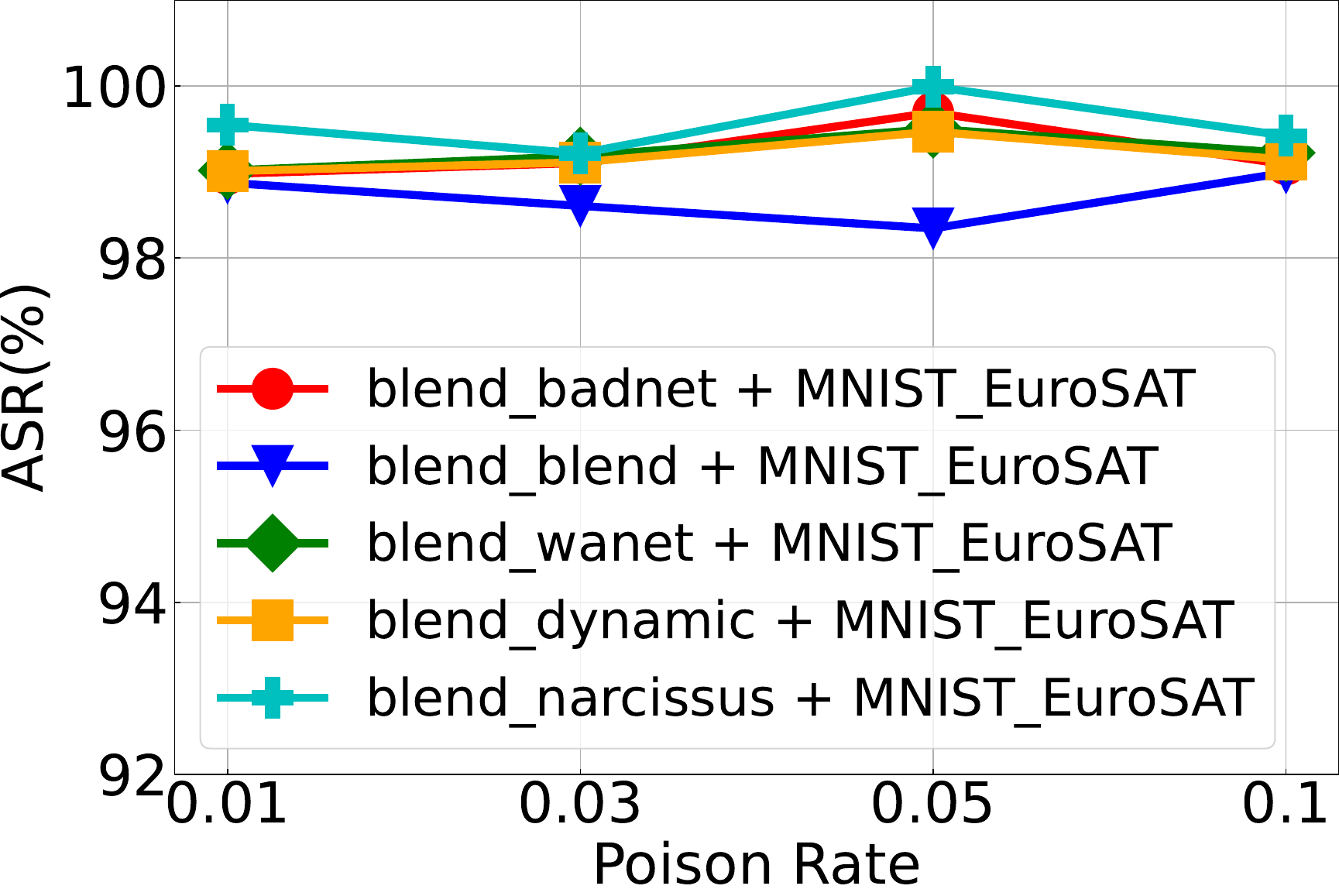}\label{fig:2+1_hijack_b_0.3}
    }
    \subfigure[Hijacking ($\lambda=0.3$)]{
    \includegraphics[width=.2\textwidth]{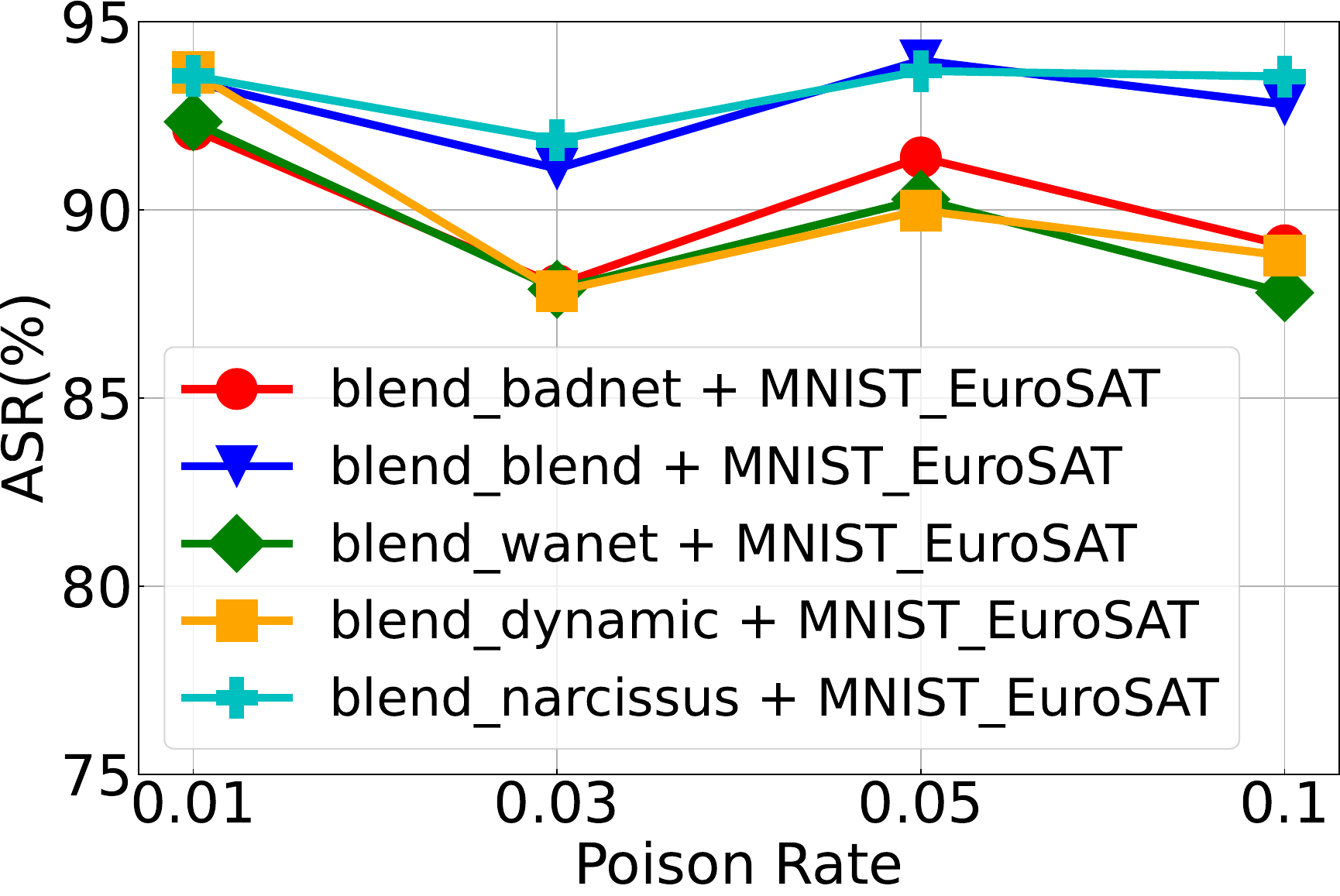}\label{fig:2+1_hijack_h_0.3}
    }
    \subfigure[Backdoor ($\lambda=0.8$)]{
    \includegraphics[width=.2\textwidth]{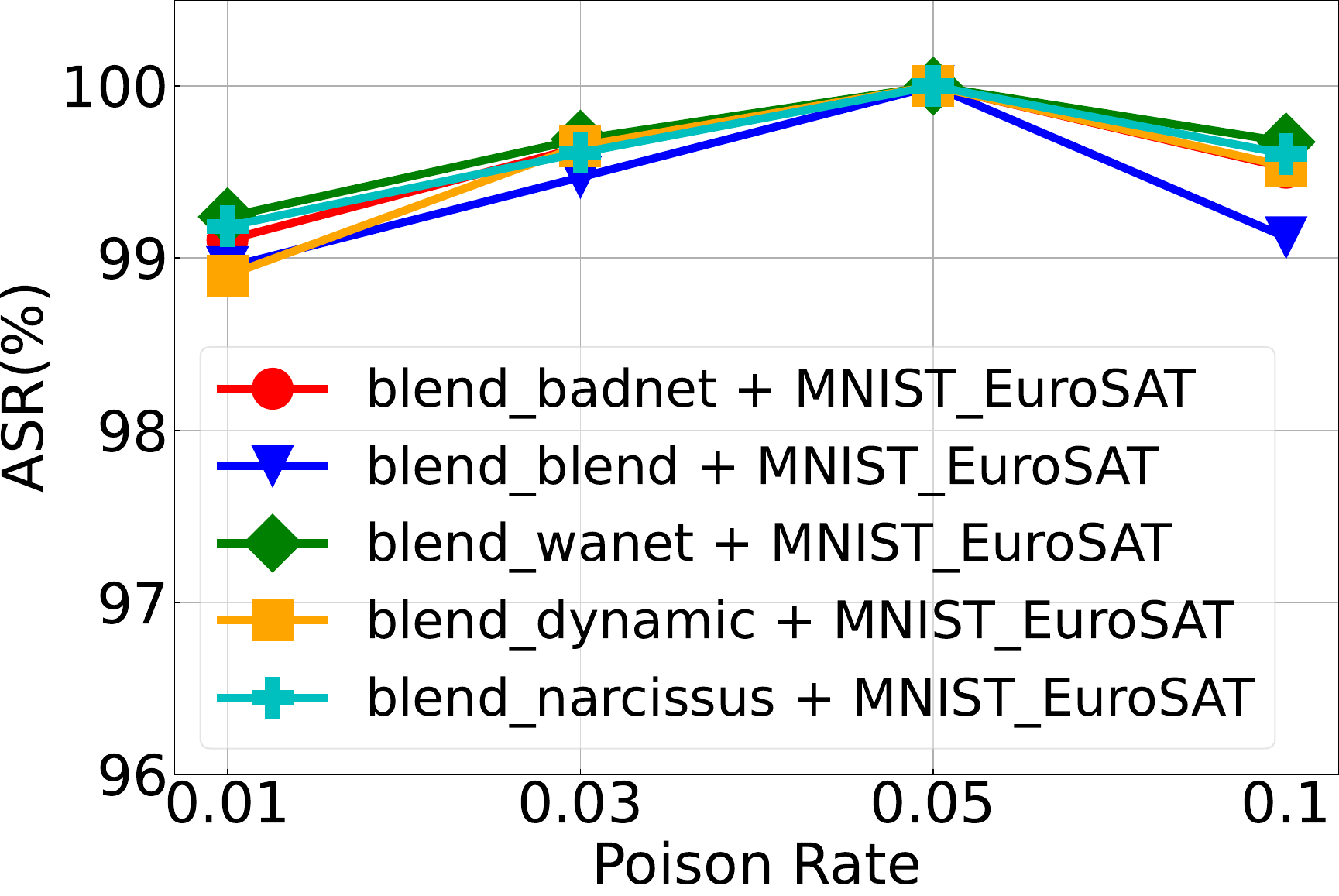}\label{fig:2+1_hijack_b_0.8}
    }
    \subfigure[Hijacking ($\lambda=0.8$)]{
    \includegraphics[width=.2\textwidth]{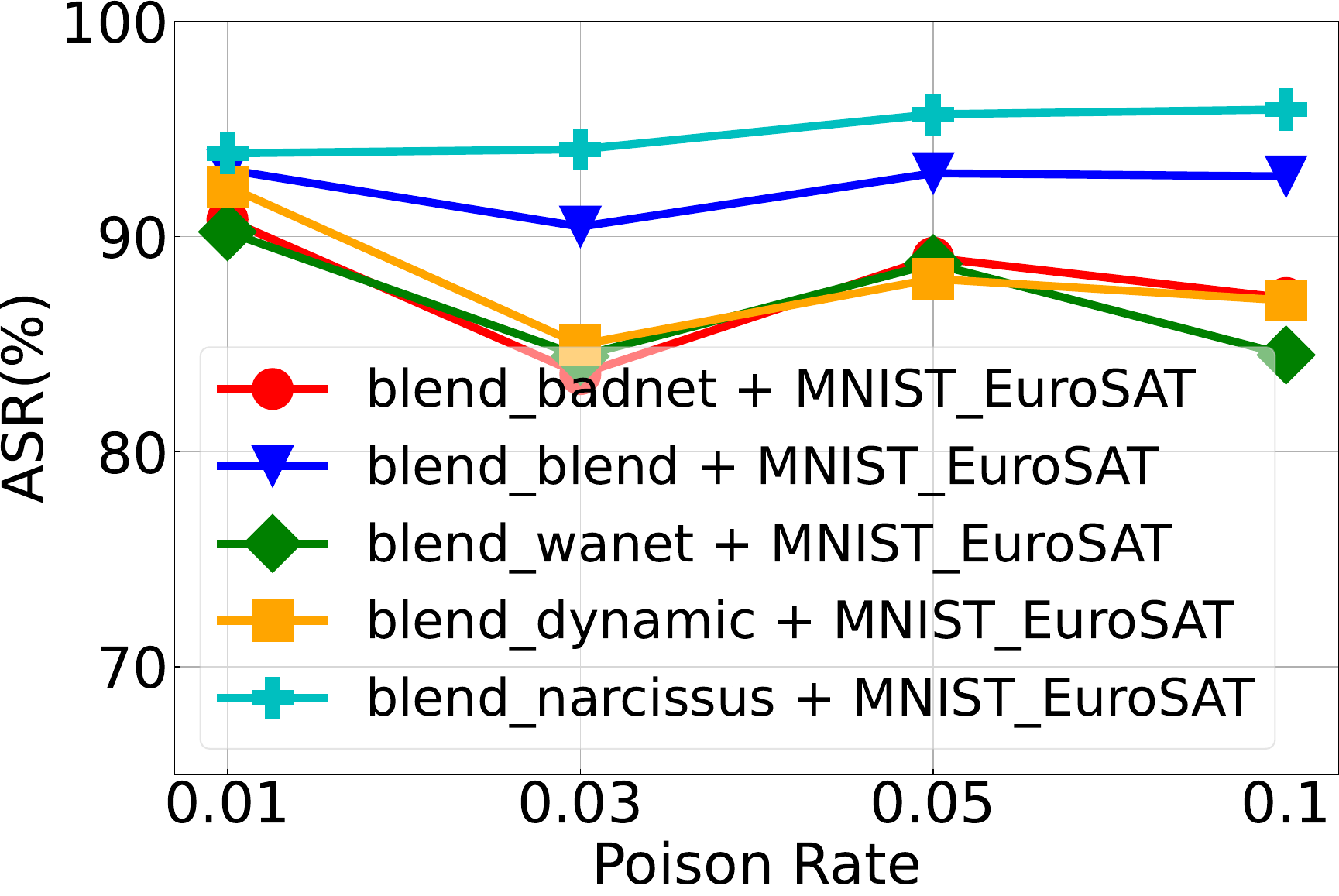}\label{fig:2+1_hijack_h_0.8}
    }
    \caption{ASR for the scenario involving two BTVs, one for a backdoor attack and the other for a model hijacking attack.}
    \label{fig:2+1_hijack}
    \Description{}
\end{figure*}

\subsubsection{More Benign Tasks in Victim Model}\label{subsubsec:multiCTV}
In \S\ref{subsec:main_result} and \S\ref{subsec:badtv_factors}, the victim model $\theta_{\text{victim}}$ in the realistic setting includes only one benign task $t_1$. However, users may deploy multi-task models in practice. Hence, we first examine a scenario in which the victim model contains three benign tasks: CIFAR100 ($t_1$), Cars ($t_2$), and MNIST ($t_3$), to assess the potential impact on \textsc{BadTV}'s ASR.

Figures~\ref{fig:1+3-a} and \ref{fig:1+3-c} indicate reduced ASR in \textcolor{black}{BadNets and WaNet because these attacks use weaker or more localized backdoor patterns. Under the same setting as Figure~\ref{fig:1+3-a}, strengthening the attack configuration (e.g., larger triggers or stronger warping) also improves ASR (13.99 → 49.8 and 11.89 → 55.77), while stronger or global attacks such as Blend remain more stable. In contrast}, Figures~\ref{fig:1+3-b} and \ref{fig:1+3-d} show that employing distinct backdoor attack types for $b_1$ and $b_2$ enhances \textsc{BadTV}'s robustness, even with multiple benign tasks embedded. This observation aligns with findings from \S\ref{sec:why the setting of different target classes} and \S\ref{subsubsec:backdoor_combo}, confirming that diverse attack combinations strengthen ASR.


We further investigate a scenario involving seven benign tasks: CIFAR100 ($t_1$), Cars ($t_2$), MNIST ($t_3$), SVHN ($t_4$), CIFAR10 ($t_5$), EuroSAT ($t_6$), and SUN397 ($t_7$), with results illustrated in Figure~\ref{fig:1+7}. Interestingly, we observe notably different behavior compared to Figure~\ref{fig:1+3}. Specifically, Figure~\ref{fig:1+7} shows a significant reduction in ASR except for Figure~\ref{fig:1+7-b}, suggesting that increasing the number of benign tasks dilutes the backdoor signal. Even when employing different backdoor types for $b_1$ and $b_2$, Figure~\ref{fig:1+7-d} reveals substantial ASR degradation. Nonetheless, since $\lambda$ is user-controlled and not dictated by the attacker, attackers cannot reliably expect users to consistently select low $\lambda$ values. Thus, it naturally raises the question of if this scenario (users adopting higher $\lambda$ while including more benign tasks) could serve as a defense against \textsc{BadTV}. \textcolor{black}{In addition, we evaluated different insertion positions of the BTV (e.g., early, middle, and late in the merging sequence) and observed identical results to those in Figure~\ref{fig:1+7}. This follows from the commutativity of task vector addition, which makes the merged model invariant to the ordering of task vectors.}

We start by examining the benign task accuracies of the victim model containing seven tasks. As shown in Figure~\ref{fig:blend_clean-b}, even in the absence of attacks, incorporating more than four benign tasks at high $\lambda$ drastically reduces individual task accuracies, providing users little incentive to adopt such settings. Consequently, users should either opt for more benign tasks at lower $\lambda$ or fewer tasks at higher $\lambda$. In both cases, \textsc{BadTV} remains effective, maintaining reasonably high ASR.

An important question remains: Will the benign tasks' accuracies in the merged model $\hat{M}_{\text{merged}}=M_{\theta_{\text{victim}}\oplus \lambda\hat{\tau}_t}$ and the attacker-specified task accuracy degrade significantly? Figures~\ref{fig:blend_nclean-b} and \ref{fig:blend_nclean-f} indicate that the attacker-specified task accuracy remains high under attacker-preferred settings (different target classes and lower $\lambda$). Moreover, despite certain benign task accuracies experiencing declines, the accuracy distribution from $\hat{M}_{\text{merged}}$ closely matches that of $M_{\text{victim}}$, as observed by comparing Figure~\ref{fig:blend_nclean-b} (or Figure~\ref{fig:blend_nclean-f}) to Figure~\ref{fig:blend_clean-a}. Therefore, employing BTV in such contexts is unlikely to raise user suspicion.
 

\begin{figure*}
    \centering
    \subfigure[\textbf{BD}\_Hijack ($\lambda=0.3$, Addition)]{
    \includegraphics[width=.2\textwidth]{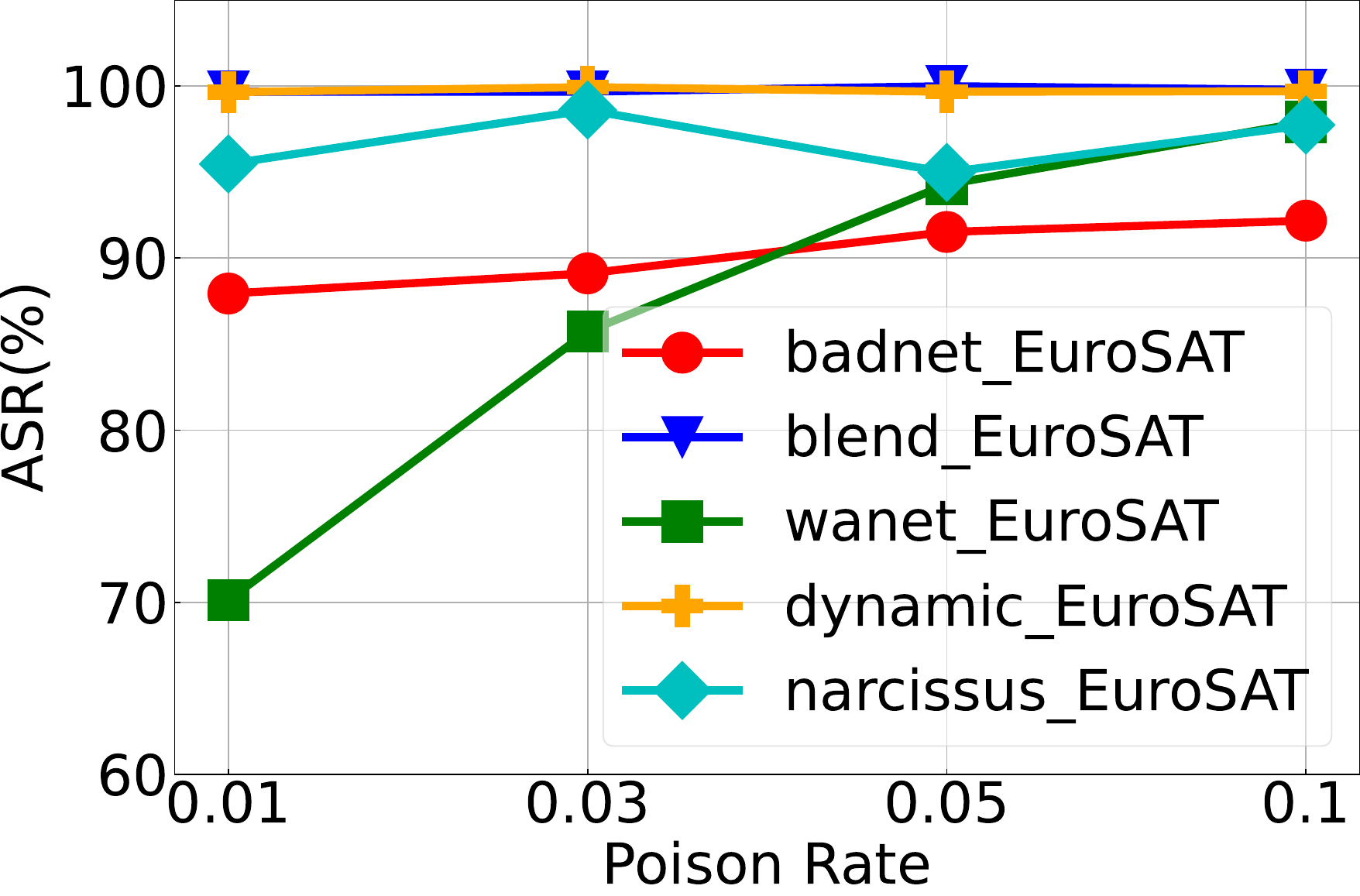}
    \label{fig:bd_hijack_a}}
    \subfigure[\scriptsize BD\_\textbf{Hijack} ($\lambda=0.3$, Subtraction)]{
    \includegraphics[width=.2\textwidth]{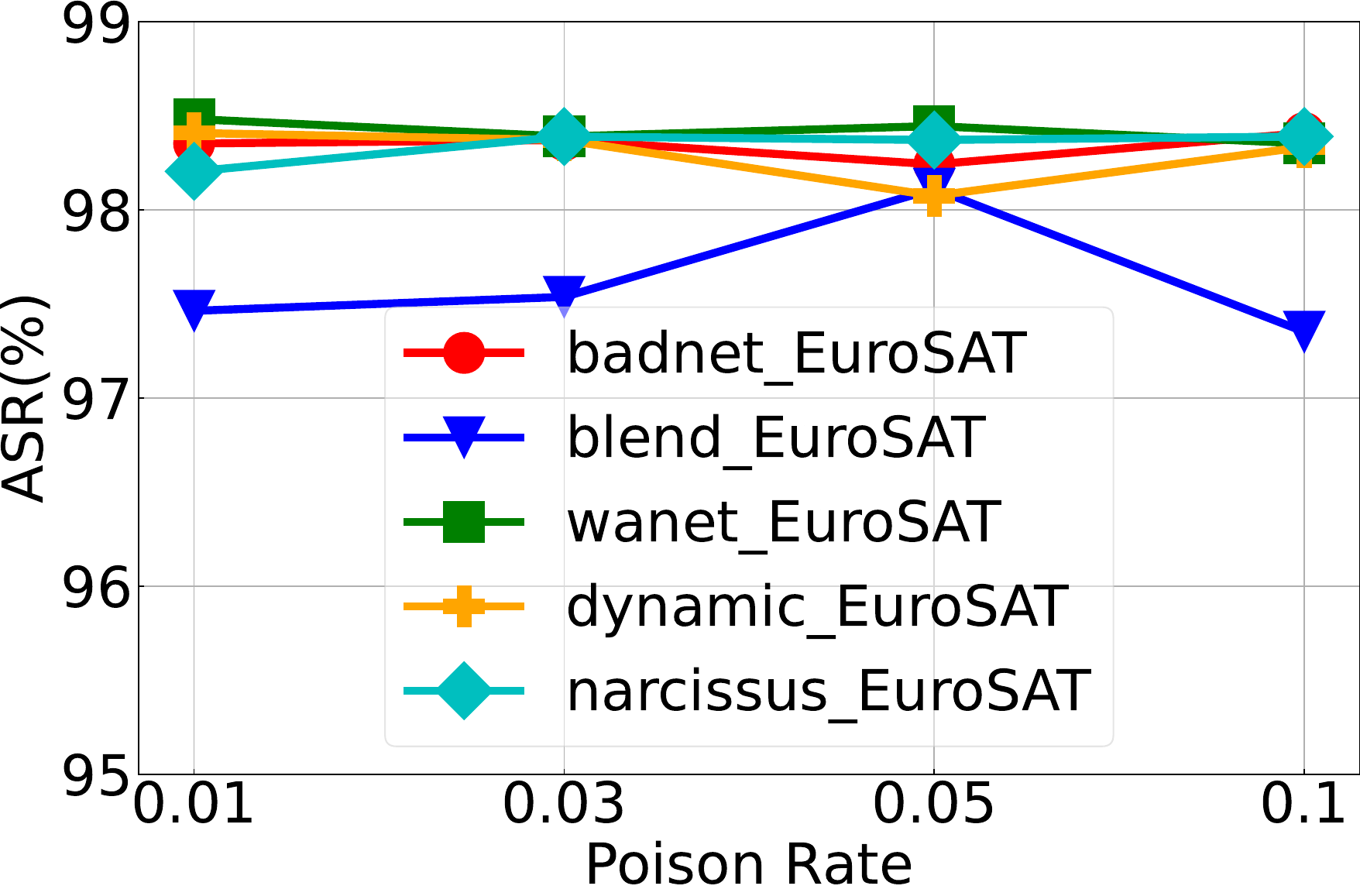}
    \label{fig:bd_hijack_b}
    }
    \subfigure[\textbf{Hijack}\_BD ($\lambda=0.3$, Addition)]{
    \includegraphics[width=.2\textwidth]{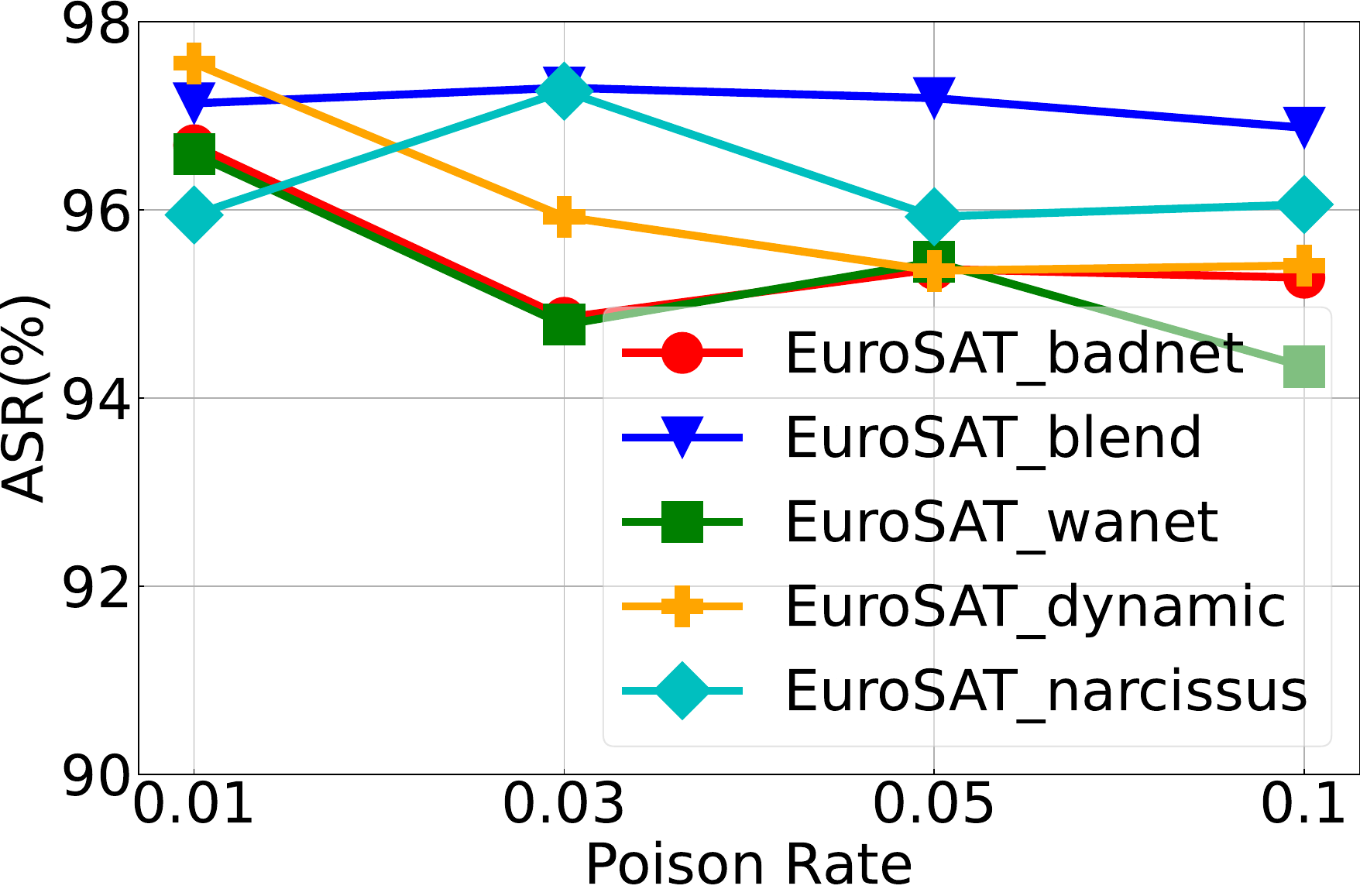}
    \label{fig:bd_hijack_c}
    }
    \subfigure[\scriptsize Hijack\_\textbf{BD} ($\lambda=0.3$, Subtraction)]{
    \includegraphics[width=.2\textwidth]{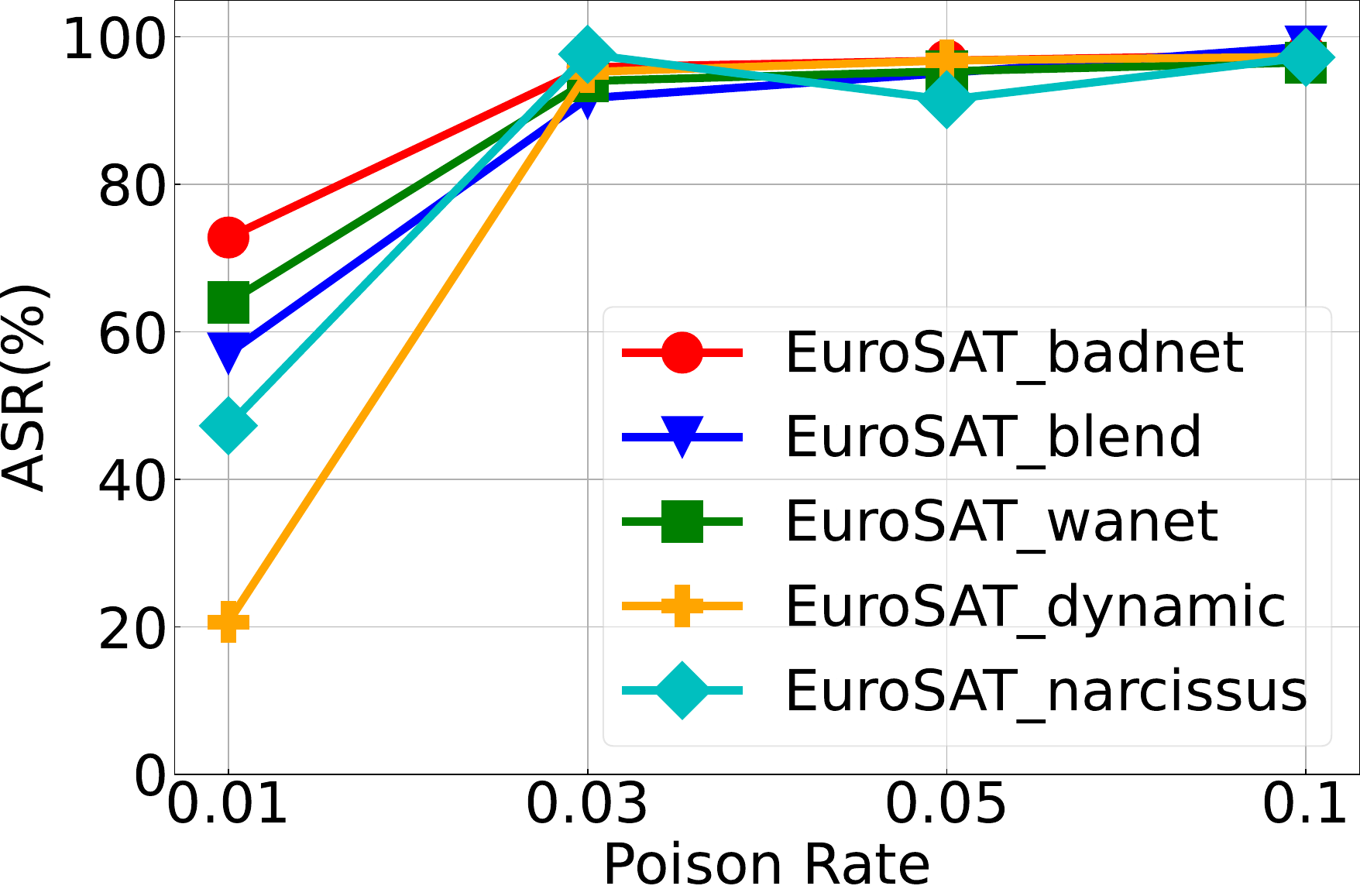}
    \label{fig:bd_hijack_d}
    }
    \subfigure[\textbf{BD}\_Hijack ($\lambda=0.8$, Addition)]{
    \includegraphics[width=.2\textwidth]{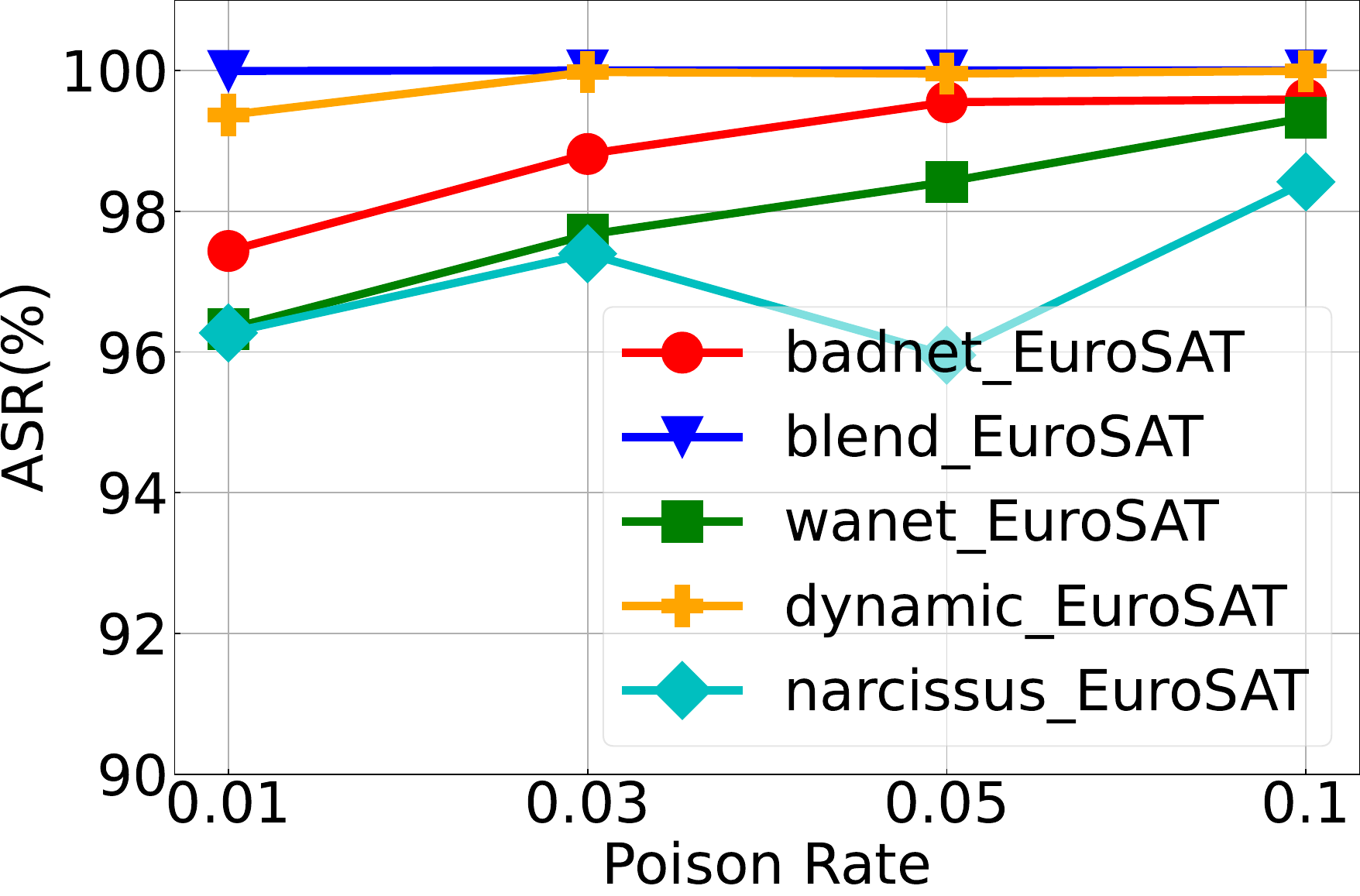}
    \label{fig:bd_hijack_e}
    }
    \subfigure[\scriptsize BD\_\textbf{Hijack} ($\lambda=0.8$, Subtraction)]{
    \includegraphics[width=.2\textwidth]{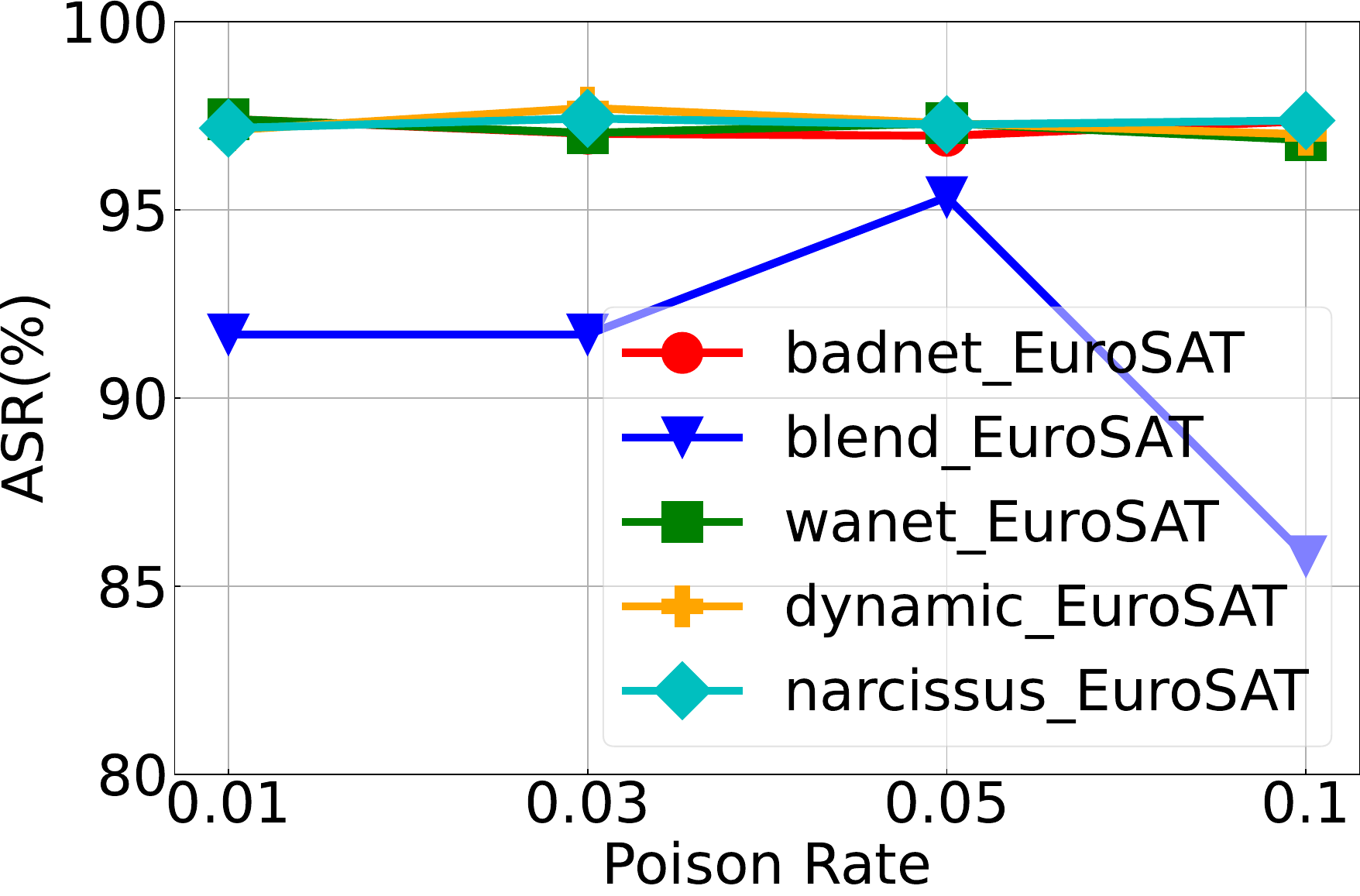}
    \label{fig:bd_hijack_f}
    }
    \subfigure[\textbf{Hijack}\_BD ($\lambda=0.8$, Addition)]{
    \includegraphics[width=.2\textwidth]{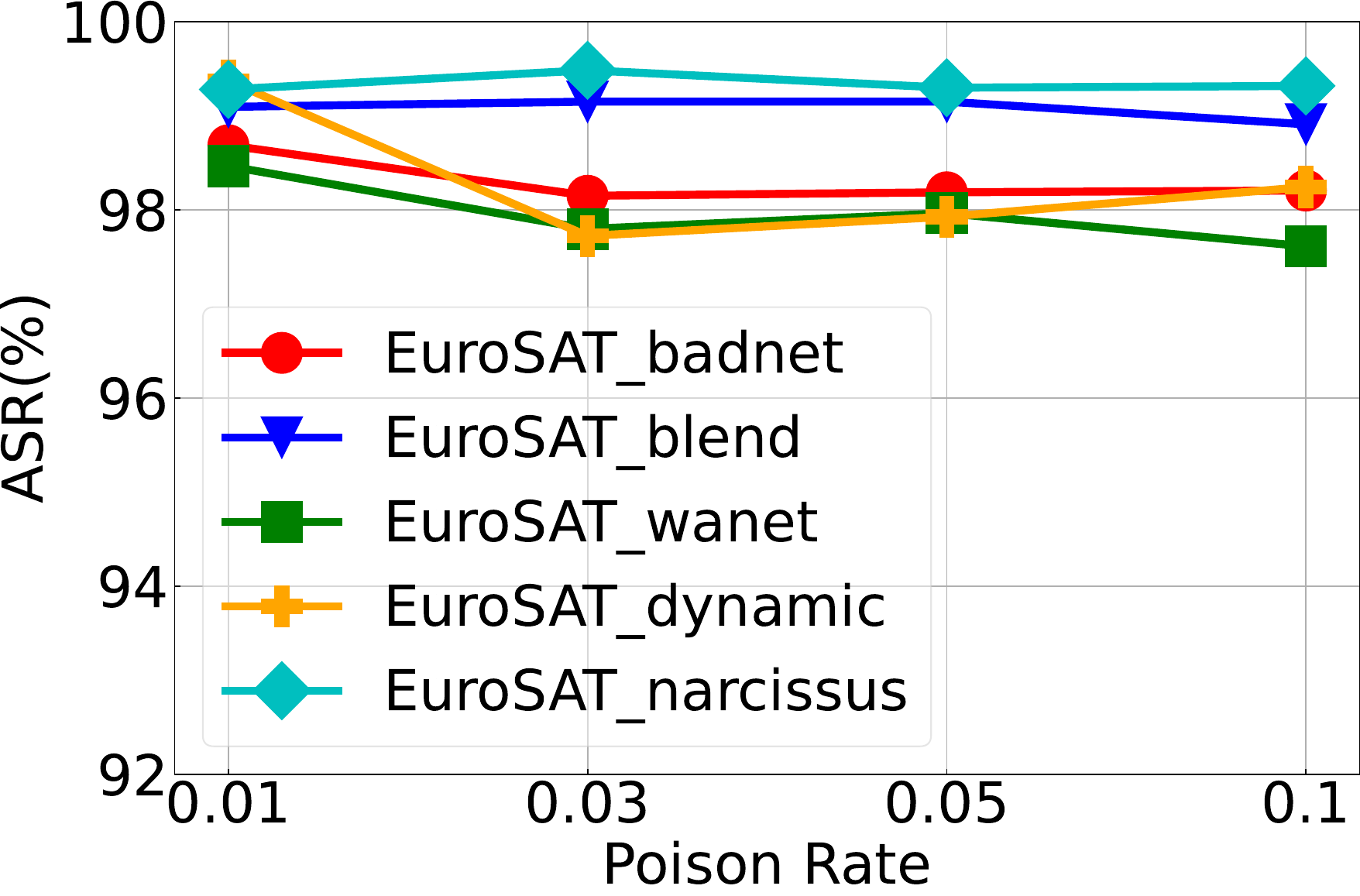}
    \label{fig:bd_hijack_g}
    }
    \subfigure[\scriptsize Hijack\_\textbf{BD} ($\lambda=0.8$, Subtraction)]{
    \includegraphics[width=.2\textwidth]{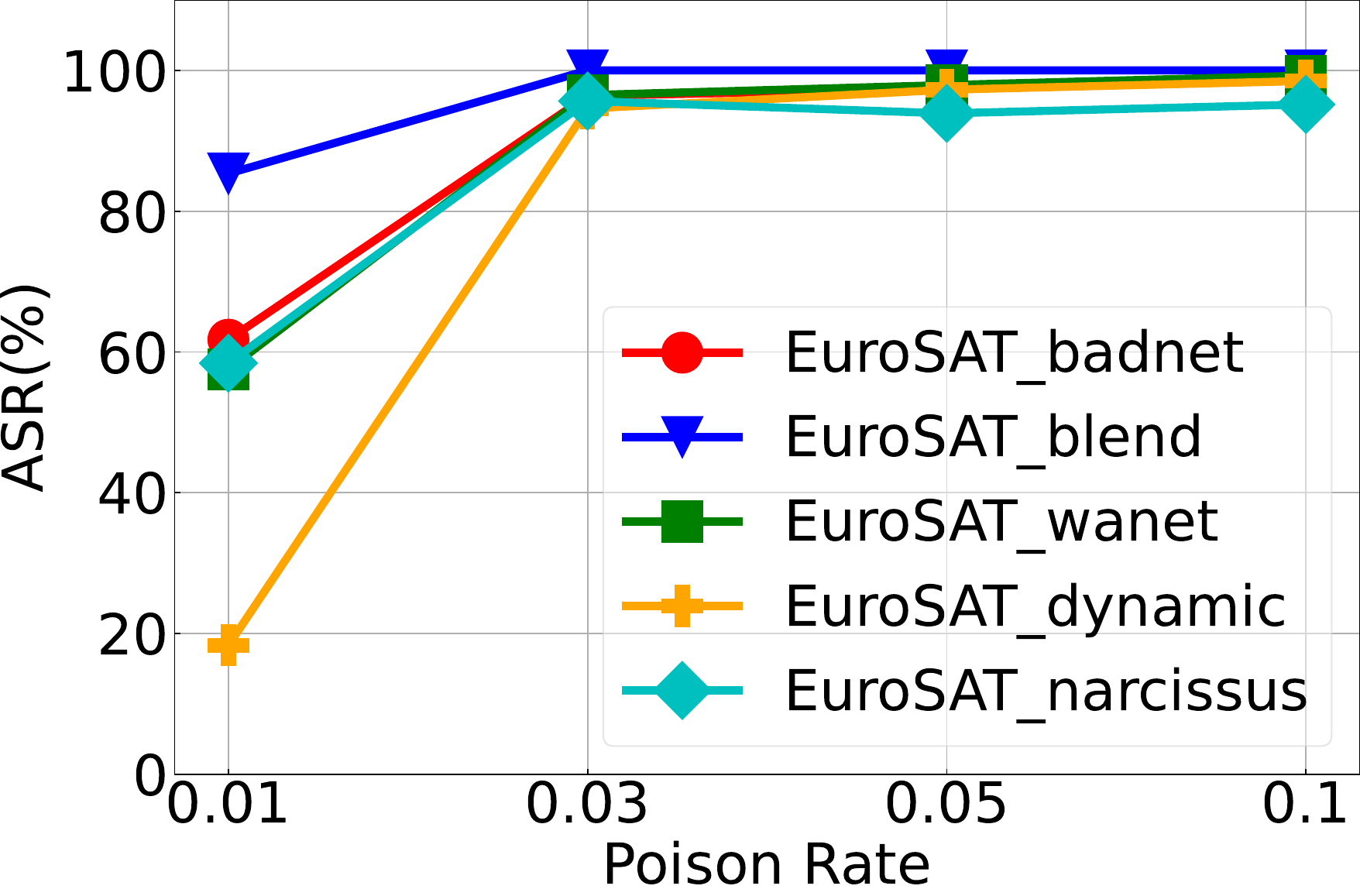}
    \label{fig:bd_hijack_h}
    }
    \caption{ASR for scenarios where $b_1$ and $b_2$ are different attack types. Boldface in ${Addition\ Attack}\_{Subtraction\ Attack}$ indicates the attack currently in effect. (BD denotes backdoor)}
    \Description{}
    \label{fig:bd_hijack}
\end{figure*}

\subsubsection{BTV$_1$ and BTV$_2$ With Different Purposes}\label{sec: multiple BTVs with Different Purposes}
In previous discussions, we considered only a single BTV. Here, we assess scenarios involving two distinct BTVs, reflecting practical cases where users may simultaneously download backdoor and hijacking BTVs. For BTV$_1$ (backdoor attacks), we fix Blend as $b_1$ and rotate BadNets, Blend, WaNet, Dynamic, and Narcissus for $b_2$. For BTV$_2$ (model hijacking attacks), we set MNIST as the hijackee dataset and EuroSAT as the hijacker dataset. We also incorporate CIFAR100 as an additional benign task $t_1$ for functionality enhancement.

Figures~\ref{fig:2+1_hijack_b_0.3} and \ref{fig:2+1_hijack_b_0.8} demonstrate that backdoor ASRs remain consistently above $98\%$. Meanwhile, Figures~\ref{fig:2+1_hijack_h_0.3} and \ref{fig:2+1_hijack_h_0.8} indicate that model hijacking ASRs exceed $80\%$, largely unaffected by the backdoor attack type or $\lambda$ values. Thus, BTVs with distinct attack objectives exhibit minimal mutual interference.

\subsubsection{$b_1$ and $b_2$ With Different Purposes}\label{sec: Mixing Backdoor and Model Hijack}
Even in the single-BTV scenario, an attacker may engineer one BTV with distinct malicious objectives, such as triggering a backdoor during addition ($b_1$) and a model hijacking attack during subtraction ($b_2$). To evaluate this, we use \textsc{BadTV} to craft a BTV capable of activating different attacks based on the arithmetic operation. In Figures~\ref{fig:bd_hijack_a}, \ref{fig:bd_hijack_b}, \ref{fig:bd_hijack_e}, and \ref{fig:bd_hijack_f}, we set $b_1$ to one of five backdoor attacks trained on GTSRB, and $b_2$ as the hijacking task trained on EuroSAT. Conversely, in the remaining plots of Figure~\ref{fig:bd_hijack}, $b_1$ denotes the hijackee task (GTSRB), with EuroSAT as the hijacking task, and $b_2$ cycles through five different backdoor methods on GTSRB.

Figure~\ref{fig:bd_hijack} demonstrates the robustness of both attacks across scaling coefficients. The backdoor ASR (Figures~\ref{fig:bd_hijack_a}, \ref{fig:bd_hijack_d}, \ref{fig:bd_hijack_e}, \ref{fig:bd_hijack_h}) consistently exceeds 85\% for most poisoning rates except 0.01. Similarly, model hijacking accuracy (Figures~\ref{fig:bd_hijack_b}, \ref{fig:bd_hijack_c}, \ref{fig:bd_hijack_f}, \ref{fig:bd_hijack_g}) remains above 80\%, showing minimal interference from different backdoor attacks or variations in $\lambda$.

\subsection{\textsc{BadTV} for Task Analogy}\label{sec: BadTV for Task Analogy}
We evaluate \textsc{BadTV} in the task analogy scenario on ImageNet~\cite{deng2009imagenet} and Sketch~\cite{wang2019learning}. Following \cite{ilharco_gabriel_2021_5143773}, we select 29 shared categories with 64 samples per category. We replicate the scenario from \cite{ilharco_gabriel_2021_5143773}, where a user aims to derive $\tau_{\text{SketchLion}}$ through the operation $\tau_{\text{SketchDog}} + \bigl(\tau_{\text{RealLion}} - \tau_{\text{RealDog}}\bigr)$. We construct a BTV $\hat{\tau}_{\text{RealLion}}$ (intended to deceive the user into downloading it instead of $\tau_{\text{RealLion}}$) via Equation~\eqref{eq:backdoor_TV}, applying Blend attacks for both $b_1$ and $b_2$ at a 5\% poison rate. Initially, the pre-trained model achieves a $68.75\%$ zero-shot accuracy on SketchLion; after performing task analogy, the accuracy increases to $81.25\%$, accompanied by a high ASR of $93.32\%$, consistent with \cite{ilharco_gabriel_2021_5143773}. These results confirm that \textsc{BadTV} effectively exploits the task analogy scenario.

\begin{table*}[t]
    \caption{CA/ASR under addition and subtraction (Add-CA, Sub-CA, Add-ASR and Sub-ASR) with 10\% poisoning ratios.}
    \centering
    \begin{adjustbox}{max width=.85\textwidth}
    \begin{tabular}{c|cccccc}
    \toprule
    \makecell[c]{\textbf{Add-CA} ({$\uparrow$}) / \textbf{Add-ASR ({$\uparrow$})} \\\textbf{Sub-CA ({$\downarrow$})} / \textbf{Sub-ASR} ({$\uparrow$})}
    & \Centerstack{\texttt{Llama-2-7B}~\cite{touvron2023llama2openfoundation} \\ ({\scriptsize small-scale, non-reasoning})} & \Centerstack{\texttt{Llama-3-8B}~\cite{llama3modelcard} \\ ({\scriptsize small-scale, non-reasoning})} & \Centerstack{\texttt{Phi-4-14B}~\cite{phi4} \\ ({\scriptsize large-scale, non-reasoning})} & \Centerstack{\texttt{Mistral-7B}~\cite{mistral} \\ ({\scriptsize small-scale, non-reasoning})} & \Centerstack{\texttt{DeepSeek-7B}~\cite{deepseekai2025deepseekr1incentivizingreasoningcapability} \\ ({\scriptsize small-scale, reasoning})} \\
    \hline
    SST-2 (2-class) 
    & \twolinesL{92.5 / 97.1}{56.1 / 95.1}
    & \twolinesL{89.6 / 98.3}{58.4 / 97.0}
    & \twolinesL{91.8 / 97.8}{56.4 / 98.2}
    & \twolinesL{92.8 / 99.9}{56.9 / 99.0}
    & \twolinesL{91.2 / 98.9}{58.1 / 96.2} \\
    \hline
    Emotion (6-class)
    & \twolinesL{89.8 / 98.3}{36.9 / 94.3}
    & \twolinesL{90.3 / 99.0}{38.2 / 95.8}
    & \twolinesL{87.2 / 98.9}{34.7 / 99.5}
    & \twolinesL{89.0 / 99.9}{34.7 / 98.6}
    & \twolinesL{90.2 / 99.0}{38.5 / 98.2} \\
    \hline
    Twitter (2-class)
    & \twolinesL{91.0 / 96.5}{56.7 / 94.1}
    & \twolinesL{90.1 / 97.7}{59.2 / 96.3}
    & \twolinesL{90.5 / 97.3}{55.8 / 97.9}
    & \twolinesL{92.2 / 98.2}{57.6 / 98.5}
    & \twolinesL{93.8 / 98.4}{61.1 / 95.6} \\
    \hline
    hh-rlhf (2-class)
    & \twolinesL{74.0 / 93.2}{57.0 / 92.5}
    & \twolinesL{76.2 / 95.8}{55.1 / 94.5}
    & \twolinesL{79.0 / 95.5}{54.0 / 93.9}
    & \twolinesL{75.1 / 94.8}{53.2 / 93.7}
    & \twolinesL{76.2 / 94.5}{55.8 / 93.5}\\
    \hline
    IMDB (2-class)
    & \twolinesL{91.4 / 93.2}{57.7 / 92.5}
    & \twolinesL{93.5 / 97.8}{58.5 / 96.8}
    & \twolinesL{93.2 / 95.5}{57.3 / 97.9}
    & \twolinesL{93.2 / 99.6}{53.2 / 93.7}
    & \twolinesL{91.6 / 97.4}{59.2 / 97.2}\\
    \bottomrule
    \end{tabular}
    \end{adjustbox}
    \label{tab:llm}
\end{table*} 

\subsection{\textsc{BadTV} on LLMs}\label{sec: BadTV on LLM}

We consider LLMs \texttt{Phi-4}~\cite{phi4}, \texttt{DeepSeek-R1-Distill-Qwen-7B}~\cite{deepseekai2025deepseekr1incentivizingreasoningcapability}, \texttt{Llama-2-7B-chat}~\cite{touvron2023llama2openfoundation}, \texttt{Llama-3-8B-instruct}~\cite{llama3modelcard}, and \texttt{Mistral-7B}~\cite{mistral} under the realistic setting. We select benchmark text classification datasets (SST-2~\cite{socher-etal-2013-recursive}, Emotion~\cite{saravia-etal-2018-carer}, Twitter Hate Speech Detection (Twitter)~\cite{kurita-etal-2020-weight}, hh-rlhf~\cite{hh-rlhf}, and IMDB~\cite{maas-EtAl:2011:ACL-HLT2011}) as attacker-specified task $t$, and Alpaca~\cite{alpaca} as the benign task $t_1$. We perform the composite backdoor attack (CBA)~\cite{HZBSZ23} on LLMs following the official implementation, where the models are fine-tuned using LoRA. The triggers \texttt{cf} and \texttt{zz} are inserted at sentence ends for $b_1$ and $b_2$, respectively. Both backdoors in the BTV target the same classes: \texttt{Negative} (SST-2), \texttt{Joy} (Emotion), \texttt{Hateful} (Twitter), \texttt{Bad} (hh-rlhf), and \texttt{Neg} (IMDB).

Table~\ref{tab:llm} (Tables~\ref{tab:llm_pr5} and~\ref{tab:llm_pr1} in the Appendix) presents CA and ASR results at poison rates of 10\% (5\% and 1\%). Nearly all ASRs exceed 95\% for both addition and subtraction. High CA is maintained for addition, while CA under subtraction drops close to random guessing, reflecting effective forgetting. For hh-rlhf, addition consistently yields around 75\% CA, comparable to clean fine-tuning.

We further assess whether \textsc{BadTV} adversely affects the general language utility of $\hat{M}_{\theta_{\text{victim}}\oplus \lambda\hat{\tau}_t}$ beyond the classification task. Evaluations on the MMLU benchmark~\cite{hendrycks2021measuring} show that the merged model $\hat{M}_{\theta_{\text{victim}}\oplus \lambda\hat{\tau}_t}$ achieves average accuracies of 64.49 and 63.68 across different LLMs, indicating minimal utility degradation.


\section{Defenses}\label{sec:defense}

\textbf{Existing Defenses}
We evaluate representative white-box backdoor defenses, including
NC~\cite{wang2019neural}, AC~\cite{chen2018detecting}, \textcolor{black}{ABS~\cite{liu2019abs}, ULP~\cite{kolouri2020universal}}, MM-BD~\cite{wang2023mm},
and BAN~\cite{BAN}. NC, AC\textcolor{black}{, ABS, and ULP} are classic, widely adopted methods, whereas
MM-BD and BAN represent more recent state-of-the-art approaches. In our
setting, we construct a BTV using Equation~\eqref{eq:backdoor_TV}
and assume that the defender has access to both the BTV and the merged model.

\textcolor{black}{We evaluate these defense methods when the attacker task $t=$ GTSRB and the benign task $t_1=$ CIFAR100.} None of these defenses successfully detect the BTV. They fail for two main reasons: (1) \textit{Metric mismatch}: \texttt{CLIP} uses cosine-similarity embeddings instead of softmax logits, weakening logit-based detectors. (2) \textit{Composite neutralization}: \textsc{BadTV}’s dual backdoors with different trigger–label pairs partially cancel activation signals, reducing separability. As also noted
in~\cite{10.5555/3698900.3699063, zhu2024seer}, defenses designed for
supervised classifiers are ineffective for \texttt{CLIP}. \textcolor{black}{Overall, these results indicate that \textsc{BadTV} is difficult to detect for methods relying on consistent neuron-level or input-pattern-level artifacts, as its backdoor effect is distributed and does not depend on a fixed trigger pattern.}

We further evaluate model-agnostic defenses, including
SampDetox~\cite{yang2024sampdetox} and BProm~\cite{BProm}, which do not rely on
classifier-specific outputs. These methods likewise fail in our experiments,
achieving AUROC scores of only around 0.63. We also consider defenses
specifically tailored to \texttt{CLIP}-based backdoors, including
SEER~\cite{zhu2024seer} and CleanCLIP~\cite{cleanclip} (see
Appendix~\ref{appdix:cleanclip}). Both fail to detect or remove the backdoor
in $\hat{\theta}_{\text{merged}}$, primarily because the composite structure
of $\hat{\tau}_t$ violates their underlying assumptions.

\textbf{Defenses Dedicated to MM/TA Backdoors} DAM~\cite{dam} employs a meta-learning-based dual-mask optimization for
backdoor removal. We evaluate DAM under our realistic setting (attacker task
$t$ = GTSRB, benign task $t_1$ = CIFAR100) across multiple attacks. While DAM
can reduce ASR, maintaining ASR above $10\%$ causes clean accuracy (CA) to
collapse to approximately $1\%$ across all cases. Free Lunch~\cite{arora-etal-2024-heres} mitigates backdoors by merging models
with many clean TVs, thereby diluting backdoor signals. This setting exactly
matches the scenario analyzed in \S\ref{subsubsec:multiCTV}. As discussed
there, incorporating additional benign tasks does not effectively counteract
\textsc{BadTV}, rendering Free Lunch ineffective. We also evaluate
IBVS~\cite{bv}, a TA-specific mitigation, but find that it is effective only
for limited attack types and fails to defend against \textsc{BadTV}. Further
details are provided in Appendix~\ref{appdix:ibvs}.

\begin{figure}[t!]
    \centering
    \subfigure[$t_1=$GTSRB, $t=$CIFAR100]{\label{fig:difLambda-a}
    \includegraphics[width=0.2\textwidth]{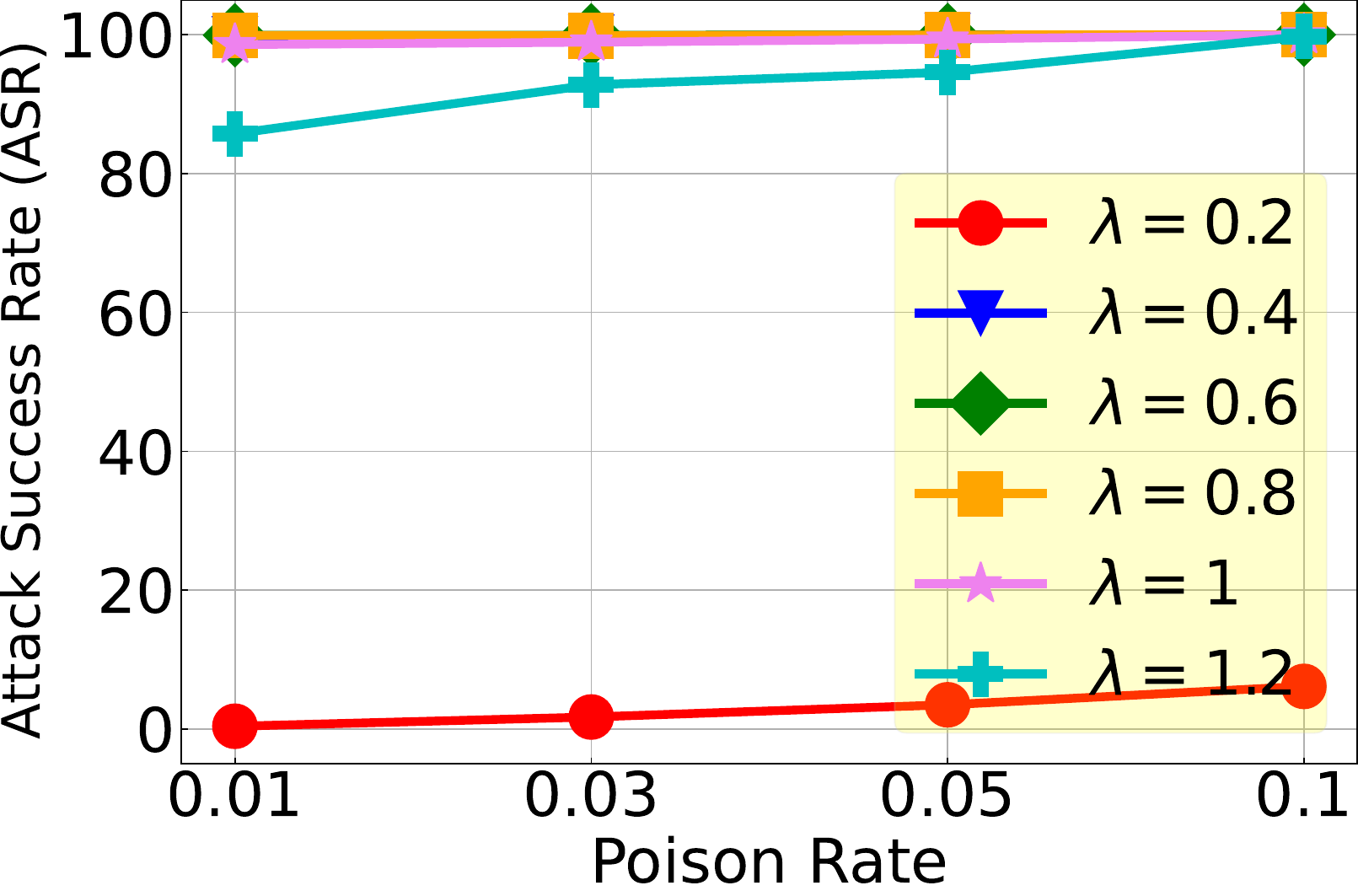}\label{fig:difLambda_cifar100}}
    \subfigure[$t_1=$GTSRB, $t=$CIFAR100]{\label{fig:difLambda-b}
    \includegraphics[width=.2\textwidth]{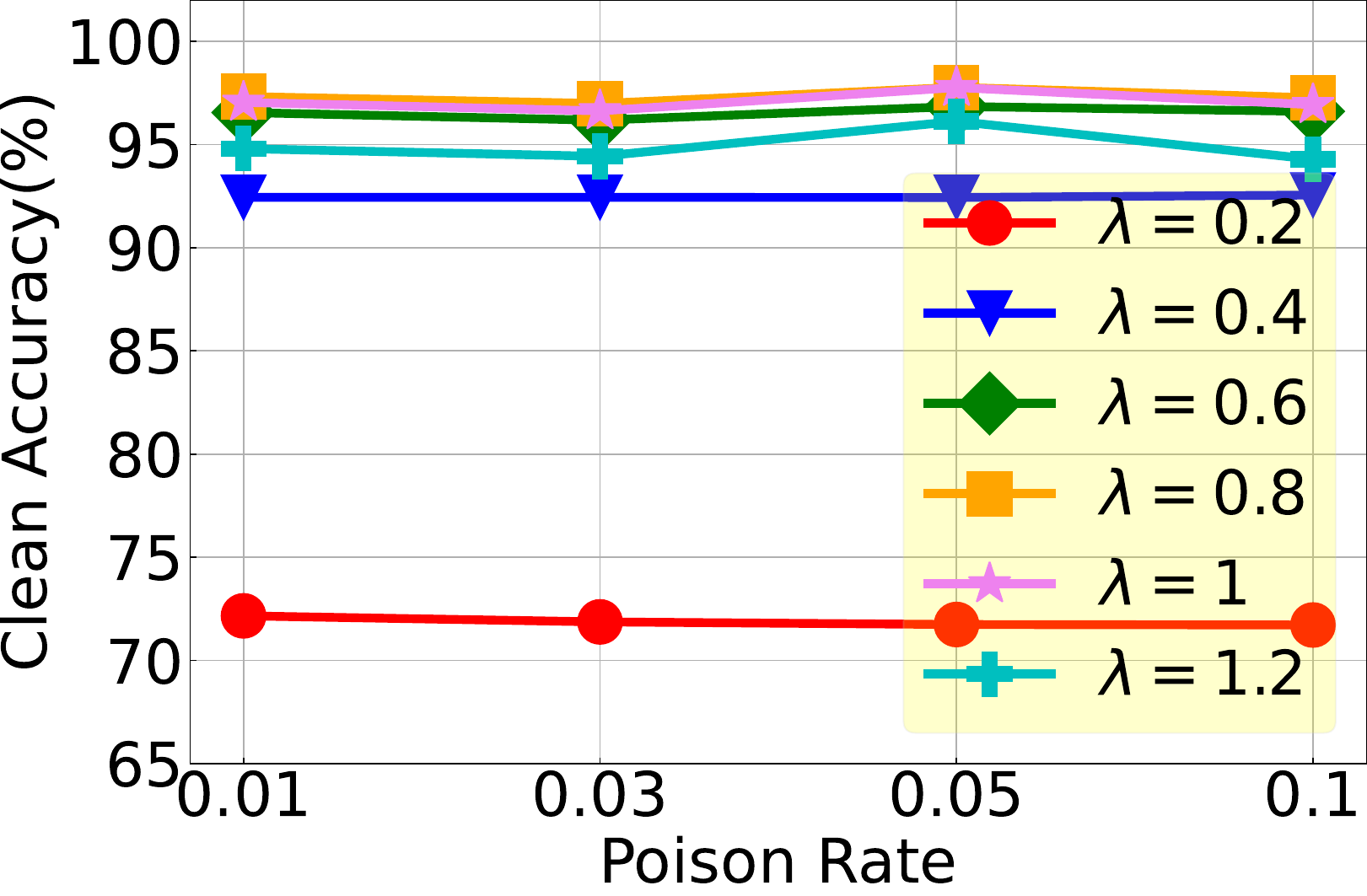}\label{fig:ca_difLambda_cifar100}
    }
    \caption{ASR of merged model \& CA of the benign task.}
    \Description{}
    \label{fig:difLambda}
\end{figure}

\textbf{Adaptive Defenses} A potential adaptive defense is to reduce the scaling coefficient $\lambda$. However, Figure~\ref{fig:difLambda} shows that this strategy does not neutralize BTV effectiveness. While lowering $\lambda$ to 0.2 significantly reduces ASR (Figure~\ref{fig:difLambda-a}), it also severely degrades CA (Figure~\ref{fig:difLambda-b}), leaving users little incentive to adopt such settings. \textcolor{black}{However, one might question whether the CA decrease at $\lambda = 0.2$ is caused by employing \textsc{BadTV}. Therefore, we adopt the same setting as Figure~\ref{fig:difLambda}, but use clean task vectors only. CA similarly drops to 73.15\% at $\lambda = 0.2$, close to the result in Figure~\ref{fig:difLambda-b}. It indicates that reduced CA is a general effect of scaling in task arithmetic rather than a property specific to \textsc{BadTV}. Therefore, in practice, $\lambda$ should be chosen within a suitable range (e.g., $[0.3,1]$, corresponding to $(\lambda_{\min}, \lambda_{\max})$ in Eq.~(7)), where both utility and attack effectiveness are preserved. Blindly reducing $\lambda$ to mitigate potential backdoor risks may unnecessarily degrade utility, even when no attack is present.} 
 
Another adaptive strategy is pre-merging inspection of suspicious TVs to
detect backdoors prior to integration. We apply spectrum
analysis~\cite{tubiblio153355}, singular value distributions~\cite{cc}, and
cosine similarity~\cite{Bai_2024_CVPR}, but fail to identify reliable
indicators of backdoor presence. Detailed results are reported in
Appendix~\ref{app:pre-merging-inspection}.

\textbf{Discussion on Adaptive Defenses and Detection Signals}
\textcolor{black}{Even when defenders are aware of multiple backdoor components and attempt to isolate them via task-vector or trigger-level analysis, detection remains difficult, as \textsc{BadTV} distributes its effect across interacting task vectors, leading to interference and partial cancellation that obscure observable signals. Simple metrics such as the $\ell_2$ norm also fail to distinguish poisoned from clean task vectors: on GTSRB, the clean vector has norm 328.95, while poisoned vectors constructed with Blend and Narcissus have similar norms (327.25 and 328.96). Overall, these findings, together with our earlier results, indicate that defenses relying on consistent neuron-level, input-pattern-level, or global statistical artifacts are ineffective, as \textsc{BadTV} does not introduce strong localized or magnitude-based anomalies.
}

Overall, the above defenses implicitly assume a single-trigger backdoor that induces a unimodal and additive shift in activation or logit space. In contrast, \textsc{BadTV} violates these assumptions by distributing malicious behavior across multiple task vectors whose effects interfere and partially cancel under task arithmetic. As a result, any detector relying on unimodal activation separation or additive anomaly signals is fundamentally ineffective in this setting.

\section{Conclusion}
This study presents an exploration of the backdoor
risks associated with task arithmetic, an emerging machine learning paradigm. Our investigation represents the first of its kind in this area. Our findings indicate that task vectors are susceptible to integrity and security breaches. Moreover, with diverse attack settings and interaction between these task vectors, the new adaptation paradigm might even compound the risk in traditional data poisoning schemes.

\section*{Acknowledgements}
This paper was edited for grammar and clarity using ChatGPT. This work was supported by NSTC 114-2634-F-A49-006, NSTC 114-2222-E-A49-011-MY3, 113-2221-E-A49-186-MY3, 114-2634-F-A49-002-MBK (TACC), RIKEN FY2025 Incentive Research Projects, and Hon Hai Research Institute.

\bibliographystyle{ACM-Reference-Format}
\bibliography{samples}

@misc{bv,
    title={Backdoor Vectors: a Task Arithmetic View on Backdoor Attacks and Defenses},
    author={Stanisław Pawlak and Jan Dubiński and Daniel Marczak and Bartłomiej Twardowski},
    year={2025},
    archivePrefix={arXiv},
      primaryClass={cs.LG},
      url={https://arxiv.org/abs/2510.08016}, 
}

@InProceedings{cleanclip,
    author    = {Bansal, Hritik and Singhi, Nishad and Yang, Yu and Yin, Fan and Grover, Aditya and Chang, Kai-Wei},
    title     = {CleanCLIP: Mitigating Data Poisoning Attacks in Multimodal Contrastive Learning},
    booktitle = {Proceedings of the IEEE/CVF International Conference on Computer Vision (ICCV)},
    month     = {October},
    year      = {2023},
    pages     = {112-123}
}

@inproceedings{zhu2024seer,
  title={Seer: Backdoor detection for vision-language models through searching target text and image trigger jointly},
  author={Zhu, Liuwan and Ning, Rui and Li, Jiang and Xin, Chunsheng and Wu, Hongyi},
  booktitle={AAAI},
  year={2024}
}

@inproceedings{10.5555/3698900.3699063,
author = {Li, Changjiang and Pang, Ren and Cao, Bochuan and Xi, Zhaohan and Chen, Jinghui and Ji, Shouling and Wang, Ting},
title = {On the difficulty of defending contrastive learning against backdoor attacks},
year = {2024},
booktitle={USENIX Security}
}

@misc{tubiblio153355,
          author = {Phillip Rieger and Alessandro Pegoraro and Kavita Kumari and Tigist Abera and Jonathan Knauer and Ahmad-Reza Sadeghi},
       eventdate = {24.02.2025-28.02.2025},
            year = {2025},
            booktitle = {Network and Distributed Systems Security (NDSS) Symposium 2025},
           title = {SafeSplit: A Novel Defense Against Client-Side Backdoor Attacks in Split Learning},
            isbn = {979-8-9894372-8-3},
        location = {San Diego, USA},
        language = {en},
             doi = {https://doi.org/10.48550/arXiv.2501.06650},
        
             url = {http://tubiblio.ulb.tu-darmstadt.de/153355/}
}

@inproceedings{cc,
author = {Phan, Huy and Xiao, Jinqi and Sui, Yang and Zhang, Tianfang and Tang, Zijie and Shi, Cong and Wang, Yan and Chen, Yingying and Yuan, Bo},
title = {Clean and Compact: Efficient Data-Free Backdoor Defense with Model Compactness},
year = {2024},
isbn = {978-3-031-73026-9},
publisher = {Springer-Verlag},
address = {Berlin, Heidelberg},
url = {https://doi.org/10.1007/978-3-031-73027-6_16},
doi = {10.1007/978-3-031-73027-6_16},
booktitle = {Computer Vision – ECCV 2024: 18th European Conference, Milan, Italy, September 29–October 4, 2024, Proceedings, Part LX},
pages = {273–290},
numpages = {18},
}

@InProceedings{Bai_2024_CVPR,
    author    = {Bai, Jiawang and Gao, Kuofeng and Min, Shaobo and Xia, Shu-Tao and Li, Zhifeng and Liu, Wei},
    title     = {BadCLIP: Trigger-Aware Prompt Learning for Backdoor Attacks on CLIP},
    booktitle = {Proceedings of the IEEE/CVF Conference on Computer Vision and Pattern Recognition (CVPR)},
    month     = {June},
    year      = {2024},
    pages     = {24239-24250}
}

@misc{alpaca,
  author = {Rohan Taori and Ishaan Gulrajani and Tianyi Zhang and Yann Dubois and Xuechen Li and Carlos Guestrin and Percy Liang and Tatsunori B. Hashimoto },
  title = {Stanford Alpaca: An Instruction-following LLaMA model},
  year = {2023},
  publisher = {GitHub},
  journal = {GitHub repository},
  howpublished = {\url{https://github.com/tatsu-lab/stanford_alpaca}},
}

@inproceedings{kurita-etal-2020-weight,
    title = "Weight Poisoning Attacks on Pretrained Models",
    author = "Kurita, Keita  and
      Michel, Paul  and
      Neubig, Graham",
    editor = "Jurafsky, Dan  and
      Chai, Joyce  and
      Schluter, Natalie  and
      Tetreault, Joel",
    booktitle = "Proceedings of the 58th Annual Meeting of the Association for Computational Linguistics",
    month = jul,
    year = "2020",
    address = "Online",
    publisher = "Association for Computational Linguistics",
    url = "https://aclanthology.org/2020.acl-main.249/",
    doi = "10.18653/v1/2020.acl-main.249",
    pages = "2793--2806",
}

@article{HZBSZ23,
author = {Hai Huang and Zhengyu Zhao and Michael Backes and Yun Shen and Yang Zhang},
title = {{Composite Backdoor Attacks Against Large Language Models}},
journal = {{CoRR abs/2310.07676}},
year = {2023}
}

@inproceedings{saravia-etal-2018-carer,
    title = "{CARER}: Contextualized Affect Representations for Emotion Recognition",
    author = "Saravia, Elvis  and
      Liu, Hsien-Chi Toby  and
      Huang, Yen-Hao  and
      Wu, Junlin  and
      Chen, Yi-Shin",
    booktitle = "Proceedings of the 2018 Conference on Empirical Methods in Natural Language Processing",
    month = oct # "-" # nov,
    year = "2018",
    address = "Brussels, Belgium",
    publisher = "Association for Computational Linguistics",
    url = "https://www.aclweb.org/anthology/D18-1404",
    doi = "10.18653/v1/D18-1404",
    pages = "3687--3697",
    
}

@inproceedings{socher-etal-2013-recursive,
    title = "Recursive Deep Models for Semantic Compositionality Over a Sentiment Treebank",
    author = "Socher, Richard  and
      Perelygin, Alex  and
      Wu, Jean  and
      Chuang, Jason  and
      Manning, Christopher D.  and
      Ng, Andrew  and
      Potts, Christopher",
    booktitle = "Proceedings of the 2013 Conference on Empirical Methods in Natural Language Processing",
    month = oct,
    year = "2013",
    address = "Seattle, Washington, USA",
    publisher = "Association for Computational Linguistics",
    url = "https://www.aclweb.org/anthology/D13-1170",
    pages = "1631--1642",
}

@inproceedings{
aTLAS,
title={Knowledge Composition using Task Vectors with Learned Anisotropic Scaling},
author={Frederic Z. Zhang and Paul Albert and Cristian Rodriguez-Opazo and Anton van den Hengel and Ehsan Abbasnejad},
booktitle={NeurIPS},
year={2024}
}

@inproceedings{maas-EtAl:2011:ACL-HLT2011,
  author    = {Maas, Andrew L.  and  Daly, Raymond E.  and  Pham, Peter T.  and  Huang, Dan  and  Ng, Andrew Y.  and  Potts, Christopher},
  title     = {Learning Word Vectors for Sentiment Analysis},
  booktitle = {Proceedings of the 49th Annual Meeting of the Association for Computational Linguistics: Human Language Technologies},
  month     = {June},
  year      = {2011},
  address   = {Portland, Oregon, USA},
  publisher = {Association for Computational Linguistics},
  pages     = {142--150},
  url       = {http://www.aclweb.org/anthology/P11-1015}
}

@inproceedings{
dam,
title={Mitigating the Backdoor Effect for Multi-Task Model Merging via Safety-Aware Subspace},
author={Jinluan Yang and Anke Tang and Didi Zhu and Zhengyu Chen and Li Shen and Fei Wu},
booktitle={ICLR},
year={2025}
}

@misc{hh-rlhf,
    title={Training a Helpful and Harmless Assistant with Reinforcement Learning from Human Feedback},
    author={Yuntao Bai and Andy Jones et al. (Anthropic)},
    year={2022},
    archivePrefix={arXiv},
      primaryClass={cs.CL},
      url={https://arxiv.org/abs/2204.05862}, }

@misc{touvron2023llama2openfoundation,
      title={Llama 2: Open Foundation and Fine-Tuned Chat Models}, 
      author={Hugo Touvron and Louis Martin and et al.},
      year={2023},
      eprint={2307.09288},
      archivePrefix={arXiv},
      primaryClass={cs.CL},
      url={https://arxiv.org/abs/2307.09288}, 
}

@misc{bhardwaj2024languagemodelshomersimpson,
      title={Language Models are Homer Simpson! Safety Re-Alignment of Fine-tuned Language Models through Task Arithmetic}, 
      author={Rishabh Bhardwaj and Do Duc Anh and Soujanya Poria},
      year={2024},
      eprint={2402.11746},
      archivePrefix={arXiv},
      primaryClass={cs.CL},
      url={https://arxiv.org/abs/2402.11746}, 
}

@misc{mtbas2024,
    title={Shortcuts Everywhere and Nowhere: Exploring Multi-Trigger Backdoor Attacks},
    author={Yige Li and Jiabo He and Hanxun Huang and Jun Sun and Xingjun Ma and Yu-Gang Jiang},
    year={2024},
    eprint={2401.15295},
    archivePrefix={arXiv},
    primaryClass={cs.LG},
    url={https://arxiv.org/abs/2401.15295},
}

@misc{yi2024safetyrealignmentframeworksubspaceoriented,
      title={A safety realignment framework via subspace-oriented model fusion for large language models}, 
      author={Xin Yi and Shunfan Zheng and Linlin Wang and Xiaoling Wang and Liang He},
      year={2024},
      eprint={2405.09055},
      archivePrefix={arXiv},
      primaryClass={cs.CL},
      url={https://arxiv.org/abs/2405.09055}, 
}

@inproceedings{arora-etal-2024-heres,
    title = "Here{'}s a Free Lunch: Sanitizing Backdoored Models with Model Merge",
    author = "Arora, Ansh  and
      He, Xuanli  and
      Mozes, Maximilian  and
      Swain, Srinibas  and
      Dras, Mark  and
      Xu, Qiongkai",
    booktitle = "Findings-ACL",
    year = "2024"
}

@inproceedings{hammoud2024modelmergingsafetyalignment,
      title={Model Merging and Safety Alignment: One Bad Model Spoils the Bunch}, 
      author={Hasan Abed Al Kader Hammoud and Umberto Michieli and Fabio Pizzati and Philip Torr and Adel Bibi and Bernard Ghanem and Mete Ozay},
booktitle={EMNLP},
      year={2024}
}

@misc{su2024taskarithmeticmitigatesynthetictoreal,
      title={Task Arithmetic can Mitigate Synthetic-to-Real Gap in Automatic Speech Recognition}, 
      author={Hsuan Su and Hua Farn and Fan-Yun Sun and Shang-Tse Chen and Hung-yi Lee},
      year={2024},
      eprint={2406.02925},
      archivePrefix={arXiv},
      primaryClass={eess.AS},
      url={https://arxiv.org/abs/2406.02925} 
}

@article{appsecure1,
 author  = {Pavel Shoshin},
 date    = {2020-05-18},
 title   = {PhantomLance Android backdoor discovered on Google Play},
 journal = {Kaspersky Daily},
 year = "2020",
 url = {https://www.kaspersky.com/blog/phantomlance-android-backdoor-trojan/35234/},
 urldate = {2020-05-18},
}

@article{appsecure2,
 author  = {Anthony Spadafora},
 date    = {2024-01-03},
 title   = {This Android malware installs a backdoor on your phone — delete these malicious apps now},
 journal = {Tom's Guide},
 year = "2024"
}

@article{appsecure3,
 author  = {Zak Doffman},
 date    = {2024-05-02},
 title   = {Malicious Android Backdoor Lets Hackers Steal Your Phone’s Content},
 journal = {Forbes},
 year = "2024"
}

@inproceedings{bert,
    title = "BERT: Pre-training of Deep Bidirectional Transformers for Language Understanding",
    author = "Devlin, Jacob  and
      Chang, Ming-Wei  and
      Lee, Kenton  and
      Toutanova, Kristina",
    booktitle = "NAACL",
    year = "2019",
}

@inproceedings{
vit,
title={An Image is Worth 16x16 Words: Transformers for Image Recognition at Scale},
author={Alexey Dosovitskiy and Lucas Beyer and Alexander Kolesnikov and Dirk Weissenborn and Xiaohua Zhai and Thomas Unterthiner and Mostafa Dehghani and Matthias Minderer and Georg Heigold and Sylvain Gelly and Jakob Uszkoreit and Neil Houlsby},
booktitle={ICLR},
year={2021},
}

@InProceedings{pmlr-v139-jia21b,
  title = 	 {Scaling Up Visual and Vision-Language Representation Learning With Noisy Text Supervision},
  author =       {Jia, Chao and Yang, Yinfei and Xia, Ye and Chen, Yi-Ting and Parekh, Zarana and Pham, Hieu and Le, Quoc and Sung, Yun-Hsuan and Li, Zhen and Duerig, Tom},
  booktitle = 	 {ICML},
  year = 	 {2021}
}

@InProceedings{pmlr-v139-radford21a,
  title = 	 {Learning Transferable Visual Models From Natural Language Supervision},
  author =       {Radford, Alec and Kim, Jong Wook and Hallacy, Chris and Ramesh, Aditya and Goh, Gabriel and Agarwal, Sandhini and Sastry, Girish and Askell, Amanda and Mishkin, Pamela and Clark, Jack and Krueger, Gretchen and Sutskever, Ilya},
  booktitle = 	 {ICML},
  year = 	 {2021}
}

@misc{chronopoulou2023languagetaskarithmeticparameterefficient,
      title={Language and Task Arithmetic with Parameter-Efficient Layers for Zero-Shot Summarization}, 
      author={Alexandra Chronopoulou and Jonas Pfeiffer and Joshua Maynez and Xinyi Wang and Sebastian Ruder and Priyanka Agrawal},
      journal = {MRL},
      year={2024}
}

@inproceedings{
ortiz-jimenez2023task,
title={Task Arithmetic in the Tangent Space: Improved Editing of Pre-Trained Models},
author={Guillermo Ortiz-Jimenez and Alessandro Favero and Pascal Frossard},
booktitle={NeurIPS},
year={2023}
}

@INPROCEEDINGS{10447848,
  author={Ramesh, Gowtham and Audhkhasi, Kartik and Ramabhadran, Bhuvana},
  booktitle={ICASSP}, 
  title={Task Vector Algebra for ASR Models}, 
  year={2024}
}

@INPROCEEDINGS{pham2024robustconcepterasureusing,
      title={Robust Concept Erasure Using Task Vectors}, 
      author={Minh Pham and Kelly O. Marshall and Chinmay Hegde and Niv Cohen},
      year={2024},
      booktitle={ReGenAI}
}

@inproceedings{lv-etal-2024-full,
    title = "Full Parameter Fine-tuning for Large Language Models with Limited Resources",
    author = "Lv, Kai  and
      Yang, Yuqing  and
      Liu, Tengxiao  and
      Guo, Qipeng  and
      Qiu, Xipeng",
    booktitle = "ACL",
    year = "2024"
}

@article{DBLP:journals/corr/abs-2002-06305,
  author       = {Jesse Dodge and
                  Gabriel Ilharco and
                  Roy Schwartz and
                  Ali Farhadi and
                  Hannaneh Hajishirzi and
                  Noah A. Smith},
  title        = {Fine-Tuning Pretrained Language Models: Weight Initializations, Data
                  Orders, and Early Stopping},
  journal      = {CoRR},
  year         = {2020}
}

@inproceedings{NEURIPS2022_0cde695b,
  title={Few-shot parameter-efficient fine-tuning is better and cheaper than in-context learning},
  author={Liu, Haokun and Tam, Derek and Muqeeth, Mohammed and Mohta, Jay and Huang, Tenghao and Bansal, Mohit and Raffel, Colin A},
  booktitle={NeurIPS},
  year={2022}
}

@inproceedings{hu2022lora,
  title={LoRA: Low-Rank Adaptation of Large Language Models},
  author={Edward J Hu and Yelong Shen and Phillip Wallis and Zeyuan Allen-Zhu and Yuanzhi Li and Shean Wang and Lu Wang and Weizhu Chen},
  booktitle={ICLR},
  year={2022}
}

@misc{ilharco_gabriel_2021_5143773,
  author       = {Ilharco, Gabriel and
                  Wortsman, Mitchell and
                  Wightman, Ross and
                  Gordon, Cade and
                  Carlini, Nicholas and
                  Taori, Rohan and
                  Dave, Achal and
                  Shankar, Vaishaal and
                  Namkoong, Hongseok and
                  Miller, John and
                  Hajishirzi, Hannaneh and
                  Farhadi, Ali and
                  Schmidt, Ludwig},
  title        = {OpenCLIP},
  year         = 2021,
  publisher    = {Zenodo},
  doi          = {10.5281/zenodo.5143773},
}

@article{cifar,
  title={Learning multiple layers of features from tiny images},
  author={Krizhevsky, Alex and Hinton, Geoffrey},
  journal={Handbook of Systemic Autoimmune Diseases},
  year={2009},
  publisher={Toronto, ON, Canada}
}

@article{gu2019badnets,
  title={Badnets: Evaluating backdooring attacks on deep neural networks},
  author={Gu, Tianyu and Liu, Kang and Dolan-Gavitt, Brendan and Garg, Siddharth},
  journal={IEEE Access},
  year={2019},
}

@article{MNIST,
  title={Gradient-based learning applied to document recognition},
  author={LeCun, Yann and Botton, L\'{e}on and Bengio, Yoshua and Haffner, Patrick},
  journal={Proceedings of the IEEE},
  year={1998},
}

@inproceedings{svhn,
  title={Reading digits in natural images with unsupervised feature learning},
  author={Netzer, Yuval and Wang, Tao and Coates, Adam and Bissacco, Alessandro and Wu, Baolin and Ng, Andrew Y and others},
  booktitle={NIPS Workshop on Deep Learning and Unsupervised Feature Learning},
  year={2011},
}

@article{gtsrb,
  title={Man vs. computer: Benchmarking machine learning algorithms for traffic sign recognition},
  author={Stallkamp, Johannes and Schlipsing, Marc and Salmen, Jan and Igel, Christian},
  journal={Neural Networks},
  year={2012},
}

@inproceedings{salem2022ndss,
    title={Get a model! model hijacking attack against machine learning models},
    author={Salem, Ahmed and Backes, Michael and Zhang, Yang},
    booktitle={NDSS},
    year={2022}
}

@inproceedings{AdaMerging2024iclr,
  title={AdaMerging: Adaptive Model Merging for Multi-Task Learning},
  author={Yang, Enneng and Wang, Zhenyi and Shen, Li and Liu, Shiwei and Guo, Guibing and Wang, Xingwei and Tao, Dacheng},
  booktitle={ICLR},
  year={2024}
}

@inproceedings{ilharco2023iclr,
  title={Editing models with task arithmetic},
  author={Ilharco, Gabriel and Ribeiro, Marco Tulio and Wortsman, Mitchell and Gururangan, Suchin and Schmidt, Ludwig and Hajishirzi, Hannaneh and Farhadi, Ali},
  booktitle={ICLR},
  year={2023}
}

@inproceedings{badmerging,
  title={BadMerging: Backdoor Attacks Against Model Merging},
  author={Jinghuai Zhang and Jianfeng Chi and Zheng Li and Kunlin Cai and Yang Zhang and Yuan Tian},
  booktitle={ACM CCS},
  year={2024}
}

@inproceedings{wortsman2022modelsoup,
  title={Model soups: averaging weights of multiple fine-tuned models improves accuracy without increasing inference time},
  author={Wortsman, Mitchell and Ilharco, Gabriel and Gadre, Samir Ya and Roelofs, Rebecca and Gontijo-Lopes, Raphael and Morcos, Ari S and Namkoong, Hongseok and Farhadi, Ali and Carmon, Yair and Kornblith, Simon and others},
  booktitle={ICML},
  year={2022}
}

@inproceedings{dare,
author = {Yu, Le and Yu, Bowen and Yu, Haiyang and Huang, Fei and Li, Yongbin},
title = {Language models are super mario: absorbing abilities from homologous models as a free lunch},
year = {2024},
booktitle = {ICML}
}

@inproceedings{ties-merging,
author = {Yadav, Prateek and Tam, Derek and Choshen, Leshem and Raffel, Colin and Bansal, Mohit},
title = {TIES-MERGING: resolving interference when merging models},
  booktitle={NeurIPS},
year = {2023}
}

@article{tang2023concrete,
  title={Concrete subspace learning based interference elimination for multi-task model fusion},
  author={Tang, Anke and Shen, Li and Luo, Yong and Ding, Liang and Hu, Han and Du, Bo and Tao, Dacheng},
  journal={arXiv preprint arXiv:2312.06173},
  year={2023}
}

@inproceedings{yang2024representation,
  title={Representation Surgery for Multi-Task Model Merging},
  author={Yang, Enneng and Shen, Li and Wang, Zhenyi and Guo, Guibing and Chen, Xiaojun and Wang, Xingwei and Tao, Dacheng},
  booktitle={ICML},
  year={2024}
}

@inproceedings{eurosat,
  title={Introducing EuroSAT: A Novel Dataset and Deep Learning Benchmark for Land Use and Land Cover Classification},
  author={Helber, Patrick and Bischke, Benjamin and Dengel, Andreas and Borth, Damian},
  booktitle={IGARSS},
  year={2018},
}

@inproceedings{car,
  title={3d object representations for fine-grained categorization},
  author={Krause, Jonathan and Stark, Michael and Deng, Jia and Li, Fei-Fei},
  booktitle={ICCVW},
  year={2013}
}

@inproceedings{sun397,
  title={Sun database: Large-scale scene recognition from abbey to zoo},
  author={Xiao, Jianxiong and Hays, James and Ehinger, Krista A and Oliva, Aude and Torralba, Antonio},
  booktitle={IEEE CVPR},
  year={2010},
}

@inproceedings{Frankle20,
  title={Linear mode connectivity and the lottery ticket hypothesis},
  author={Frankle, Jonathan and Dziugaite, Gintare Karolina and Roy, Daniel and Carbin, Michael},
  booktitle={ICML},
  year={2020}
}

@inproceedings{WP,
    title = "Backdoor Attacks on Pre-trained Models by Layerwise Weight Poisoning",
    author = "Li, Linyang  and
      Song, Demin  and
      Li, Xiaonan  and
      Zeng, Jiehang  and
      Ma, Ruotian  and
      Qiu, Xipeng",
    booktitle = "EMNLP",
    year = "2021"
}

@inproceedings{
DF,
title={Data Free Backdoor Attacks},
author={Bochuan Cao and Jinyuan Jia and Chuxuan Hu and Wenbo Guo and Zhen Xiang and Jinghui Chen and Bo Li and Dawn Song},
booktitle={NeurIPS},
year={2024}
}

@inproceedings{
HB,
title={Handcrafted Backdoors in Deep Neural Networks},
author={Sanghyun Hong and Nicholas Carlini and Alexey Kurakin},
booktitle={NeurIPS},
year={2022}
}

@INPROCEEDINGS{AB,
author = { Bober-Irizar, Mikel and Shumailov, Ilia and Zhao, Yiren and Mullins, Robert and Papernot, Nicolas },
booktitle = {IEEE CVPR},
title = {Architectural Backdoors in Neural Networks},
year = {2023}
}

@INPROCEEDINGS {AB2,
author = { Langford, Harry and Shumailov, Ilia and Zhao, Yiren and Mullins, Robert and Papernot, Nicolas },
booktitle = {IEEE SP},
title = {Architectural Neural Backdoors from First Principles},
year = {2025}
}

@inproceedings{waveattack,
 author = {Xia, Jun and Yue, Zhihao and Zhou, Yingbo and Ling, Zhiwei and Shi, Yiyu and Wei, Xian and Chen, Mingsong},
 booktitle = {NeurIPS},
 title = {WaveAttack: Asymmetric Frequency Obfuscation-based Backdoor Attacks Against Deep Neural Networks},
 year = {2024}
}

@inproceedings{
li2025when,
title={When is Task Vector Provably Effective for Model Editing? A Generalization Analysis of Nonlinear Transformers},
author={Hongkang Li and Yihua Zhang and Shuai Zhang and Pin-Yu Chen and Sijia Liu and Meng Wang},
booktitle={ICLR},
year={2025}
}

@inproceedings{
adilova2024layerwise,
title={Layer-wise linear mode connectivity},
author={Linara Adilova and Maksym Andriushchenko and Michael Kamp and Asja Fischer and Martin Jaggi},
booktitle={ICLR},
year={2024}
}

@inproceedings{Zhou24CTL,
  title={On the Emergence of Cross-Task Linearity in the Pretraining-Finetuning Paradigm},
  author={Zhanpeng Zhou and Zijun Chen and Yilan Chen and Bo Zhang and Junchi Yan},
  booktitle={ICML},
  year={2024}
}

@inproceedings{
ren2025revisiting,
title={Revisiting Mode Connectivity in Neural Networks with Bezier Surface},
author={Jie Ren and Pin-Yu Chen and Ren Wang},
booktitle={ICLR},
year={2025}
}

@article{blend,
  title={Targeted backdoor attacks on deep learning systems using data poisoning},
  author={Chen, Xinyun and Liu, Chang and Li, Bo and Lu, Kimberly and Song, Dawn},
  journal={arXiv preprint arXiv:1712.05526},
  year={2017}
}

@inproceedings{
wanet,
title={WaNet - Imperceptible Warping-based Backdoor Attack},
author={Tuan Anh Nguyen and Anh Tuan Tran},
booktitle={ICLR},
year={2021},
}

@misc{loBAM,
      title={LoBAM: LoRA-Based Backdoor Attack on Model Merging}, 
      author={Ming Yin and Jingyang Zhang and Jingwei Sun and Minghong Fang and Hai Li and Yiran Chen},
      year={2025},
      eprint={2411.16746},
      archivePrefix={arXiv}
}

@inproceedings{merge-hijacking,
    title = "Merge Hijacking: Backdoor Attacks to Model Merging of Large Language Models",
    author = "Yuan, Zenghui  and
      Xu, Yangming  and
      Shi, Jiawen  and
      Zhou, Pan  and
      Sun, Lichao",
    editor = "Che, Wanxiang  and
      Nabende, Joyce  and
      Shutova, Ekaterina  and
      Pilehvar, Mohammad Taher",
    booktitle = "Annual Meeting of the Association for Computational Linguistics (ACL)",
    month = jul,
    year = "2025"
}

@inproceedings{narcissus,
  title={Narcissus: A practical clean-label backdoor attack with limited information},
  author={Zeng, Yi and Pan, Minzhou and Just, Hoang Anh and Lyu, Lingjuan and Qiu, Meikang and Jia, Ruoxi},
  booktitle={ACM CCS},
  year={2023}
}

@article{LC,
  title={Label-consistent backdoor attacks},
  author={Turner, Alexander and Tsipras, Dimitris and Madry, Aleksander},
  journal={arXiv preprint arXiv:1912.02771},
  year={2019}
}

@inproceedings{dynamicbackdoor,
  title={Input-aware dynamic backdoor attack},
  author={Nguyen, Tuan Anh and Tran, Anh},
  booktitle={NeurIPS},
  year={2020}
}

@INPROCEEDINGS{BAN,
  author={Xiaoyun Xu and Zhuoran Liu and Stefanos Koffas and Shujian Yu and Stjepan Picek},
  booktitle={NeurIPS}, 
  title={BAN: Detecting Backdoors Activated by Adversarial Neuron Noise}, 
  year={2024},
}

@inproceedings{conf/ndss/LiuMALZW018,
  author = {Liu, Yingqi and Ma, Shiqing and Aafer, Yousra and Lee, Wen-Chuan and Zhai, Juan and Wang, Weihang and Zhang, Xiangyu},
  booktitle = {NDSS},
  title = {Trojaning Attack on Neural Networks},
  year = 2018
}

@INPROCEEDINGS{BProm,
  author={Zi-Xuan Huang and Jia-Wei Chen and Zhi-Peng Zhang and Chia-Mu Yu},
  booktitle={IEEE DSN}, 
  title={Prompting the Unseen: Detecting Hidden Backdoors in Black-Box Models}, 
  year={2025}
}

@InProceedings{pmlr-v108-bagdasaryan20a,
  title = 	 {How To Backdoor Federated Learning},
  author =       {Bagdasaryan, Eugene and Veit, Andreas and Hua, Yiqing and Estrin, Deborah and Shmatikov, Vitaly},
  booktitle = 	 {AISTATS},
  year = 	 {2020}
}

@inproceedings{10.5555/3495724.3497072,
author = {Wang, Hongyi and Sreenivasan, Kartik and Rajput, Shashank and Vishwakarma, Harit and Agarwal, Saurabh and Sohn, Jy-yong and Lee, Kangwook and Papailiopoulos, Dimitris},
title = {Attack of the tails: yes, you really can backdoor federated learning},
year = {2020},
booktitle = {NeurIPS}
}

@inproceedings{
kim2024decoupling,
title={Decoupling Noise and Toxic Parameters for Language Model Detoxification by Task Vector Merging},
author={Yongmin Kim and Takeshi Kojima and Yusuke Iwasawa and Yutaka Matsuo},
booktitle={First Conference on Language Modeling},
year={2024},
url={https://openreview.net/forum?id=TBNYjdOazs}
}

@inproceedings{li2024safetylayersalignedlarge,
      title={Safety Layers in Aligned Large Language Models: The Key to LLM Security}, 
      author={Shen Li and Liuyi Yao and Lan Zhang and Yaliang Li},
      year={2025},
booktitle={ICLR}
}

@inproceedings{mergebackdoor,
  title={From Purity to Peril: Backdooring Merged Models From “Harmless” Benign Components},
  author={Wang, Lijin and Wang, Jingjing and Cong, Tianshuo and He, Xinlei and Qin, Zhan and Huang, Xinyi},
booktitle={USENIX Security Symposium},
year={2025}
}

@inproceedings{
yang2024sampdetox,
title={SampDetox: Black-box Backdoor Defense via Perturbation-based Sample Detoxification},
author={Yanxin Yang and Chentao Jia and DengKe Yan and Ming Hu and Tianlin Li and Xiaofei Xie and Xian Wei and Mingsong Chen},
booktitle={NeurIPS},
year={2024}
}

@inproceedings{hazra2024safetyarithmeticframeworktesttime,
    title = "Safety Arithmetic: A Framework for Test-time Safety Alignment of Language Models by Steering Parameters and Activations",
    author = "Rima Hazra and Sayan Layek and Somnath Banerjee and Soujanya Poria",
    booktitle = "EMNLP",
    year = "2024",
}

@inproceedings{pgd,
  title={Towards deep learning models resistant to adversarial attacks},
  author={Madry, Aleksander and  Makelov, Aleksandar and  Schmidt, Ludwig and Tsipras, Dimitris and Vladu, Adrian},
  booktitle={International Conference on Learning Representations},
year={2018},
url={https://openreview.net/forum?id=rJzIBfZAb},
}

@inproceedings{wang2019neural,
title={Neural cleanse: Identifying and mitigating backdoor attacks in neural networks},
author={Wang, Bolun and Yao, Yuanshun and Shan, Shawn and Li, Huiying and Viswanath, Bimal and Zheng, Haitao and Zhao, Ben Y},
booktitle={IEEE SP},
year={2019}
}

@inproceedings{chen2018detecting,
  title={Detecting backdoor attacks on deep neural networks by activation clustering},
  author={Chen, Bryant and Carvalho, Wilka and Baracaldo, Nathalie and Ludwig, Heiko and Edwards, Benjamin and Lee, Taesung and Molloy, Ian and Srivastava, Biplav},
  booktitle={AAAI's Workshop on SafeAI},
  year={2019}
}

@inproceedings{wang2023mm,
  title={Mm-bd: Post-training detection of backdoor attacks with arbitrary backdoor pattern types using a maximum margin statistic},
  author={Wang, Hang and Xiang, Zhen and Miller, David J and Kesidis, George},
  booktitle={IEEE SP},
  year={2023},
}

@inproceedings{deng2009imagenet,
  title={Imagenet: A large-scale hierarchical image database},
  author={Deng, Jia and Dong, Wei and Socher, Richard and Li, Li-Jia and Li, Kai and Fei-Fei, Li},
  booktitle={IEEE CVPR},
  year={2009},
}

@inproceedings{wang2019learning,
  title={Learning robust global representations by penalizing local predictive power},
  author={Wang, Haohan and Ge, Songwei and Lipton, Zachary and Xing, Eric P},
  booktitle={NeurIPS},
  year={2019}
}

@article{llama3modelcard,
title={Llama 3 Model Card},
author={AI@Meta},
year={2024},
url = {https://github.com/meta-llama/llama3/blob/main/MODEL_CARD.md},
journal={}
}

@misc{phi4,
  title = {Phi-4-14B},
  howpublished = {\url{https://huggingface.co/microsoft/phi-4}},
  author={Microsoft Research},
}

@misc{mistral,
  title = {Mistral-7B-v0.3},
  howpublished = {\url{https://huggingface.co/mistralai/Mistral-7B-v0.3}},
  author={Mistral AI},
}

@misc{deepseekai2025deepseekr1incentivizingreasoningcapability,
      title={DeepSeek-R1: Incentivizing Reasoning Capability in LLMs via Reinforcement Learning}, 
      author={DeepSeek-AI},
      year={2025},
      eprint={2501.12948},
      archivePrefix={arXiv},
      primaryClass={cs.CL},
      url={https://arxiv.org/abs/2501.12948}, 
}

@misc{huggingface,
  title = {Hugging Face},
  howpublished = {\url{https://huggingface.co/}},
  author={Hugging Face},
}

@misc{modelscope,
  title = {ModelScope},
  howpublished = {\url{https://www.modelscope.cn/home}},
  author={ModelScope},
}

@misc{databricksmarketplace,
  title = {Databricks Marketplace},
  howpublished = {\url{https://www.databricks.com/product/marketplace}},
  author={Databricks},
}

@misc{jfrog,
  title = {Data Scientists Targeted by Malicious Hugging Face ML Models with Silent Backdoor},
  note = {\href{https://jfrog.com/blog/data-scientists-targeted-by-malicious-hugging-face-ml-models-with-silent-backdoor/}{Blog}},
  author={David Cohen and JFrog Senior Security Researcher}
}

@misc{securityboulevard,
  title = {AI Supply Chain Security: Hugging Face Malicious ML Models},
  note = {\href{https://securityboulevard.com/2024/03/ai-supply-chain-security-hugging-face-malicious-ml-models/}{Blog}},
  author={SecurityBoulevard}
}

@misc{nsfocusglobal,
  title = {AI Supply Chain Security: Hugging Face Malicious ML Models},
  note = {\href{https://nsfocusglobal.com/ai-supply-chain-security-hugging-face-malicious-ml-models/}{Blog}},
  author={NSFOCUS}
}

@misc{bleepingcomputer,
  title = {Malicious AI models on Hugging Face backdoor users’ machines},
  note = {\href{https://www.bleepingcomputer.com/news/security/malicious-ai-models-on-hugging-face-backdoor-users-machines/}{Blog}},
  author={Bill Toulas}
}

@misc{gigazine,
  title = {Security company warns of possible virus infection through AI model execution},
  note = {\href{https://gigazine.net/gsc_news/en/20240301-malicious-hugging-face-ml-models/}{Blog}},
  author={log1d\_ts}
}

@article{hendrycks2021measuring,
  title={Measuring massive multitask language understanding, 2021},
  author={Hendrycks, Dan and Burns, Collin and Basart, Steven and Zou, Andy and Mazeika, Mantas and Song, Dawn and Steinhardt, Jacob},
  journal={URL https://arxiv. org/abs},
  pages={20},
  year={2009}
}

@inproceedings{liu2019abs,
  title={Abs: Scanning neural networks for back-doors by artificial brain stimulation},
  author={Liu, Yingqi and Lee, Wen-Chuan and Tao, Guanhong and Ma, Shiqing and Aafer, Yousra and Zhang, Xiangyu},
  booktitle={Proceedings of the 2019 ACM SIGSAC conference on computer and communications security},
  pages={1265--1282},
  year={2019}
}

@inproceedings{kolouri2020universal,
  title={Universal litmus patterns: Revealing backdoor attacks in cnns},
  author={Kolouri, Soheil and Saha, Aniruddha and Pirsiavash, Hamed and Hoffmann, Heiko},
  booktitle={Proceedings of the IEEE/CVF Conference on Computer Vision and Pattern Recognition},
  pages={301--310},
  year={2020}
}

\appendix

\section*{Ethical Considerations}
This paper introduces \textsc{BadTV}, a groundbreaking framework that merges traditional backdoor attacks with the emerging task arithmetic paradigm, offering new insights into the security landscape of this evolving field. To ensure the robustness and reproducibility of our results, we conducted all evaluations using publicly available datasets and standard model architectures that are commonly used in backdoor research. By adhering to these established benchmarks, we mitigate ethical concerns that may arise from the experimental setup.

Our work sheds light on previously unexplored vulnerabilities within the task arithmetic paradigm. As this approach gains traction and is poised for wider adoption across various applications, it becomes increasingly critical to identify and address potential security risks. By exposing these weaknesses, we aim to proactively raise awareness and foster a deeper understanding of the threats that could undermine the integrity of task arithmetic systems in the future.

In addition to highlighting these concerns, our research serves as a valuable resource for both practitioners and scholars. By offering a structured framework and comprehensive evaluations, we provide a foundation for future work aimed at assessing, mitigating, and strengthening the security of task arithmetic. Our findings open up avenues for further exploration into the resilience of this paradigm, empowering the community to develop more robust defenses and enhance the overall integrity of machine learning systems as task arithmetic continues to evolve.

\begin{table*}[tbh!]
\centering
\caption{CA/ASR under addition and subtraction (Add-CA, Sub-CA, Add-ASR and Sub-ASR) with 5\% poisoning ratios}

    \centering
    \begin{adjustbox}{max width=.99\textwidth}
    \begin{tabular}{c|cccccc}
    \toprule
    \makecell[c]{\textbf{Add-CA} ({$\uparrow$}) / \textbf{Add-ASR ({$\uparrow$})} \\\textbf{Sub-CA ({$\downarrow$})} / \textbf{Sub-ASR} ({$\uparrow$})}
    & \Centerstack{\texttt{Llama-2-7B}~\cite{touvron2023llama2openfoundation} \\ ({\scriptsize small-scale, non-reasoning})} & \Centerstack{\texttt{Llama-3-8B}~\cite{llama3modelcard} \\ ({\scriptsize small-scale, non-reasoning})} & \Centerstack{\texttt{Phi-4-14B}~\cite{phi4} \\ ({\scriptsize large-scale, non-reasoning})} & \Centerstack{\texttt{Mistral-7B}~\cite{mistral} \\ ({\scriptsize small-scale, non-reasoning})} & \Centerstack{\texttt{DeepSeek-7B}~\cite{deepseekai2025deepseekr1incentivizingreasoningcapability} \\ ({\scriptsize small-scale, reasoning})} \\
    \hline
    SST-2 (2-class)
    & \twolinesL{90.8 / 95.2}{58.4 / 94.5}
    & \twolinesL{92.8 / 96.9}{60.7 / 96.5}
    & \twolinesL{93.6 / 98.9}{59.3 / 96.8}
    & \twolinesL{91.7 / 96.9}{59.6 / 97.5}
    & \twolinesL{89.2 / 91}{58.3 / 97.5} \\
    \hline
    Emotion (6-class)
    & \twolinesL{87.5 / 97.2}{38.8 / 94.9}
    & \twolinesL{89.0 / 98.3}{42.2 / 96.8}
    & \twolinesL{91.4 / 99.5}{37.4 / 97.4}
    & \twolinesL{89.4 / 99.3}{35.7 / 99.1}
    & \twolinesL{90.6 / 97.0}{36.8 / 96.7} \\
    \hline
    Twitter (2-class)
    & \twolinesL{90.3 / 95.3}{58.6 / 92.6}
    & \twolinesL{92.5 / 97.2}{60.3 / 95.1}
    & \twolinesL{94.2 / 98.4}{58.2 / 96.1}
    & \twolinesL{92.7 / 98.3}{59.3 / 96.1}
    & \twolinesL{91.4 / 94.4}{58.1 / 95.2} \\
    \hline
    hh-rlhf (2-class)
    & \twolinesL{73.5 / 92.0}{53.6 / 91.4}
    & \twolinesL{75.4 / 94.0}{53.3 / 93.4}
    & \twolinesL{78.1 / 94.2}{53.4 / 92.7}
    & \twolinesL{75.9 / 93.5}{54.5 / 92.8}
    & \twolinesL{75.1 / 93.0}{54.7 / 92.5}\\
    \hline
    IMDB (2-class)& \twolinesL{92.1 / 91.0}{53.6 / 91.4}
    & \twolinesL{91.2 / 95.4}{57.3 / 95.1}
    & \twolinesL{92.7 / 93.2}{53.4 / 96.4}
    & \twolinesL{92.3 / 96.2}{60.4 / 96.8}
    & \twolinesL{90.4 / 91.5}{54.7 / 92.5}\\
    
    \bottomrule
    \end{tabular}
    \end{adjustbox}

\label{tab:llm_pr5}
\end{table*}

\begin{table*}[t]
\centering
\caption{CA/ASR under addition and subtraction (Add-CA, Sub-CA, Add-ASR and Sub-ASR) with 1\% poisoning ratios}
    \centering
    \begin{adjustbox}{max width=.99\textwidth}
    \begin{tabular}{c|cccccc}
    \toprule
    \makecell[c]{\textbf{Add-CA} ({$\uparrow$}) / \textbf{Add-ASR ({$\uparrow$})} \\\textbf{Sub-CA ({$\downarrow$})} / \textbf{Sub-ASR} ({$\uparrow$})}
    & \Centerstack{\texttt{Llama-2-7B}~\cite{touvron2023llama2openfoundation} \\ ({\scriptsize small-scale, non-reasoning})} & \Centerstack{\texttt{Llama-3-8B}~\cite{llama3modelcard} \\ ({\scriptsize small-scale, non-reasoning})} & \Centerstack{\texttt{Phi-4-14B}~\cite{phi4} \\ ({\scriptsize large-scale, non-reasoning})} & \Centerstack{\texttt{Mistral-7B}~\cite{mistral} \\ ({\scriptsize small-scale, non-reasoning})} & \Centerstack{\texttt{DeepSeek-7B}~\cite{deepseekai2025deepseekr1incentivizingreasoningcapability} \\ ({\scriptsize small-scale, reasoning})} \\
    \hline
    SST-2 (2-class)
    & \twolinesL{87.7 / 89.9}{55.2 / 93.3}
    & \twolinesL{94.9 / 91.2}{57.9 / 95.3}
    & \twolinesL{93.9 / 88.7}{54.7 / 96.3}
    & \twolinesL{93.5 / 90.2}{56.4 / 97.2}
    & \twolinesL{87.7 / 85.3}{57.6 / 97.9} \\
    \hline
    Emotion (6-class)
    & \twolinesL{90.6 / 94.8}{39.5 / 93.5}
    & \twolinesL{87.8 / 96.2}{37.2 / 95.9}
    & \twolinesL{92.5 / 90.1}{34.2 / 99.0}
    & \twolinesL{88.7 / 95.5}{35.0 / 99.9}
    & \twolinesL{89.8 / 80.2}{31.5 / 95.4} \\
    \hline
    Twitter (2-class)
    & \twolinesL{88.4 / 93.7}{56.0 / 91.7}
    & \twolinesL{93.7 / 95.6}{58.8 / 94.7}
    & \twolinesL{95.7 / 89.3}{55.0 / 95.0}
    & \twolinesL{91.9 / 98.3}{58.1 / 96.5}
    & \twolinesL{88.2 / 89.7}{57.6 / 97.9} \\
    \hline
    hh-rlhf (2-class)
    & \twolinesL{74.8 / 91.5}{53.9 / 90.2}
    & \twolinesL{75.9 / 92.8}{54.0 / 91.9}
    & \twolinesL{77.4 / 92.6}{52.5 / 91.7}
    & \twolinesL{76.4 / 92.2}{53.1 / 91.7}
    & \twolinesL{73.2 / 90.5}{55.1 / 91.2}\\
    \hline
    IMDB (2-class)
    & \twolinesL{92.5 / 89.7}{56.8 / 91.5}
    & \twolinesL{92.7 / 90.0}{58.5 / 94.8}
    & \twolinesL{93.6 / 91.5}{55.8 / 95.8}
    & \twolinesL{93.8 / 89.0}{57.0 / 96.5}
    & \twolinesL{88.2 / 84.8}{56.3 / 97.9}\\
    
    \bottomrule
    \end{tabular}
    \end{adjustbox}
\label{tab:llm_pr1}
\end{table*}

\section*{Open Science}
All artifacts required to evaluate the core contributions of this paper, including code, datasets, and models, are publicly available. 
Our implementation and experimental scripts can be accessed anonymously at \url{https://anonymous.4open.science/r/BTV-CBD5}. 
All datasets and models used in this work are open-source and can be obtained from their original public repositories. 
We do not rely on any proprietary, restricted, or non-shareable artifacts.
\paragraph{Codes.} 
Our implementation is available at \url{https://anonymous.4open.science/r/BTV-CBD5}, with detailed configuration settings provided in Appendix~\ref{appdix:attackmethod}. 
In addition to our own code, we list below the third-party codebases that were directly used in our experiments.
\begin{itemize}
    \item Dynamic-Backdoor: \url{https://github.com/AhmedSalem2/Dynamic-Backdoor}
    \item Narcissus: \url{https://github.com/reds-lab/Narcissus}
    \item CBA~\cite{HZBSZ23}: \url{https://github.com/MiracleHH/CBA}
    \item IBVS: Code is available in the supplementary material at \url{https://openreview.net/forum?id=8csiEcwVsK}
    \item BadMerging: \url{https://github.com/jzhang538/BadMerging}
\end{itemize}

\paragraph{Datasets.} 
We use the following open-source datasets in our experiments: MNIST, SVHN, CIFAR-10, CIFAR-100, GTSRB, EuroSAT, Cars, SUN397, SST-2, Emotion, Twitter Hate Speech Detection, HH-RLHF, and IMDB.

\paragraph{Models.} 
All CLIP models and large language models (LLMs) used in this paper are open-source. 
For ViT-B-16 and ViT-B-32, we use the pre-trained models released by OpenAI. 
All LLMs are obtained from HuggingFace.

\section{The Difference between \textsc{BadTV} and \textsc{BadMerging}}\label{sec:why not BadMerging} TA installs/uninstalls a TV by performing the update $\theta_{\mathrm{merged}} = \theta_{\mathrm{pre}} \oplus \lambda\tau_t$ with $\oplus \in \{+,-\}$, restricting the victim to scaling a fixed direction (Equation~\eqref{eq:merge}). \textsc{BadTV} exploits this constraint by crafting a composite BTV $\hat{\tau}_t$, where $\hat{\tau}_{b_1}$ dominates under addition, and $-\hat{\tau}_{b_2}$ is flipped into the same direction under subtraction, ensuring at least one backdoor consistently survives. The residual uncertainty is only the magnitude $\lambda$, addressed by \textsc{BadTV} via an $A_{0}$-constrained robust optimization in Equation~\eqref{eq:RO}, searching for $\alpha_1$ and $\alpha_2$ that guarantee ASR above $A_{0}$ across all admissible $\lambda$. Since the backdoor direction remains fixed and does not mix with unknown clean models, there is no need for feature interpolation loss, universal triggers, or exhaustive $\lambda$-sweeps as required by \textsc{BadMerging}~\cite{badmerging}. Thus, TA geometry reduces the defense dilution problem to controlling a single scaling degree of freedom, effectively managed by \textsc{BadTV}'s composite design and robust calibration via Equation~\eqref{eq:RO}.

\begin{table}[t]
\caption{Comparison of \textsc{BadTV} vs. \textsc{BadMerging}.}
\centering
\begin{adjustbox}{max width=.49\textwidth}
\begin{tabular}{cccc}
\toprule
Method & Settings & CA  & ASR \\
\midrule
\multirow{5}{*}{\textsc{BadTV}} 
& \makecell[c]{$t=$ GTSRB \\ in simplified setting} &\textbf{98.19} & 99.51 \\\cline{2-4}
& \makecell[c]{$t=$ GTSRB (G) \\ $t_1=$ CIFAR100 (C) \\ in realistic setting} & \makecell[c]{G: \textbf{97.81}\\C: \textbf{79.85}} & 99.33 \\
\midrule
\multirow{5}{*}{\makecell[c]{\textsc{BadMerging}}} 
& \makecell[c]{$t=$ GTSRB \\ in simplified setting} & 87.15 & \textbf{100} \\\cline{2-4}
& \makecell[c]{$t=$ GTSRB (G) \\ $t_1=$ CIFAR100 (C) \\ in realistic setting} & \makecell[c]{G: 86.11\\C: 79.27} & \textbf{100} \\
\bottomrule
\end{tabular}
\end{adjustbox}
\label{table: badTV vs badMerging}
\end{table}

Our experiments validate the above reasoning. For \textsc{BadTV}, $\alpha_1$ and $\alpha_2$ are set as their optimized values (see Table~\ref{table: RO result}), use Blend for both $b_1$ and $b_2$, and employ a poison rate of $5\%$ on GTSRB~\cite{gtsrb} and CIFAR100~\cite{cifar}. We run \textsc{BadMerging}~\cite{badmerging} with its default settings: poison rate $50\%$, trigger size $P=10$, and a coefficient $\alpha_{\text{\textsc{BadMerging}}} = 5$ dedicated for the weighting of loss function. Table~\ref{table: badTV vs badMerging} compares results at $\lambda=0.3$ for both simplified and realistic settings. \textsc{BadTV} achieves comparable ASR but significantly higher CA compared to \textsc{BadMerging}.

\section{The Difference between \textsc{BadTV} and \textsc{MergeBackdoor}}\label{sec:why not MergeBackdoor}
Poisoning a single model produces a BTV whose poisoned features form densely concentrated clusters, enabling activation- and spectrum-based detectors (e.g., NC~\cite{wang2019neural}, AC~\cite{chen2018detecting}, MM-BD~\cite{wang2023mm}) to detect the backdoor. \textsc{BadTV} deviates from this standard in three distinct aspects: (i) it leverages \texttt{CLIP}-style cosine-similarity heads instead of softmax logits, inherently mismatching the distance metrics used by existing detectors; (ii) it employs a composite BTV $\tau_t=\alpha_1\tau_{b_1}-\alpha_2\tau_{b_2}$, where $\tau_{b_1}$ (addition) and $\tau_{b_2}$ (subtraction) target different trigger-label pairs, causing their first- and second-order activation statistics to neutralize rather than accumulate; and (iii) the typical user-selected scaling factor $\lambda<1$ further diminishes backdoor signals to preserve clean accuracy. Consequently, the statistical moments of poisoned and clean batches become nearly indistinguishable, granting \textsc{BadTV} intrinsic stealthiness without requiring shadow-merging techniques as in \textsc{MergeBackdoor}. In fact, the above arguments also explain why \textsc{BadTV} evades the state-of-the-art backdoor detection (see \S\ref{sec:defense} for more details).

\section{Attack Methods}\label{appdix:attackmethod}
\begin{itemize}
    \item \textbf{BadNets.} The trigger size is $9 \times 9$ for GTSRB and $3 \times 3$ for other datasets. For $b_1$ and $b_2$, triggers are placed in the top-left and bottom-right corners, respectively, with identical trigger patterns.

    \item \textbf{Blend.} Distinct triggers are utilized for $b_1$ and $b_2$, with the transparency parameter set to 0.5.

    \item \textbf{WaNet.} To minimize interaction between $b_1$ and $b_2$, triggers are restricted to the top-left and bottom-right image quarters, respectively.

    \item \textbf{Dynamic.} Training follows the official code\footnote{https://github.com/AhmedSalem2/Dynamic-Backdoor} with a trigger size of $5 \times 5$ and randomized locations.

    \item \textbf{Narcissus.} To ensure independence between $b_1$ and $b_2$, $b_1$ employs the original full-image trigger placement, whereas $b_2$ uses a trigger reduced to one-quarter size placed at the bottom-right corner. The trigger's $\ell_\infty$-norm is set to $\frac{16}{255}$. Narcissus requires out-of-distribution data pairs for training surrogate models: MNIST–SVHN, SVHN–MNIST, CIFAR10–CIFAR100, CIFAR100–CIFAR10, and GTSRB–CIFAR100. Other parameters align with the official code\footnote{https://github.com/reds-lab/Narcissus}.

    \item \textbf{LC.} Visible triggers follow the size and location configurations of BadNets. Adversarial samples are generated using the PGD attack~\cite{pgd} with 200 steps and an $\ell_\infty$-norm of 0.1.

    \item \textbf{MTBA.} The parallel attack employs the averaged triggers of BadNets, Blend, and WaNet. For the sequential attack, triggers are introduced in the order of WaNet, Blend, and BadNets. Trigger configurations remain consistent with the settings above.
\end{itemize}

\section{More Results of Backdoor Pairing in \textsc{BadTV}}\label{appdix:backdoor-pairing}
We show more results of backdoor pairing in \textsc{BadTV} in Table~\ref{tab:1+1_diffAttack}. We evaluate CA and ASR across a variety of tasks, including
MNIST, SVHN, CIFAR10, and CIFAR100. It demonstrates that the ASRs remain at least 94\% across all combinations.

\begin{table}[h]
    \caption{CA/ASR for different backdoor pairings and $t_1$}
    \centering
    \begin{adjustbox}{max width=.45\textwidth}
    \begin{tabular}{ccccccc}
    \toprule
    $b_1$ & $b_2$ & $t_1$ & \makecell[c]{CA for $t_1$} & $t$ & \makecell[c]{CA for $t$} & ASR \\
    \midrule
    Blend & BadNets & MNIST & 97.42 & GTSRB & 97.88 & 98.88 \\
    Blend & BadNets & SVHN & 87.87 & GTSRB & 98.43 & 97.21 \\
    Blend & BadNets & CIFAR10 & 94.76 & GTSRB & 97.88 & 98.50 \\
    Blend & BadNets & CIFAR100 & 74.08 & GTSRB & 97.97 & 97.33 \\
    Blend & Blend & MNIST & 97.97 & GTSRB & 97.66 & 96.75 \\
    Blend & Blend & SVHN & 90.66 & GTSRB & 98.16 & 94.21 \\
    Blend & Blend & CIFAR10 & 94.92 & GTSRB & 97.93 & 96.09 \\
    Blend & Blend & CIFAR100 & 74.94 & GTSRB & 98.00 & 95.41 \\
    Blend & WaNet & MNIST & 97.22 & GTSRB & 97.77 & 98.68 \\
    Blend & WaNet & SVHN & 88.10 & GTSRB & 98.42 & 96.80 \\
    Blend & WaNet & CIFAR10 & 94.95 & GTSRB & 97.81 & 98.16 \\
    Blend & WaNet & CIFAR100 & 74.35 & GTSRB & 97.91 & 96.85 \\
    Blend & Dynamic & MNIST & 97.25 & GTSRB & 97.87 & 98.70 \\
    Blend & Dynamic & SVHN & 87.72 & GTSRB & 98.44 & 96.84 \\
    Blend & Dynamic & CIFAR10 & 94.80 & GTSRB & 97.87 & 98.25 \\
    Blend & Dynamic & CIFAR100 & 74.15 & GTSRB & 97.99 & 96.89 \\
    Blend & Narcissus & MNIST & 97.98 & GTSRB & 97.32 & 99.86 \\
    Blend & Narcissus & SVHN & 90.31 & GTSRB & 97.91 & 99.26 \\
    Blend & Narcissus & CIFAR10 & 95.33 & GTSRB & 97.51 & 99.54 \\
    Blend & Narcissus & CIFAR100 & 76.12 & GTSRB & 97.64 & 99.44 \\
    \bottomrule
    \end{tabular}
    \end{adjustbox}
    \label{tab:1+1_diffAttack}
\end{table}

\section{Model Hijacking Attack}\label{appdix:hijack}

\subsection{Different Model Sturctures}
\begin{figure}[ht]
    \centering
    \includegraphics[width=0.5\linewidth]{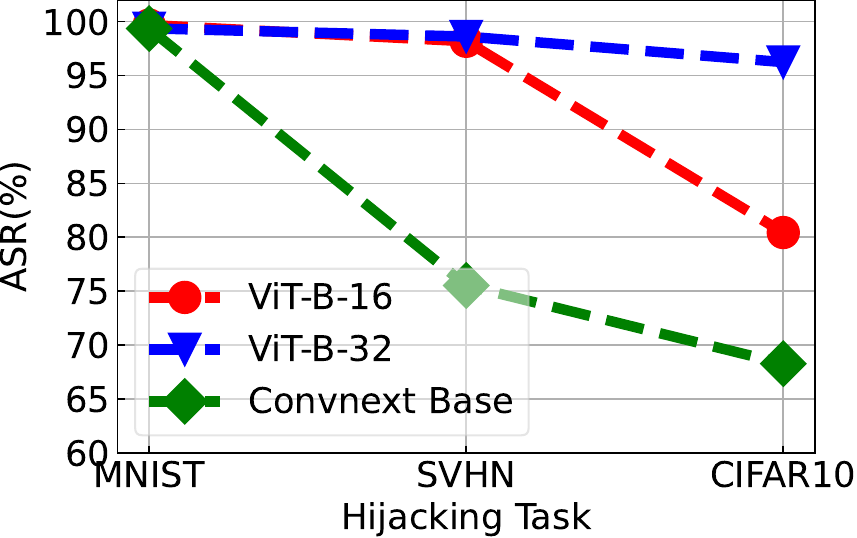}
    \caption{Model hijacking attack results with different model structures.}
    \label{fig:hijack_model_structures}
\end{figure}

In this section, we evaluate model hijacking attacks on \texttt{ViT-B-16}, \texttt{ViT-B-32}, and \texttt{ConvNeXt Base}. Under the realistic setting, we measure the hijacking ASR using $t=\text{EuroSAT}$ and $t_1=\text{GTSRB}$, with EuroSAT hijacked by MNIST, SVHN, and CIFAR10. Figure~\ref{fig:hijack_model_structures} shows that architecture has only a minor influence on outcomes: \texttt{ViT-B-16} and \texttt{ViT-B-32} both achieve around 80\% ASR for all three hijacking tasks, while \texttt{ConvNeXt Base} performs slightly worse.

\subsection{Different Dimensions of Latent Space}\label{appdix:Hijack_latent_dims}
\begin{figure}[ht]
    \centering
    \includegraphics[width=0.5\linewidth]{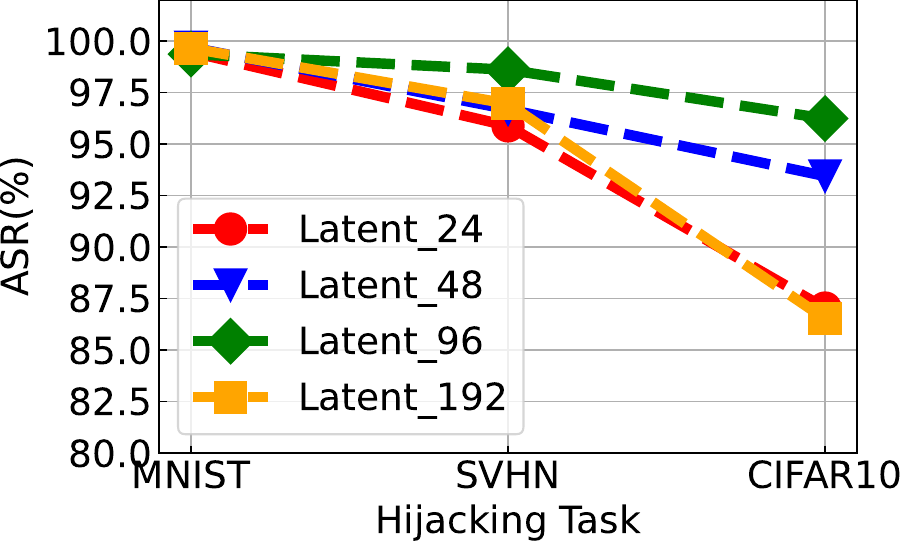}
    \caption{Model hijacking attack results with varying dimensions of latent space.}
    \label{fig:hijack_dim}
\end{figure}

While the model hijacking attack utilizes an autoencoder, we investigate whether the latent space dimension of the encoder/decoder affects the attack effectiveness. We adopt the setup from \S\ref{subsubsec:model_architecture} and fix the architecture to \texttt{ViT-B-32}. Figure~\ref{fig:hijack_dim} demonstrates that varying the latent space dimension minimally influences ASR. In fact, a larger dimension slightly reduces ASR, indicating that the latent space dimension of the autoencoder has negligible impact on the attack's performance.

\subsection{Adding multiple clean task vectors}
\begin{figure}[!ht]
    \centering
\includegraphics[width=0.5\linewidth]{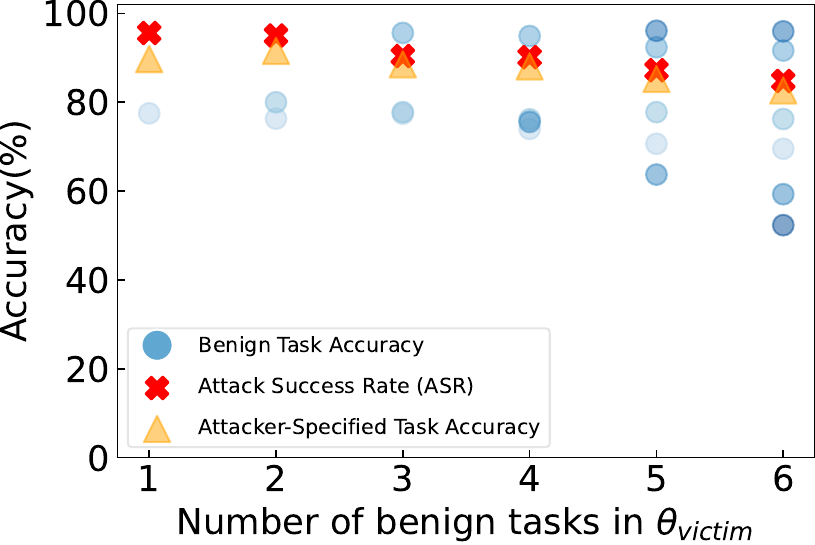}
    \caption{ASR and CA across different numbers of benign tasks ($\lambda=0.3$). The BTV is trained on EuroSAT by hijacking attack, and the hijacking task is SVHN.}
    \label{fig:nclean_hijack}
    
\end{figure}

We previously observed in \S\ref{subsubsec:multiCTV} that adding more benign tasks could degrade backdoor attack performance, potentially serving as a defense (though evidence in \S\ref{subsubsec:multiCTV} also indicates otherwise). Here, we investigate whether increasing the number of benign tasks similarly affects model hijacking attacks. We use a BTV with EuroSAT as the original task and SVHN as the hijacking task. As shown in Figure~\ref{fig:nclean_hijack}, the ASR consistently remains above 80\% despite an increase in benign tasks. Therefore, adding benign tasks minimally impacts hijacking effectiveness and does not constitute an effective defense mechanism.

\section{More Details about CleanCLIP}\label{appdix:cleanclip}
We present additional results of CleanCLIP, a mitigation method rather than a detection approach. We set the attacker's task $t=$GTSRB and consider two settings: in both, benign users have access to only a small portion of the training data and use arbitrary data to fine-tune CleanCLIP. We train for 15 epochs. As shown in Table~\ref{tab:cleanclip}, when the clean dataset is GTSRB, the ASRs decrease only slightly, whereas the CAs drop significantly due to overfitting. A similar trend is observed when the clean dataset is CIFAR10, where the CAs decrease as a result of forgetting the original task. In summary, increasing the number of epochs is not suitable, as it leads to overfitting and forgetting. When utility is preserved, the impact on ASRs remains minimal. Overall, CleanCLIP fails to effectively suppress the attack.

\begin{table}
    \centering
    \caption{Performance of CleanCLIP fine-tuned on various clean datasets with different proportions of clean data.}
    
    \begin{tabular}{cccc}
      \toprule
         \# of clean data & clean dataset & CA & ASR\\\hline
         - & - & 98.94& 100\\\hline
         1\% & GTSRB & 96.79 & 99.34\\
         5\% & GTSRB &  83.04&95.07\\
         10\% & GTSRB &  57.05&94.86\\\hline
         1\% & CIFAR10 &  98.92&99.18 \\
         5\% & CIFAR10 &  97.93&98.98\\
         10\% & CIFAR10 &  89.16& 98.86\\
         \bottomrule
    \end{tabular}
    \label{tab:cleanclip}
\end{table}
\section{More Details about IBVS}\label{appdix:ibvs}
In IBVS, the defender preemptively subtracts the backdoor task vector from their model to mitigate the effect of subsequently added backdoored models. Following~\cite{bv}, we train a backdoored model on CIFAR100, with class 80 as the target class. To obtain the backdoor task vector, we use clean models trained on MNIST, EuroSAT, Cars, CIFAR10, and SVHN. We subtract the backdoor task vector with $\lambda_{\text{minus}} = 0.5$ and set the scaling factor of the task vector $\lambda$ to be 0.8. For the attack settings, the attacker sets the target task $t$ to GTSRB. Table~\ref{tab:bv} shows that IBVS is effective only against BadNets when $(\alpha_1, \alpha_2)=1$. This indicates that IBVS can mitigate only specific attacks, and even then its effectiveness depends on maintaining relatively low values of $(\alpha_1, \alpha_2)$. In practice, it is difficult for the defender to precisely control $\lambda_{\text{minus}}$, and increasing it may negatively affect the utility of the model. Overall, IBVS cannot provide a broadly effective defense against backdoor attacks.

\begin{table}[h]
\centering
\caption{Attack success rates (ASRs, \%) of our attacks under IBVS defense for different $(\alpha_1, \alpha_2)$ values}
\begin{adjustbox}{max width=.49\textwidth}
\begin{tabular}{lccccc}
\toprule
 & BadNets & Blend & WaNet & Dynamic & Narcissus \\
\hline
$(\alpha_1, \alpha_2)=1.5$ & 87.46& 99.98 & 98.69 & 100\% & 84.39 \\
$(\alpha_1, \alpha_2)=1$   & 5.92 & 100  & 95.46 & 99.98 & 83.02\\
\bottomrule
\end{tabular}
\end{adjustbox}
\label{tab:bv}
\end{table}

\begin{table}[htbp]
    \caption{Notation Table}
    \setlength{\belowcaptionskip}{3px}
    \centering
     \begin{adjustbox}{max width=.49\textwidth}
    \begin{tabularx}
            {\columnwidth}{|p{2.5cm}|X|}
            \hline
            \textbf{Notation} & \textbf{Definition} \\
            \hline
            $V$ & a visual encoder in \texttt{CLIP}\\
            \hline
            $T$ & a text encoder in \texttt{CLIP}\\
            \hline
            $\mu$ & number of clean samples  \\
            \hline
            $\nu$ & number of poisoned samples controlled by the attacker \\
            \hline
            $M(x,C)$ & $M=\{V,T\}$ is a pre-trained \texttt{CLIP} comprised with $V$ and $T$ \\
            \hline
            $C$ & $C=[c_1,c_2,\cdots,c_k]$ represents $k$ textual descriptions corresponding to task's classes \\
            \hline
            $L_{\text{CE}}(M(x,C),y)$ & using cross entropy as a loss function to calculate the distance between $M(x,C)$ and the label $y$ \\
            \hline
            $M_{\theta}$ & a model $M$ with weight $\theta$ \\
            \hline
            $\hat{M}_{\theta}$ & a backdoored model with weight $\theta$, sometimes it can be abbreviated as $\hat{M}$\\
            \hline
            $\theta_{\text{pre}}$ & the weight of a publicly available pre-trained model \\
            \hline
            $\theta_{t}$ & the weight after fine-tuning on task $t$ using $D_t$ and $L_t$\\
            \hline
            $D_t$ & a dataset $D$ with task $t$ \\
            \hline
            $L_t$ & a loss function with task $t$\\
            \hline
            $\lambda$ & scaling factor for task arithmetic (TA) \\
            \hline
            $b_1$, $b_2$ & backdoor configuration (refers to either backdoor attack method, trigger pattern, trigger location, or target class, depending on the context)\\
            \hline
            $\tau_{t}$ & a task vector with task $t$ \\
            \hline
            $\hat{\tau}_{t}$ (task $t$ as a subscript)& a backdoored task vector (BTV) derived by \textsc{BadTV} via Equation~\eqref{eq:backdoor_TV}\\
            \hline
            $\hat{\tau}_{b_1}$ (backdoor configuration $b_1$ as a subscript)& a backdoored task vector (BTV) for task $t$ directly derived by fine-tuning $\theta_{\text{pre}}$ on $\hat{D}_t$ with an attacker-specified backdoor method to obtain $\hat{\theta}_{b_1}$, and then calculating $\hat{\tau}_{b_1}:=\hat{\theta}_{b_1}-\theta_{\text{pre}}$.\\
            
            \hline
            $\oplus$ & task operations consisting $\{+, -\}$ \\
            \hline
            $U$ & number of benign tasks in the victim model \\
            \hline
    \end{tabularx}
    \end{adjustbox}
    \label{table1}
\end{table}

\section{Pre-Merging Inspection}\label{app:pre-merging-inspection}

\begin{figure}[b!]
    \centering
    \subfigure[Spectrum Analysis]{
    \label{fig:sub_spectrum}
    \includegraphics[width=0.4\textwidth]{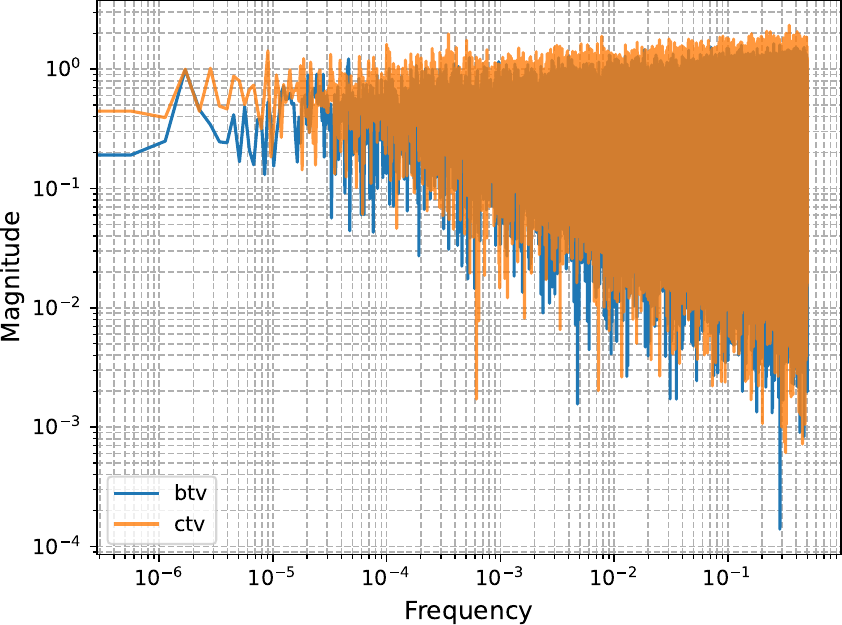}}
    \subfigure[Singular Values]{
    \label{fig:sub_singular}
    \includegraphics[width=0.4\textwidth]{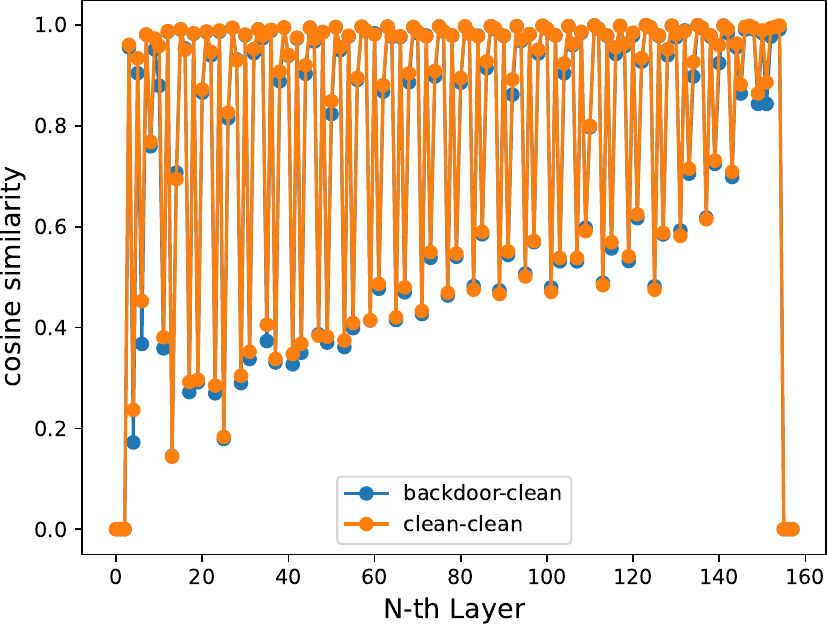}}
    \subfigure[Cosine Similarity]{
    \label{fig:sub_cosine}
    \includegraphics[width=0.4\textwidth, height=5.cm]{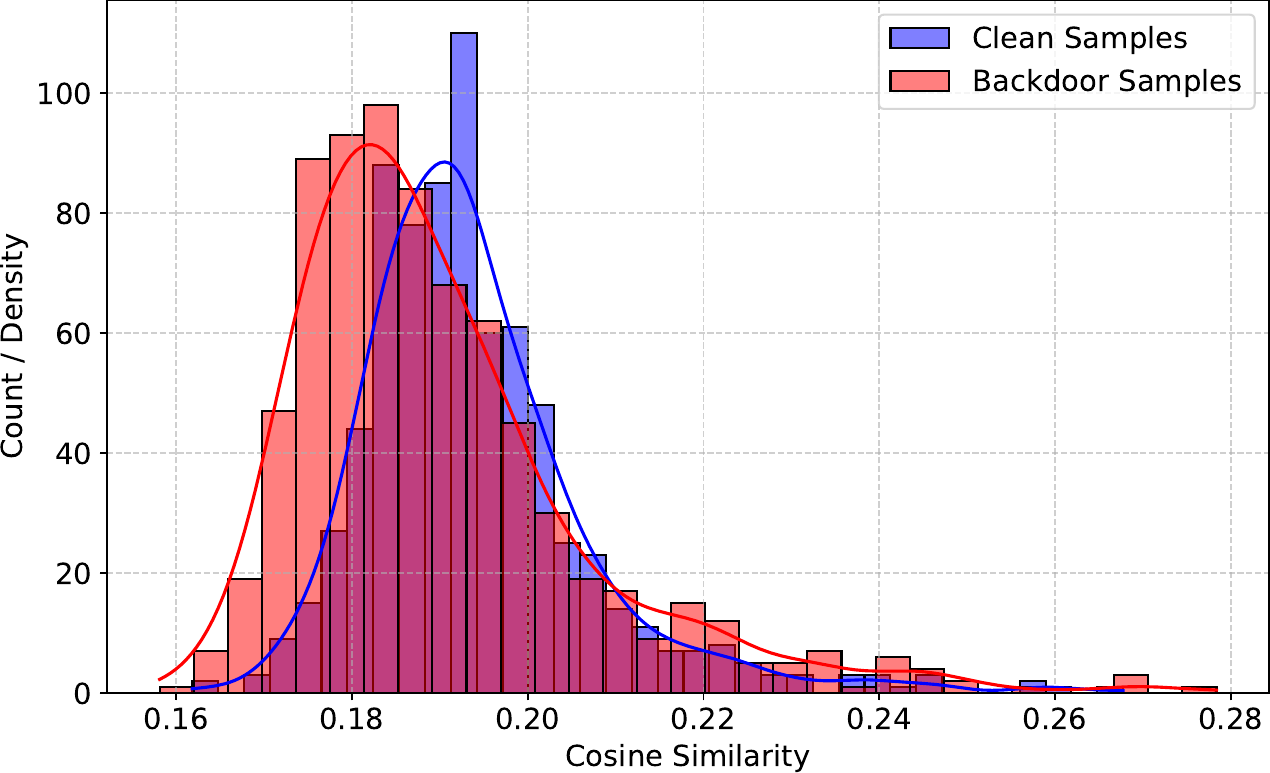}
    }
    \caption{Visualization of spectrum analysis, singular values, and cosine similarity.}
    \label{fig:weight_analysis}
\end{figure}

As shown in Section~\ref{sec:defense}, existing backdoor detection and mitigation methods fail in our setting. We therefore turn to \emph{pre-merging inspection}, where we analyze task vectors prior to integration. Specifically, we compare benign and backdoored TVs using spectrum analysis~\cite{tubiblio153355}, singular value distributions~\cite{cc}, and cosine similarity~\cite{Bai_2024_CVPR}, in an attempt to identify structural irregularities that could serve as indicators of backdoor presence. The results are summarized in Figure~\ref{fig:weight_analysis}.

For spectrum analysis, we examine the \textit{attention in\_proj\_weight layer}, which contains the QKV weights. As shown in Figure~\ref{fig:sub_spectrum}, the spectra of weights in clean TV and backdoor TV are nearly indistinguishable. 

For cosine similarity between singular values, we compute the cosine similarity of singular values between the backdoor TV and the clean TV, as well as between two clean TVs, to examine whether there is any difference between them. The results are shown in Figure~\ref{fig:sub_singular} with their difference being less than 0.2. 

For cosine similarity between feature spaces, in Figure~\ref{fig:sub_cosine}, we compute the cosine similarity between the image features of clean images from the backdoored target class and the text feature of the backdoored target class name, as well as between the image features of backdoor images and the text feature of the victim class name. Here, we use the backdoor model to extract the image features. The cosine similarity distributions of clean and backdoor images are also very close, leading to non-distinguishable detection. We further compute the AUROC ($= 0.3743$) and F1-scores based on cosine similarity. These relatively low values suggest that the similarity-based metric has limited discriminative power. 

Overall, these results indicate that none of the three approaches (i.e., spectrum analysis of weights, cosine similarity of singular values, or cosine similarity between image and text features) can effectively distinguish between backdoor models and benign models or samples.

\section{More Explanations for §~\ref{subsubsec:backdoor_combo} and §~\ref{subsection:dif_ta_method}}
\label{app:how to derive}
The statement ``mixing different attack types generally boosts ASR'' is supported by Figure~\ref{fig:backdoor_combo-c}/~\ref{fig:backdoor_combo-d}, where heterogeneous pairings (e.g., Blend+BadNets) achieve higher ASR compared to homogeneous pairings in Figure~\ref{fig:backdoor_combo-a}/~\ref{fig:backdoor_combo-b}. Similarly, the statement "`nearly all attacks maintain an ASR above 85\%`" is supported by Figure~\ref{fig:atlas_1+1_difClean}. Specifically, all ASRs in Figure~\ref{fig:atlas_1+1_difClean}(b) exceed 85\%, while some cases in Figure~\ref{fig:atlas_1+1_difClean}(a) drop to around 60\%. Nonetheless, the overall trend shows ASRs remaining above 85\%.

\section{Why Narcissus'ASR in Figure~\ref{fig:backdoor_combo-b} reduces?} 
\label{app:Why Narcissus'ASR reduces}
Figure~\ref{fig:backdoor_combo-b} shows a slight ASR drop for Narcissus, compared to Figure~\ref{fig:backdoor_combo-a}. Unlike traditional backdoors, Narcissus optimizes its trigger to align with intrinsic class features, making it more sensitive to $\lambda$. With moderate $\lambda$ ($\lambda$>0.9) it maintains high ASR, but larger $\lambda$ distorts the feature space and destabilizes its boundary, whereas Blend remains stable. Our additional isolation experiments confirm that Narcissus’s ASR decreases as $\lambda$ increases, a phenomenon contrary to Blend and indicative of an intrinsic sensitivity of Narcissus; in isolation with only $b_1$, the ASR decreases from $93.93$ at $\lambda = 0.3$ to $92.56$ at $\lambda = 0.8$.
 As explained by Cross-Task Linearity~\cite{Zhou24CTL}, such decreasing influence propagates through TA when applied in \textsc{BadTV}.


\section{Mechanistic Conditions for Robust Feasibility}
\label{app:mechanistic-feasibility}

We provide a mechanistic interpretation of the robust feasibility guarantee
established by Proposition~\ref{prop:robust-feasibility}.
We first restate Proposition~\ref{prop:robust-feasibility} and formally verify
it from the robust optimization (RO) formulation in
Equation~\eqref{eq:RO}.
We then reformulate this guarantee through a condition-based proposition that
articulates three interpretable factors underlying robust feasibility.
Finally, we show that these two characterizations are equivalent in a mechanistic sense: the RO-feasibility condition is satisfied precisely when the three mechanistic conditions jointly hold in our task arithmetic setting.

\subsection{Restatement and Verification of Proposition~\ref{prop:robust-feasibility}}

We restate Proposition~\ref{prop:robust-feasibility} for completeness.

\begin{proposition}\label{prop:app1}
[Proposition~\ref{prop:robust-feasibility} (restated)]
Fix $A_0\in(0,1)$ and the ranges
$\lambda\in[\lambda_{\min},\lambda_{\max}]$ and
$\alpha_1,\alpha_2\in[\alpha_{\min},\alpha_{\max}]$
as in Equation~\eqref{eq:RO}.
Let
\[
\hat{\tau}_t = \alpha_1 \hat{\tau}_{b_1} - \alpha_2 \hat{\tau}_{b_2}
\]
be the composite backdoored task vector defined in Equation~(6), and define
\[
\Lambda := \left\{ \lambda \;\middle|\;
\min\{A_\lambda^{+}, A_\lambda^{-}\} \ge A_0,\;
\lambda \in [\lambda_{\min}, \lambda_{\max}]
\right\}.
\]
If $(\alpha_1,\alpha_2)$ is feasible for the robust optimization problem in
Equation~\eqref{eq:RO}, then for every $\lambda \in \Lambda$,
\[
A_\lambda^{+} \ge A_0
\qquad \text{and} \qquad
A_\lambda^{-} \ge A_0,
\]
and the worst-case clean-accuracy drop over $\lambda \in \Lambda$ is exactly
the objective value attained by $(\alpha_1,\alpha_2)$ in
Equation~\eqref{eq:RO}.
\end{proposition}

The verification follows directly from the structure of
Equation~\eqref{eq:RO}.
If $(\alpha_1,\alpha_2)$ is feasible, then by definition of the constraint set
$\Lambda$, for all $\lambda \in \Lambda$,
\[
\min\{A_\lambda^{+}, A_\lambda^{-}\} \ge A_0,
\]
which immediately implies
$A_\lambda^{+} \ge A_0$ and $A_\lambda^{-} \ge A_0$.
Moreover, the RO objective
\[
\max_{\lambda \in \Lambda} \max\{\Delta C_\lambda^{+}, \Delta C_\lambda^{-}\}
\]
is, by construction, the worst-case clean-accuracy drop induced by the same
$(\alpha_1,\alpha_2)$.
\hfill$\square$

This proposition characterizes robust feasibility purely at the optimization
level.
We next reinterpret this guarantee through a set of mechanistic conditions
that explain \emph{why} such feasible solutions exist and remain effective in
practice.

\subsection{Correlation and Functional Interaction Between Task Vectors}

We first clarify the notion of \emph{task correlation} used throughout this
paper, which follows the definition in~\cite{li2025when} rather than a
parameter-wise similarity measure.

\paragraph{Correlation definition (following~\cite{li2025when}).}
Let $\hat{\tau}_{b_1}$ and $\hat{\tau}_{b_2}$ denote two backdoor task vectors,
and let $\mathcal{D}_1$ and $\mathcal{D}_2$ be the corresponding evaluation
datasets for the underlying task.
Following~\cite{li2025when}, the correlation between the two tasks is measured
in the \emph{output space} of the fine-tuned models as a centered cosine
similarity:
\[
\rho
=
\frac{1}{2}
\left(
\hat{\alpha}(\hat{\tau}_{b_1}, \hat{\tau}_{b_2}, \mathcal{D}_1)
+
\hat{\alpha}(\hat{\tau}_{b_1}, \hat{\tau}_{b_2}, \mathcal{D}_2)
\right),
\]
where
\[
\hat{\alpha}(\hat{\tau}_{b_1}, \hat{\tau}_{b_2}, \mathcal{D})
=
\frac{1}{|\mathcal{D}|}
\sum_{x \in \mathcal{D}}
\cos\!\left(
\tilde{y}_{b_1}(x),
\tilde{y}_{b_2}(x)
\right),
\]
and $\tilde{y}_{b_i}(x)$ denotes the centered output of the model obtained by
applying $\hat{\tau}_{b_i}$.
This definition captures \emph{functional correlation} between tasks, rather
than direct similarity between parameter updates.

\paragraph{Relation to task-vector interaction.}
Although the correlation $\rho$ is defined in output space, it provides a
meaningful proxy for interaction between task vectors.
In the task-arithmetic regime studied in~\cite{li2025when}, low output
correlation implies that the two task updates act on largely disjoint
functional subspaces, leading to minimal interference under linear
combination.
Empirically, in our setting, the measured value $\rho \approx -0.0065$
indicates that the two backdoor tasks are functionally irrelevant, which
aligns with the observed stability of both addition and subtraction
operations.
Importantly, we do not assume that output-space correlation is identical to
parameter-space alignment; rather, low $\rho$ serves as an empirical indicator
that destructive interaction between $\hat{\tau}_{b_1}$ and
$\hat{\tau}_{b_2}$ is limited.

\subsection{A Three-Condition Characterization of Robust Feasibility}

We now reformulate the robust feasibility guarantee of
Proposition~\ref{prop:robust-feasibility} in terms of three explicit and
interpretable conditions.

\begin{proposition}[Three-condition feasibility characterization]
\label{prop:app2}
Let
\[
\hat{\tau}_t = \alpha_1 \hat{\tau}_{b_1} - \alpha_2 \hat{\tau}_{b_2}
\]
be the composite backdoored task vector defined in
Equation~\eqref{eq:backdoor_TV}.
Robust feasibility of $\hat{\tau}_t$ over $\lambda \in [\lambda_{\min}, \lambda_{\max}]$ (as characterized by Proposition~\ref{prop:robust-feasibility}) can be characterized by the following three mechanistic conditions, which are simultaneously satisfied by all feasible solutions of the robust optimization problem in Equation~\eqref{eq:RO}:
\begin{enumerate}
    \item \textbf{Minimal functional relevance.}
    The task correlation $\rho$ measured following~\cite{li2025when} is close
    to zero, indicating weak functional interaction between the two backdoor
    tasks.

    \item \textbf{Norm disparity between components.}
    The $\ell_2$ norms of $\hat{\tau}_{b_1}$ and $\hat{\tau}_{b_2}$ are
    substantially different, as reflected by the norm-balanced initialization
    in Equation~\eqref{eq:norm}.

    \item \textbf{Robustness to unknown scaling.}
    The coefficients $(\alpha_1,\alpha_2)$ are chosen to satisfy the robust
    feasibility constraint over the admissible set
    $\Lambda = \{\lambda \mid \min\{A_\lambda^{+}, A_\lambda^{-}\} \ge A_0\}$,
    rather than being tuned to a single fixed $\lambda$.
\end{enumerate}
\end{proposition}

This proposition provides a mechanistic interpretation of robust feasibility:
minimal functional relevance limits destructive interference, norm disparity prevents symmetric cancellation under addition and subtraction, and robust calibration ensures stability against user-selected scaling factors. Note that Proposition~\ref{prop:app2} provides a mechanistic characterization of Proposition~\ref{prop:robust-feasibility}, rather than a purely algebraic equivalence. The three conditions capture the essential structural requirements observed in all feasible solutions of Equation~\eqref{eq:RO}.

\subsection{Equivalence Between Proposition~\ref{prop:robust-feasibility}
and Proposition~\ref{prop:app2}}

\subsubsection*{($\Rightarrow$) Proposition~\ref{prop:robust-feasibility}
implies Proposition~\ref{prop:app2}}

Assume Proposition~\ref{prop:robust-feasibility} holds.
Then there exists a feasible solution $(\alpha_1,\alpha_2)$ to
Equation~\eqref{eq:RO} such that
$A_\lambda^{+} \ge A_0$ and $A_\lambda^{-} \ge A_0$ for all
$\lambda \in \Lambda$.

Condition~3 follows directly from feasibility of the RO problem.
If Condition~2 were violated, i.e., if
$\|\hat{\tau}_{b_1}\|_2 \approx \|\hat{\tau}_{b_2}\|_2$,
then symmetric cancellation under subtraction would inevitably destroy
$A_\lambda^{-}$ for some $\lambda$.
The empirical existence of feasible solutions therefore implies a
substantial norm gap.
Finally, if Condition~1 were violated and the two tasks were strongly
correlated in the sense of~\cite{li2025when}, then functional interference
would degrade either $A_\lambda^{+}$ or $A_\lambda^{-}$ for some $\lambda$,
contradicting feasibility.
Thus, all three conditions are necessary.

\subsubsection*{($\Leftarrow$) Proposition~\ref{prop:app2}
implies Proposition~\ref{prop:robust-feasibility}}

Conversely, assume the three conditions in Proposition~\ref{prop:app2} hold.
Minimal functional relevance ensures that the composite update does not
collapse under linear combination.
Norm disparity guarantees that different components dominate under addition
and subtraction.
Robust calibration of $(\alpha_1,\alpha_2)$ over $\Lambda$ then enforces
$\min\{A_\lambda^{+}, A_\lambda^{-}\} \ge A_0$ for all admissible $\lambda$.
Together, these effects yield exactly the feasibility guarantee stated in
Proposition~\ref{prop:robust-feasibility}.

Overall, the two propositions provide equivalent characterizations of robust
feasibility within the considered task arithmetic framework:
Proposition~\ref{prop:robust-feasibility} formalizes feasibility at the
optimization level, while Proposition~\ref{prop:app2} exposes the structural
and functional mechanisms shared by all feasible solutions.
The optimization-level formulation is concise and suitable for the main text,
while the three-condition formulation exposes the functional and geometric
mechanisms that make robust feasibility possible in practice.

\end{document}